\documentclass[final]{elsarticle}
\usepackage{soul}
\usepackage{lineno}
\modulolinenumbers[5]
\usepackage{units}
\usepackage{tikz}
\usetikzlibrary{positioning}
\usepackage{algorithm}
\usepackage{enumitem}
\usepackage{float}
\usepackage{algpseudocode}

\usepackage{graphicx}
\usepackage{varioref}
\usepackage{array}
\usepackage{amsmath}
\usepackage{booktabs}
\usepackage{multirow}
\usepackage{subcaption}
\usepackage{array,arydshln}
\usepackage{amssymb}
\usepackage{hhline}
\usepackage{times,rotating}
\usepackage{comment}
\usepackage[hyphens,spaces,obeyspaces]{url}
\usepackage{tikz}
 
 \tikzset{variable/.default=}
\usepackage{tabularx,ragged2e}
\newcolumntype{T}[1]{>{\raggedright\arraybackslash}p{#1}}
\newcolumntype{M}[1]{>{\centering\arraybackslash}m{#1}}

\newcolumntype{L}[1]{>{\raggedright\let\newline\\\arraybackslash\hspace{0pt}}m{#1}}
\newcolumntype{C}[1]{>{\centering\let\newline\\\arraybackslash\hspace{0pt}}m{#1}}
\newcolumntype{R}[1]{>{\raggedleft\let\newline\\\arraybackslash\hspace{0pt}}m{#1}} 

\usepackage{tikz}
\usepackage{forest}
\usetikzlibrary{positioning}
\useforestlibrary{linguistics}
\forestapplylibrarydefaults{linguistics}
\usepackage{pifont}
\newcommand{\xmark}{\ding{55}}

\usepackage{framed} 
\usepackage{multicol} 
\usepackage{nomencl} 

\usepackage{booktabs}       
\usepackage{amsfonts}       
\usepackage{nicefrac}       
\usepackage{microtype}      
\usepackage{diagbox} 
\usepackage{dsfont}
\usepackage{bm} 
\usepackage{amsmath}
\usepackage{graphicx}
\usepackage{amsfonts}
\usepackage{subcaption}
\usepackage{stmaryrd}
\usepackage{color, colortbl}
\usepackage{breakcites}
\usepackage{pdfpages}
\usepackage{rotating}
\usepackage{dirtree} 

\makenomenclature
\renewcommand*\nompreamble{\begin{multicols}{2}}
\renewcommand*\nompostamble{\end{multicols}}

\usepackage[margin=3.5cm]{geometry}

\journal{Information Fusion}

\begin{document}

\begin{frontmatter}
	
\title{Explainable Artificial Intelligence (XAI): Concepts, Taxonomies, Opportunities and Challenges toward Responsible AI}

\author[a]{Alejandro Barredo Arrieta}
\author[g]{Natalia D\'iaz-Rodr\'iguez}
\author[a,c,d]{Javier Del Ser}
\author[g,h,i]{Adrien Bennetot}
\author[f]{\\Siham Tabik}
\author[j]{Alberto Barbado}
\author[f]{Salvador Garcia}
\author[a]{Sergio Gil-Lopez}
\author[f]{Daniel Molina}
\author[j]{\\Richard Benjamins}
\author[i]{Raja Chatila}
\author[f]{and Francisco Herrera}

\address[a]{TECNALIA, 48160 Derio, Spain}
\address[g]{ENSTA, Institute Polytechnique Paris and INRIA Flowers Team, Palaiseau, France}
\address[c]{University of the Basque Country (UPV/EHU), 48013 Bilbao, Spain}
\address[d]{Basque Center for Applied Mathematics (BCAM), 48009 Bilbao, Bizkaia, Spain}
\address[h]{Segula Technologies, Parc d'activit\'e de Pissaloup, Trappes, France}
\address[i]{Institut des Syst\`emes Intelligents et de Robotique, Sorbonne Universit\`e, France}
\address[f]{DaSCI Andalusian Institute of Data Science and Computational Intelligence, University of Granada, 18071 Granada, Spain}
\address[j]{Telefonica, 28050 Madrid, Spain}
\cortext[cor1]{Corresponding author. TECNALIA. P. Tecnologico, Ed. 700. 48170 Derio (Bizkaia), Spain. E-mail: javier.delser@tecnalia.com}

\begin{abstract}
In the last few years, Artificial Intelligence (AI) has achieved a notable momentum that, if harnessed appropriately, may deliver the best of expectations over many application sectors across the field. For this to occur shortly in Machine Learning, the entire community stands in front of the barrier of explainability, an inherent problem of the latest techniques brought by sub-symbolism (e.g. ensembles or Deep Neural Networks) that were not present in the last hype of AI (namely, expert systems and rule based models). Paradigms underlying this problem fall within the so-called \textit{eXplainable} AI (XAI) field, which is widely acknowledged as a crucial feature for the practical deployment of AI models. The overview presented in this article examines the existing literature and contributions already done in the field of XAI, including a prospect toward what is yet to be reached. For this purpose we summarize previous efforts made to define explainability in Machine Learning, establishing a novel definition of explainable Machine Learning that covers such prior conceptual propositions with a major focus on the audience for which the explainability is sought. Departing from this definition, we propose and discuss about a taxonomy of recent contributions related to the explainability of different Machine Learning models, including those aimed at explaining Deep Learning methods for which a second dedicated taxonomy is built and examined in detail. This critical literature analysis serves as the motivating background for a series of challenges faced by XAI, such as the interesting crossroads of data fusion and explainability. Our prospects lead toward the concept of \emph{Responsible Artificial Intelligence}, namely, a methodology for the large-scale implementation of AI methods in real organizations with fairness, model explainability and accountability at its core. Our ultimate goal is to provide newcomers to the field of XAI with a thorough taxonomy that can serve as reference material in order to stimulate future research advances, but also to encourage experts and professionals from other disciplines to embrace the benefits of AI in their activity sectors, without any prior bias for its lack of interpretability.
\end{abstract}


\end{frontmatter}


\section{Introduction} \label{sec:intro}

Artificial Intelligence (AI) lies at the core of many activity sectors that have embraced new information technologies \cite{russell2016artificial}. While the roots of AI trace back to several decades ago, there is a clear consensus on the paramount importance featured nowadays by intelligent machines endowed with learning, reasoning and adaptation capabilities. It is by virtue of these capabilities that AI methods are achieving unprecedented levels of performance when learning to solve increasingly complex computational tasks, making them pivotal for the future development of the human society \cite{west2018future}. The sophistication of AI-powered systems has lately increased to such an extent that almost no human intervention is required for their design and deployment. When decisions derived from such systems ultimately affect humans' lives (as in e.g. medicine, law or defense), there is an emerging need for understanding how such decisions are furnished by AI methods \cite{goodman2017Fair}.
 
While the very first AI systems were easily interpretable, the last years have witnessed the rise of opaque decision systems such as Deep Neural Networks (DNNs). The empirical success of Deep Learning (DL) models such as DNNs stems from a combination of efficient learning algorithms and their huge parametric space. The latter space comprises hundreds of layers and millions of parameters, which makes DNNs be considered as complex \textit{black-box} models \cite{Castelvecchi16}. The opposite of \textit{black-box-ness} is \textit{transparency}, i.e., the search for a direct understanding of the mechanism by which a model works \cite{Lipton18}.

As black-box Machine Learning (ML) models are increasingly being employed to make important predictions in critical contexts, the demand for transparency is increasing from the various stakeholders in AI \cite{Preece18Stakeholders}. The danger is on creating and using decisions that are not justifiable, legitimate, or that simply do not allow obtaining detailed explanations of their behaviour \cite{gunning2017explainable}. Explanations supporting the output of a model are crucial, e.g., in precision medicine, where experts require far more information from the model than a simple binary prediction for supporting their diagnosis \cite{1907.07374}. Other examples include autonomous vehicles in transportation, security, and finance, among others. 

In general, humans are reticent to adopt techniques that are not directly interpretable, tractable and trustworthy \cite{Zhu18}, given the increasing demand for ethical AI \cite{goodman2017Fair}. It is customary to think that by focusing solely on performance, the systems will be increasingly opaque. This is true in the sense that there is a trade-off between the performance of a model and its transparency \cite{Dosilovic18}. However, an improvement in the understanding of a system can lead to the correction of its deficiencies. When developing a ML model, the consideration of interpretability as an additional design driver can improve its implementability for 3 reasons: 
\begin{itemize}[leftmargin=*]
\item Interpretability helps ensure impartiality in decision-making, i.e. to detect, and consequently, correct from bias in the training dataset.
\item Interpretability facilitates the provision of robustness by highlighting potential adversarial perturbations that could change the prediction. 
\item Interpretability can act as an insurance that only meaningful variables infer the output, i.e., guaranteeing that an underlying truthful causality exists in the model reasoning.
\end{itemize}

All these means that the interpretation of the system should, in order to be considered practical, provide either an understanding of the model mechanisms and predictions, a visualization of the model's discrimination rules, or hints on what could perturb the model \cite{Hall2018}.

In order to avoid limiting the effectiveness of the current generation of AI systems, \textit{eXplainable AI} (XAI) \cite{gunning2017explainable} proposes creating a suite of ML techniques that 1) produce more explainable models while maintaining a high level of learning performance (e.g., prediction accuracy), and 2) enable humans to understand, appropriately trust, and effectively manage the emerging generation of artificially intelligent partners. XAI draws as well insights from the Social Sciences \cite{Miller19} and considers the psychology of explanation. 
\begin{figure}[htpb]
        \center{\includegraphics[width=\columnwidth]{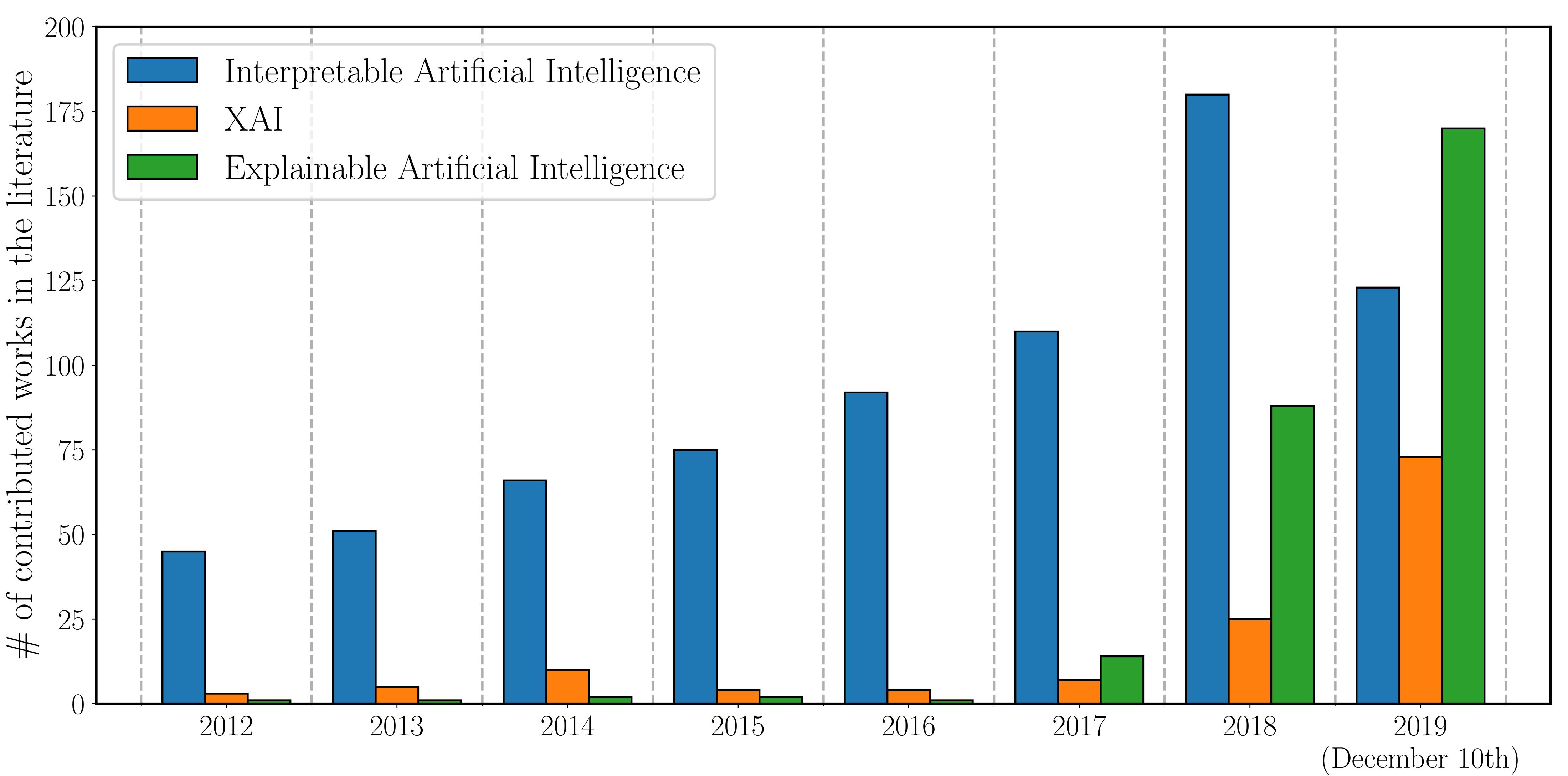}}
        \caption{Evolution of the number of total publications whose title, abstract and/or keywords refer to the field of XAI during the last years. Data retrieved from Scopus\textsuperscript{\textregistered} {\color{black}(December 10th, 2019) by using the search terms indicated in the legend when querying this database}. It is interesting to note the latent need for interpretable AI models over time (which conforms to intuition, as interpretability is a requirement in many scenarios), yet it has not been until 2017 when the interest in techniques to explain AI models has permeated throughout the research community.}
        \label{fig:xAITrend}
      \end{figure}
      
Figure \ref{fig:xAITrend} displays the rising trend of contributions on XAI and related concepts. This literature outbreak shares its rationale with the research agendas of national governments and agencies. Although some recent surveys \cite{1907.07374,Gilpin18,Dosilovic18,adadi2018peeking,biran2017explanation,Darpa2019,Guidotti19} summarize the upsurge of activity in XAI across sectors and disciplines, this overview aims to cover the creation of a complete unified framework of categories and concepts that allow for scrutiny and understanding of the field of XAI methods. Furthermore, we pose intriguing thoughts around the explainability of AI models in data fusion contexts with regards to data privacy and model confidentiality. This, along with other research opportunities and challenges identified throughout our study, serve as the pull factor toward {Responsible Artificial Intelligence}, term by which we refer to a series of AI principles to be necessarily met when deploying AI in real applications. As we will later show in detail, model explainability is among the most crucial aspects to be ensured within this methodological framework. All in all, the novel contributions of this overview can be summarized as follows:
\begin{enumerate}[leftmargin=*]
\item Grounded on a first elaboration of concepts and terms used in XAI-related research, we propose a novel definition of explainability that places \emph{audience} (Figure \ref{fig:audiences}) as a key aspect to be considered when explaining a ML model. We also elaborate on the diverse purposes sought when using XAI techniques, from trustworthiness to privacy awareness, which round up the claimed importance of purpose and targeted audience in model explainability.

\item We define and examine the different levels of transparency that a ML model can feature by itself, as well as the diverse approaches to post-hoc explainability, namely, the explanation of ML models that are not transparent by design.

\item We thoroughly analyze the literature on XAI and related concepts published to date, covering approximately 400 contributions arranged into two different taxonomies. The first taxonomy addresses the explainability of ML models using the previously made distinction between transparency and post-hoc explainability, including models that are transparent by themselves, Deep and non-Deep (i.e., \emph{shallow}) learning models. The second taxonomy deals with XAI methods suited for the explanation of Deep Learning models, using classification criteria closely linked to this family of ML methods (e.g. layerwise explanations, representation vectors, attention).

\item We enumerate a series of challenges of XAI that still remain insufficiently addressed to date. Specifically, we identify research needs around the concepts and metrics to evaluate the explainability of ML models, and outline research directions toward making Deep Learning models more understandable. We further augment the scope of our prospects toward the implications of XAI techniques in regards to confidentiality, robustness in adversarial settings, data diversity, and other areas intersecting with explainability.

\item After the previous prospective discussion, we arrive at the concept of Responsible Artificial Intelligence, a manifold concept that imposes the systematic adoption of several AI principles for AI models to be of practical use. In addition to explainability, the guidelines behind Responsible AI establish that fairness, accountability and privacy should also be considered when implementing AI models in real environments. 

\item Since Responsible AI blends together model explainability and privacy/security by design, we call for a profound reflection around the benefits and risks of XAI techniques in scenarios dealing with sensitive information and/or confidential ML models. As we will later show, the regulatory push toward data privacy, quality, integrity and governance demands more efforts to assess the role of XAI in this arena. In this regard, we provide an insight on the implications of XAI in terms of privacy and security under different data fusion paradigms.
\end{enumerate}

The remainder of this overview is structured as follows: first, Section \ref{sec:xaiwwwh} and subsections therein open a discussion on the terminology and concepts revolving around explainability and interpretability in AI, ending up with the aforementioned novel definition of interpretability (Subsections \ref{sec:terminology} and \ref{sec:what}), and a general criterion to categorize and analyze ML models from the XAI perspective. Sections \ref{sec:transparent} and \ref{sec:posthoc} proceed by reviewing recent findings on XAI for ML models (on transparent models and post-hoc techniques respectively) that comprise the main division in the aforementioned taxonomy. We also include a review on hybrid approaches among the two, to attain XAI. Benefits and caveats of the synergies among the families of methods are discussed in Section \ref{sec:challenges}, where we present a prospect of general challenges and some consequences to be cautious about. Finally, Section \ref{sec:responsibleAI} elaborates on the concept of Responsible Artificial Intelligence. Section \ref{sec:conc} concludes the survey with an outlook aimed at engaging the community around this vibrant research area, which has the potential to impact society, in particular those sectors that have progressively embraced ML as a core technology of their activity.

\section{Explainability: What, Why, What For and How?} \label{sec:xaiwwwh}

Before proceeding with our literature study, it is convenient to first establish a common point of understanding on what the term \emph{explainability} stands for in the context of AI and, more specifically, ML. This is indeed the purpose of this section, namely, to pause at the numerous definitions that have been done in regards to this concept (what?), to argue why explainability is an important issue in AI and ML (why? what for?) and to introduce the general classification of XAI approaches that will drive the literature study thereafter (how?).

\subsection{Terminology Clarification} \label{sec:terminology}

One of the issues that hinders the establishment of common grounds is the interchangeable misuse of interpretability and explainability in the literature. There are notable differences among these concepts. To begin with, interpretability refers to a passive characteristic of a model referring to the level at which a given model makes sense for a human observer. This feature is also expressed as transparency. By contrast, explainability can be viewed as an active characteristic of a model, denoting any action or procedure taken by a model with the intent of clarifying or detailing its internal functions.

To summarize the most commonly used nomenclature, in this section we clarify the distinction and similarities among terms often used in the ethical AI and XAI communities.
\begin{itemize}[leftmargin=*]
\item \textbf{Understandability} (or equivalently, \textbf{intelligibility}) denotes the characteristic of a model to make a human understand its function -- how the model works -- without any need for explaining its internal structure or the algorithmic means by which the model processes data internally \cite{Montavon18}. 

\item \textbf{Comprehensibility}: when conceived for ML models, comprehensibility refers to the ability of a learning algorithm to represent its learned knowledge in a human understandable fashion \cite{Fernandez19,gleicher2016framework,craven1996extracting}. This notion of model comprehensibility stems from the postulates of Michalski \cite{michalski1983theory}, which stated that \emph{``the results of computer induction should be symbolic descriptions of given entities, semantically and structurally similar to those a human expert might produce observing the same entities. Components of these descriptions should be comprehensible as single `chunks' of information, directly interpretable in natural language, and should relate quantitative and qualitative concepts in an integrated fashion''}. Given its difficult quantification, comprehensibility is normally tied to the evaluation of the model complexity \cite{Guidotti19}. 

\item \textbf{Interpretability}: it is defined as the ability to explain or to provide the meaning in understandable terms to a human. 

\item \textbf{Explainability}: explainability is associated with the notion of explanation as an interface between humans and a decision maker that is, at the same time, both an accurate proxy of the decision maker and comprehensible to humans \cite{Guidotti19}.

\item \textbf{Transparency}: a model is considered to be transparent if by itself it is understandable. Since a model can feature different degrees of understandability, transparent models in Section \ref{sec:transparent} are divided into three categories: simulatable models, decomposable models and algorithmically transparent models \cite{Lipton18}.
\end{itemize}

{\color{black}In all the above definitions, \emph{understandability} emerges as the most essential concept in XAI. Both transparency and interpretability are strongly tied to this concept: while transparency refers to the characteristic of a model to be, on its own, understandable for a human, understandability measures the degree to which a human can understand a decision made by a model. Comprehensibility is also connected to understandability in that it relies on the capability of the audience to understand the knowledge contained in the model. All in all, understandability is a two-sided matter: model understandability and human understandability. This is the reason why the definition of XAI given in Section \ref{sec:what} refers to the concept of \emph{audience}, as the cognitive skills and pursued goal of the users of the model have to be taken into account jointly with the intelligibility and comprehensibility of the model in use. This prominent role taken by understandability makes the concept of \emph{audience} the cornerstone of XAI, as we next elaborate in further detail}.

\subsection{What?} \label{sec:what}

Although it might be considered to be beyond the scope of this paper, it is worth noting the discussion held around general theories of explanation in the realm of philosophy \cite{diez2013Explanations}. Many proposals have been done in this regard, suggesting the need for a general, unified theory that approximates the structure and intent of an explanation. However, nobody has stood the critique when presenting such a general theory. For the time being, the most agreed-upon thought blends together different approaches to explanation drawn from diverse knowledge disciplines. A similar problem is found when addressing interpretability in AI. It appears from the literature that there is not yet a common point of understanding on what interpretability or explainability are. However, many contributions claim the achievement of interpretable models and techniques that empower explainability. 

To shed some light on this lack of consensus, it might be interesting to place the reference starting point at the definition of the term Explainable Artificial Intelligence (XAI) given by D. Gunning in \cite{gunning2017explainable}:
\begin{center}
\noindent\fbox{\parbox{0.97\textwidth}{\noindent\textit{``XAI will create a suite of machine learning techniques that enables human users to understand, appropriately trust, and effectively manage the emerging generation of artificially intelligent partners''}}}
\end{center}

This definition brings together two concepts (understanding and trust) that need to be addressed in advance. However, it misses to consider other purposes motivating the need for interpretable AI models, such as causality, transferability, informativeness, fairness and confidence \cite{Lipton18,WhatDoesExplainableAImean,TowardsInterpretability,MakingInterpretable}. We will later delve into these topics, mentioning them here as a supporting example of the incompleteness of the above definition.

As exemplified by the definition above, a thorough, complete definition of explainability in AI still slips from our fingers. A broader reformulation of this definition (e.g. \textit{``An explainable Artificial Intelligence is one that produces explanations about its functioning''}) would fail to fully characterize the term in question, leaving aside important aspects such as its purpose. To build upon the completeness, a definition of explanation is first required. 

As extracted from the Cambridge Dictionary of English Language, an explanation is \textit{``the details or reasons that someone gives to make something clear or easy to understand''} \cite{walter2008cambridge}. In the context of an ML model, this can be rephrased as: \textit{"the details or reasons a model gives to make its functioning clear or easy to understand"}. It is at this point where opinions start to diverge. Inherently stemming from the previous definitions, two ambiguities can be pointed out. First, the details or the reasons used to explain, are completely dependent of the audience to which they are presented. Second, whether the explanation has left the concept clear or easy to understand also depends completely on the audience. Therefore, the definition must be rephrased to reflect explicitly the dependence of the explainability of the model on the audience. To this end, a reworked definition could read as: 
\begin{center}
	\noindent\fbox{\parbox{0.97\textwidth}{\noindent\textit{\textit{Given a certain audience, explainability refers to the details and reasons a model gives to make its functioning clear or easy to understand.}}}}
\end{center}

Since explaining, as argumenting, may involve weighting, comparing or convincing an audience with logic-based formalizations of (counter) arguments \cite{Besnard08}, explainability might convey us into the realm of cognitive psychology and the \textit{psychology of explanations} \cite{gunning2017explainable}, since measuring whether something has been understood or put clearly is a hard task to be gauged objectively. However, measuring to which extent the internals of a model can be explained could be tackled objectively. Any means to reduce the complexity of the model or to simplify its outputs should be considered as an XAI approach. How big this leap is in terms of complexity or simplicity will correspond to how explainable the resulting model is. An underlying problem that remains unsolved is that the interpretability gain provided by such XAI approaches may not be straightforward to quantify: for instance, a model simplification can be evaluated based on the reduction of the number of architectural elements or number of parameters of the model itself (as often made, for instance, for DNNs). On the contrary, the use of visualization methods or natural language for the same purpose does not favor a clear quantification of the improvements gained in terms of interpretability. The derivation of general metrics to assess the quality of XAI approaches remain as an open challenge that should be under the spotlight of the field in forthcoming years. We will further discuss on this research direction in Section \ref{sec:challenges}.

Explainability is linked to post-hoc explainability since it covers the techniques used to convert a non-interpretable model into a explainable one. In the remaining of this manuscript, explainability will be considered as the main design objective, since it represents a broader concept. A model can be explained, but the interpretability of the model is something that comes from the design of the model itself. Bearing these observations in mind, explainable AI can be defined as follows: 
\begin{center}
	\noindent\fbox{\parbox{0.97\textwidth}{\noindent\textit{\textit{Given an audience, an \textbf{explainable} Artificial Intelligence is one that produces details or reasons to make its functioning clear or easy to understand.}}}}
\end{center}

This definition is posed here as a first contribution of the present overview, implicitly assumes that the ease of understanding and clarity targeted by XAI techniques for the model at hand reverts on different application purposes, such as a better trustworthiness of the model's output by the audience.

\subsection{Why?}

\begin{figure}[h]
\center{\includegraphics[width=0.9\columnwidth]{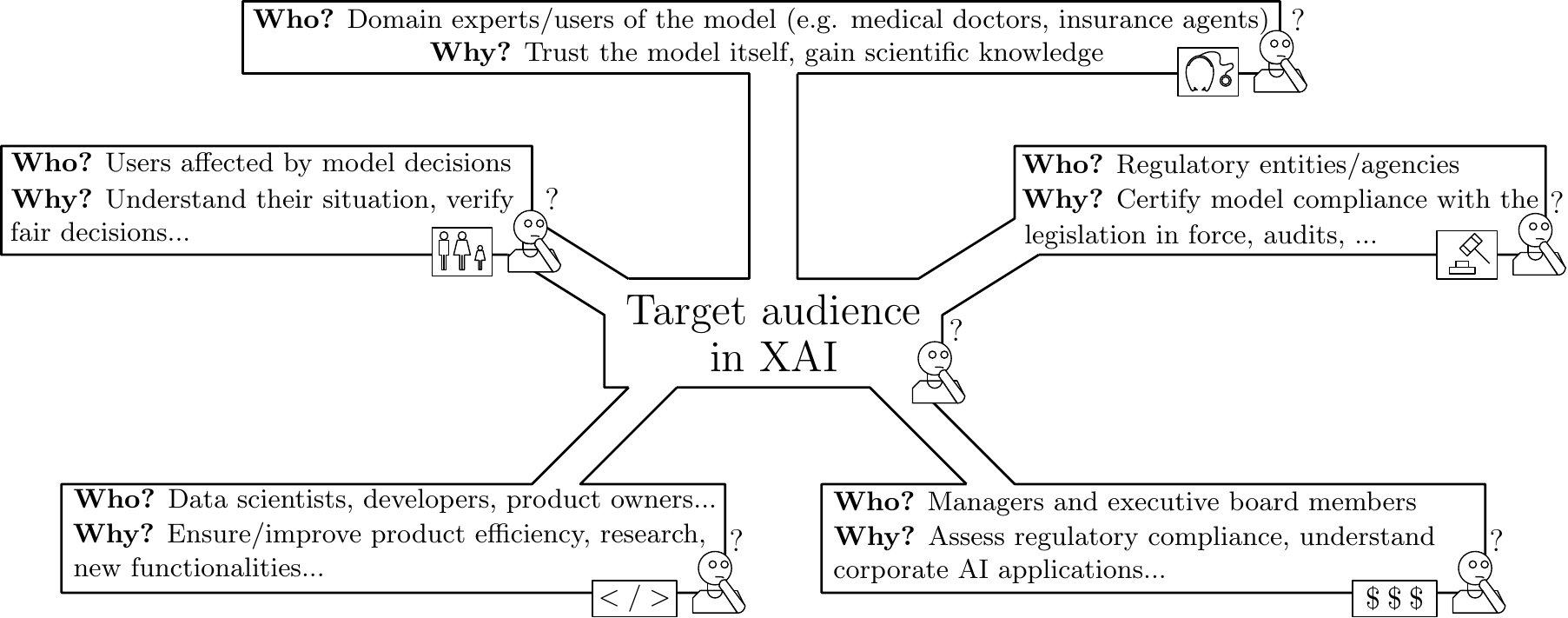}}
\caption{Diagram showing the different purposes of explainability in ML models sought by different audience profiles. Two goals occur to prevail across them: need for model understanding, and regulatory compliance. Image partly inspired by the one presented in \cite{ibm2019}, used with permission from IBM.} 
\label{fig:audiences}
\end{figure}

As stated in the introduction, explainability is one of the main barriers AI is facing nowadays in regards to its practical implementation. The inability to explain or to fully understand the reasons by which state-of-the-art ML algorithms perform as well as they do, is a problem that find its roots in two different causes, which are conceptually illustrated in Figure \ref{fig:audiences}. 

Without a doubt, the first cause is the gap between the research community and business sectors, impeding the full penetration of the newest ML models in sectors that have traditionally lagged behind in the digital transformation of their processes, such as banking, finances, security and health, among many others. In general this issue occurs in strictly regulated sectors with some reluctance to implement techniques that may put at risk their assets. 

The second axis is that of knowledge. AI has helped research across the world with the task of inferring relations that were far beyond the human cognitive reach. Every field dealing with huge amounts of reliable data has largely benefited from the adoption of AI and ML techniques. However, we are entering an era in which results and performance metrics are the only interest shown up in research studies. Although for certain disciplines this might be the fair case, science and society are far from being concerned just by performance. The search for understanding is what opens the door for further model improvement and its practical utility.

The following section develops these ideas further by analyzing the goals motivating the search for explainable AI models.

\subsection{What for?}

The research activity around XAI has so far exposed different goals to draw from the achievement of an explainable model. Almost none of the papers reviewed completely agrees in the goals required to describe what an explainable model should compel. However, all these different goals might help discriminate the purpose for which a given exercise of ML explainability is performed. Unfortunately, scarce contributions have attempted to define such goals from a conceptual perspective \cite{Lipton18,Gilpin18,WhatDoesExplainableAImean,WhatDoWeNeed}. We now synthesize and enumerate definitions for these XAI goals, so as to settle a first classification criteria for the full suit of papers covered in this review:
\begin{itemize}[leftmargin=*]
    \item \textit{Trustworthiness:} several authors agree upon the search for trustworthiness as the primary aim of an explainable AI model \cite{kim2015Trust,ribeiro2016trust}. However, declaring a model as explainable as per its capabilities of inducing trust might not be fully compliant with the requirement of model explainability. Trustworthiness might be considered as the confidence of whether a model will act as intended when facing a given problem. Although it should most certainly be a property of any explainable model, it does not imply that every trustworthy model can be considered explainable on its own, nor is trustworthiness a property easy to quantify. Trust might be far from being the only purpose of an explainable model since the relation among the two, if agreed upon, is not reciprocal. Part of the reviewed papers mention the concept of trust when stating their purpose for achieving explainability. However, as seen in Table \ref{tab:ExplainabilityGoals}, they do not amount to a large share of the recent contributions related to XAI. 
\begin{table}[t]
	\centering
	\resizebox{0.9\textwidth}{!}{%
		\begin{tabular}{C{3cm}L{5cm}L{7.5cm}}
			\specialrule{.2em}{.1em}{.1em}
			\textbf{XAI Goal} & {\color{black}\textbf{Main target audience} (Fig. 2)} & \textbf{References} \\ 
			\specialrule{.2em}{.1em}{.1em}
			Trustworthiness & {\color{black}Domain experts, users of the model affected by decisions} & \cite{Lipton18,Dosilovic18,WhatDoesExplainableAImean,ribeiro2016trust,fox2017explainable,lane2005explainable,murdoch2019interpretable,ExplanationsExpectations,chander2018working} \\ \hline
			
			Causality & {\color{black}Domain experts, managers and executive board members, regulatory entities/agencies} & \cite{murdoch2019interpretable,tickle1998truth,louizos2017causal,goudet2018learning,athey2015machine, Lopez-Paz17, barabas2017interventions} \\ \hline
			
			Transferability & {\color{black}Domain experts, data scientists} & \cite{Lipton18,caruana2015Transferability,craven1996extracting,MakingInterpretable,theodorou2017designing,WhatDoWeNeed,ribeiro2016trust,chander2018working,tickle1998truth,louizos2017causal,samek2017explainable,wadsworth2018achieving,yuan2019adversarial,letham2015interpretable,harbers2010design,aung2007comparing,weller2017challenges,freitas2014comprehensible,schetinin2007confident,martens2011performance,InterpretableDeepICU,barakat2008,ExplainYourselfAgents,ExplainableAgencyAgents,DeepTaylor,LearningHowTo,ras2018explanation,bach2016controlling,MedicinePrecision,UsingPerceptualAgents,olden2002illuminating,krause2016interacting,interpretingHeatMapSVM,tan2014unsupervised,krening2017learning,LIME,LayerWise,etchells2006orthogonal,PlanExplicabilityAgents,santoro2017simple,peng2002use,ustun2007visualisation,zhang2019interpreting,wu2018beyond,hinton2015distilling,frosst2017distilling,augasta2012reverse,zhou2003extracting,tan2016tree,fong2017interpretable}\\ 
			\hline
			Informativeness & {\color{black}All} & \cite{Lipton18,caruana2015Transferability,craven1996extracting,TowardsInterpretability,MakingInterpretable,theodorou2017designing,WhatDoWeNeed,ribeiro2016trust,lane2005explainable,murdoch2019interpretable,chander2018working,tickle1998truth,athey2015machine,samek2017explainable,letham2015interpretable,harbers2010design,aung2007comparing,weller2017challenges,freitas2014comprehensible,schetinin2007confident,martens2011performance,InterpretableDeepICU,barakat2008,ExplainYourselfAgents,ExplainableAgencyAgents,DeepTaylor,bach2016controlling,MedicinePrecision,UsingPerceptualAgents,olden2002illuminating,interpretingHeatMapSVM,tan2014unsupervised,krening2017learning,LIME,LayerWise,etchells2006orthogonal,PlanExplicabilityAgents,santoro2017simple,peng2002use,ustun2007visualisation,zhang2019interpreting,wu2018beyond,UsersAtChargeOfDesing,goebel2018explainable,belle2017logic,edwards2017slave,ExplainableAgencyAgents,lou2013accurate,xu2015show,huysmans2011Informativeness,barakat2007rule,Chaves2005,martens2007comprehensible,LearningDeepFeatures,krishnan1999extracting,fu2004,green2018fair,chouldechova2017fair,kim2018fairness,haasdonk2005feature,FeatureContributionMethod,welling2016forest,fung2005,zhang2005,linsley2018global,zhou2008low,burrell2016machine,shrikumar2016not,ImprovingInterpretability,ridgeway1998interpretable,InterpretableCNN,seo2017interpretable,larsen2000interpreting,interpretingNeuroSVM,xu2018interpreting,Intrees,domingos1998knowledge,DistillAndCompare,StatisticsForCriminalBehavior,MakingTEInterpretable,AtributeInteractions,MIRIAMAgents,QuantifyingInterpretability,nunez2002rule,nunez2006rule,kearns2017preventing,akyol2016price,erhan2010understanding,zhang2015sensitivity,quinlan1987simplifying,SingleTreeApproximation,intepretationSVM,TreeView,VisualizingUnderstanding,UnderstandingDeep,wagner2019interpretable,kanehira2019learning,apley2016visualizing,staniak2018explanations,zeiler2010deconvolutional, springenberg2014striving,kim2017interpretability,polino2018model,murdoch2017automatic,craven1994using,arbatli1997rule,johansson2009evolving,lei2016rationalizing,radford2017learning,selvaraju2016grad,shwartz2017opening,yosinski2015understanding} \\ \hline
			Confidence & {\color{black}Domain experts, developers, managers, regulatory entities/agencies} & \cite{Lipton18,theodorou2017designing,murdoch2019interpretable,samek2017explainable,yuan2019adversarial,schetinin2007confident,LearningHowTo,LayerWise,belle2017logic,edwards2017slave,LearningDeepFeatures,zhou2008low,xu2018interpreting,domingos1998knowledge, pope2019explainability}\\ \hline
			
			Fairness & {\color{black}Users affected by model decisions, regulatory entities/agencies} & \cite{Lipton18,WhatDoesExplainableAImean,theodorou2017designing,murdoch2019interpretable,wadsworth2018achieving,green2018fair,chouldechova2017fair,kim2018fairness,DistillAndCompare,StatisticsForCriminalBehavior,kearns2017preventing,gajane2017formalizing,dwork2018group,barocas-hardt-narayanan19} \\ \hline
			
			Accessibility & {\color{black}Product owners, managers, users affected by model decisions} & \cite{craven1996extracting,MakingInterpretable,WhatDoWeNeed,ribeiro2016trust,chander2018working,harbers2010design,freitas2014comprehensible,martens2011performance,ras2018explanation,krause2016interacting,interpretingHeatMapSVM,tan2014unsupervised,krening2017learning,LIME,PlanExplicabilityAgents,santoro2017simple,peng2002use,UsersAtChargeOfDesing,barakat2007rule,Chaves2005,FeatureContributionMethod,fung2005,linsley2018global,zhou2008low,ImprovingInterpretability,ridgeway1998interpretable,InterpretableCNN,seo2017interpretable,larsen2000interpreting,MIRIAMAgents,akyol2016price}\\ \hline
			
			Interactivity & {\color{black}Domain experts, users affected by model decisions} & \cite{chander2018working,harbers2010design,ExplainableAgencyAgents,UsingPerceptualAgents,krause2016interacting,PlanExplicabilityAgents,UsersAtChargeOfDesing,MIRIAMAgents} \\ \hline
			Privacy awareness & {\color{black}Users affected by model decisions, regulatory entities/agencies} & \cite{edwards2017slave}\\ 
			\specialrule{.2em}{.1em}{.1em}
		\end{tabular}%
	}
	\caption{Goals pursued in the reviewed literature toward reaching explainability{\color{black}, and their main target audience}.}
	\label{tab:ExplainabilityGoals}
\end{table}

    \item \textit{Causality:} another common goal for explainability is that of finding causality among data variables. Several authors argue that explainable models might ease the task of finding relationships that, should they occur, could be tested further for a stronger causal link between the involved variables \cite{wang1999Causality,rani2006Causality}. The inference of causal relationships from observational data is a field that has been broadly studied over time \cite{pearl2009causality}. As widely acknowledged by the community working on this topic, causality requires a wide frame of prior knowledge to prove that observed effects are causal. A ML model only discovers correlations among the data it learns from, and therefore might not suffice for unveiling a cause-effect relationship. However, causation involves correlation, so an explainable ML model could validate the results provided by causality inference techniques, or provide a first intuition of possible causal relationships within the available data. Again, Table \ref{tab:ExplainabilityGoals} reveals that causality is not among the most important goals if we attend to the amount of papers that state it explicitly as their goal.
    
    \item \textit{Transferability:} models are always bounded by constraints that should allow for their seamless transferability. This is the main reason why a training-testing approach is used when dealing with ML problems \cite{kuhn2013appliedTransferability, james2013Transferability}. Explainability is also an advocate for transferability, since it may ease the task of elucidating the boundaries that might affect a model, allowing for a better understanding and implementation. Similarly, the mere understanding of the inner relations taking place within a model facilitates the ability of a user to reuse this knowledge in another problem. There are cases in which the lack of a proper understanding of the model might drive the user toward incorrect assumptions and fatal consequences \cite{caruana2015Transferability,szegedy2013Transferability}. Transferability should also fall between the resulting properties of an explainable model, but again, not every transferable model should be considered as explainable. As observed in Table \ref{tab:ExplainabilityGoals}, the amount of papers stating that the ability of rendering a model explainable is to better understand the concepts needed to reuse it or to improve its performance is the second most used reason for pursuing model explainability.
    
    \item \textit{Informativeness:} ML models are used with the ultimate intention of supporting decision making \cite{huysmans2011Informativeness}. However, it should not be forgotten that the problem being solved by the model is not equal to that being faced by its human counterpart. Hence, a great deal of information is needed in order to be able to relate the user's decision to the solution given by the model, and to avoid falling in misconception pitfalls. For this purpose, explainable ML models should give information about the problem being tackled. Most of the reasons found among the papers reviewed is that of extracting information about the inner relations of a model. Almost all rule extraction techniques substantiate their approach on the search for a simpler understanding of what the model internally does, stating that the knowledge (information) can be expressed in these simpler proxies that they consider explaining the antecedent. This is the most used argument found among the reviewed papers to back up what they expect from reaching explainable models.
    
    \item \textit{Confidence:} as a generalization of robustness and stability, confidence should always be assessed on a model in which reliability is expected. The methods to maintain confidence under control are different depending on the model. As stated in \cite{ruppert1987Stability,basu2018Stability,yu2013stability}, stability is a must-have when drawing interpretations from a certain model. Trustworthy interpretations should not be produced by models that are not stable. Hence, an explainable model should contain information about the confidence of its working regime. 
    
    \item \textit{Fairness:} from a social standpoint, explainability can be considered as the capacity to reach and guarantee fairness in ML models. In a certain literature strand, an explainable ML model suggests a clear visualization of the relations affecting a result, allowing for a fairness or ethical analysis of the model at hand \cite{goodman2017Fair,chouldechova2017fair}. Likewise, a related objective of XAI is highlighting bias in the data a model was exposed to \cite{Burns18,Bennetot19}. The support of algorithms and models is growing fast in fields that involve human lives, hence explainability should be considered as a bridge to avoid the unfair or unethical use of algorithm's outputs.
    
    \item \textit{Accessibility:} a minor subset of the reviewed contributions argues for explainability as the property that allows end users to get more involved in the process of improving and developing a certain ML model \cite{chander2018working,UsersAtChargeOfDesing} . It seems clear that explainable models will ease the burden felt by non-technical or non-expert users when having to deal with algorithms that seem incomprehensible at first sight. This concept is expressed as the third most considered goal among the surveyed literature. 
    
    \item \textit{Interactivity:} some contributions \cite{harbers2010design,ExplainableAgencyAgents} include the ability of a model to be interactive with the user as one of the goals targeted by an explainable ML model. Once again, this goal is related to fields in which the end users are of great importance, and their ability to tweak and interact with the models is what ensures success.
    
    \item \textit{Privacy awareness:} almost forgotten in the reviewed literature, one of the byproducts enabled by explainability in ML models is its ability to assess privacy. ML models may have complex representations of their learned patterns. Not being able to understand what has been captured by the model \cite{Castelvecchi16} and stored in its internal representation may entail a privacy breach. Contrarily, the ability to explain the inner relations of a trained model by non-authorized third parties may also compromise the differential privacy of the data origin. Due to its criticality in sectors where XAI is foreseen to play a crucial role, confidentiality and privacy issues will be covered further in Subsections \ref{ssec:robust_adv} and \ref{ssec:privacydatafusion}, respectively.
    
\end{itemize}

This subsection has reviewed the goals encountered among the broad scope of the reviewed papers. All these goals are clearly under the surface of the concept of explainability introduced before in this section. To round up this prior analysis on the concept of explainability, the last subsection deals with different strategies followed by the community to address explainability in ML models.

\subsection{How?}

The literature makes a clear distinction among models that are interpretable by design, and those that can be explained by means of external XAI techniques. This duality could also be regarded as the difference between interpretable models and model interpretability techniques; a more widely accepted classification is that of \emph{transparent} models and post-hoc explainability. This same duality also appears in the paper presented in \cite{Guidotti19} in which the distinction its authors make refers to the methods to solve the transparent box design problem against the problem of explaining the black-box problem. This work, further extends the distinction made among transparent models including the different levels of transparency considered.

Within transparency, three levels are contemplated: algorithmic transparency, decomposability and simulatability\footnote{The alternative term \textit{simulability} is also used in the literature to refer to the capacity of a system or process to be simulated. However, we note that this term does not appear in current English dictionaries.}. Among post-hoc techniques we may distinguish among \textit{text explanations}, \textit{visualizations}, \textit{local explanations}, \textit{explanations by example}, \textit{explanations by simplification} and \textit{feature relevance}. In this context, there is a broader distinction proposed by \cite{WhatDoesExplainableAImean} discerning between 1) opaque systems, where the mappings from input to output are invisible to the user; 2) interpretable systems, in which users can mathematically analyze the mappings; and 3) comprehensible systems, in which the models should output symbols or rules along with their specific output to aid in the understanding process of the rationale behind the mappings being made. This last classification criterion could be considered included within the one proposed earlier, hence this paper will attempt at following the more specific one.

\subsubsection{Levels of Transparency in Machine Learning Models} \label{sec:transparent-models}

\begin{figure}[ht]
        \center{\includegraphics[width=0.95\columnwidth]{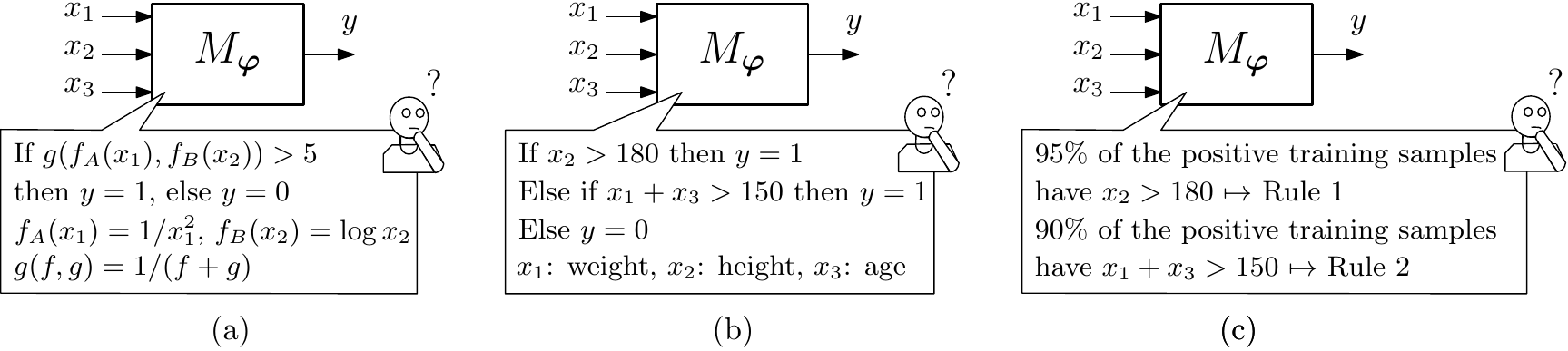}}
        \caption{Conceptual diagram exemplifying the different levels of transparency characterizing a ML model $M_{\bm{\varphi}}$, with $\bm{\varphi}$ denoting the parameter set of the model at hand: (a) simulatability; (b) decomposability; (c) algorithmic transparency. {\color{black}Without loss of generality, the example focuses on the ML model as the explanation target. However, other targets for explainability may include a given example, the output classes or the dataset itself.}}
        \label{fig:transparentML}
      \end{figure}

Transparent models convey some degree of interpretability by themselves. Models belonging to this category can be also approached in terms of the domain in which they are interpretable, namely, algorithmic transparency, decomposability and simulatability. As we elaborate next in connection to Figure \ref{fig:transparentML}, each of these classes contains its predecessors, e.g. a  \textit{simulatable} model is at the same time a model that is decomposable and algorithmically transparent:
\begin{itemize}[leftmargin=*]
\item \textit{Simulatability} denotes the ability of a model of being simulated or thought about strictly by a human, hence complexity takes a dominant place in this class. This being said, simple but extensive (i.e., with \textit{too large} amount of rules) rule based systems fall out of this characteristic, whereas a single perceptron neural network falls within. This aspect aligns with the claim that sparse linear models are more interpretable than dense ones \cite{tibshirani1996Simulatability}, and that an interpretable model is one that can be easily presented to a human by means of text and \textit{visualizations} \cite{ribeiro2016trust}. Again, endowing a decomposable model with simulatability requires that the model has to be self-contained enough for a human to think and reason about it as a whole.

\item \textit{Decomposability} stands for the ability to explain each of the parts of a model (input, parameter and calculation). It can be considered as intelligibility as stated in \cite{lou2012Decomposability}. This characteristic might empower the ability to understand, interpret or explain the behavior of a model. However, as occurs with algorithmic transparency, not every model can fulfill this property. Decomposability requires every input to be readily interpretable (e.g. cumbersome features will not fit the premise). The added constraint for an algorithmically transparent model to become decomposable is that every part of the model must be understandable by a human without the need for additional tools.

\item \textit{Algorithmic Transparency} can be seen in different ways. It deals with the ability of the user to understand the process followed by the model to produce any given output from its input data. Put it differently, a linear model is deemed transparent because its error surface can be understood and reasoned about, allowing the user to understand how the model will act in every situation it may face \cite{james2013Transferability}. Contrarily, it is not possible to understand it in deep architectures as the loss landscape might be opaque \cite{kawaguchi2016Transparency,AlgorithmicTransparency} since it cannot be fully observed and the solution has to be approximated through heuristic optimization (e.g. through stochastic gradient descent). The main constraint for algorithmically transparent models is that the model has to be fully explorable by means of mathematical analysis and methods.
\end{itemize}

\subsubsection{Post-hoc Explainability Techniques for Machine Learning Models}

Post-hoc explainability targets models that are not readily interpretable by design by resorting to diverse means to enhance their interpretability, such as \textit{text explanations}, \textit{visual explanations}, \textit{local explanations}, \textit{explanations by example}, \textit{explanations by simplification} and \textit{feature relevance explanations} techniques. Each of these techniques covers one of the most common ways humans explain systems and processes by themselves.  

Further along this river, actual techniques, or better put, actual group of techniques are specified to ease the future work of any researcher that intends to look up for an specific technique that suits its knowledge. Not ending there, the classification also includes the type of data in which the techniques has been applied. Note that many techniques might be suitable for many different types of data, although the categorization only considers the type used by the authors that proposed such technique. Overall, post-hoc explainability techniques are divided first by the intention of the author (explanation technique e.g. Explanation by simplification), then, by the method utilized (actual technique e.g. sensitivity analysis) and finally by the type of data in which it was applied (e.g. images).

\begin{itemize}[leftmargin=*]
\item \textit{Text explanations} deal with the problem of bringing explainability for a model by means of learning to generate \textit{text explanations} that help explaining the results from the model \cite{Bennetot19}. \textit{Text explanations} also include every method generating symbols that represent the functioning of the model. These symbols may portrait the rationale of the algorithm by means of a semantic mapping from model to symbols.

\item \textit{Visual explanation} techniques for post-hoc explainability aim at visualizing the model's behavior. Many of the visualization methods existing in the literature come along with dimensionality reduction techniques that allow for a human interpretable simple visualization. Visualizations may be coupled with other techniques to improve their understanding, and are considered as the most suitable way to introduce complex interactions within the variables involved in the model to users not acquainted to ML modeling.

\item \textit{Local explanations} tackle explainability by segmenting the solution space and giving explanations to less complex solution subspaces that are relevant for the whole model. These explanations can be formed by means of techniques with the differentiating property that these only explain part of the whole system's functioning. 
    
\item \textit{Explanations by example} consider the extraction of data examples that relate to the result generated by a certain model, enabling to get a better understanding of the model itself. Similarly to how humans behave when attempting to explain a given process, \textit{explanations by example} are mainly centered in extracting representative examples that grasp the inner relationships and correlations found by the model being analyzed.
    
\item \textit{Explanations by simplification} collectively denote those techniques in which a whole new system is rebuilt based on the trained model to be explained. This new, simplified model usually attempts at optimizing its resemblance to its antecedent functioning, while reducing its complexity, and keeping a similar performance score. An interesting byproduct of this family of post-hoc techniques is that the simplified model is, in general, easier to be implemented due to its reduced complexity with respect to the model it represents.
    
\item Finally, \textit{feature relevance explanation} methods for post-hoc explainability clarify the inner functioning of a model by computing a relevance score for its managed variables. These scores quantify the affection (sensitivity) a feature has upon the output of the model. A comparison of the scores among different variables unveils the importance granted by the model to each of such variables when producing its output. \textit{Feature relevance} methods can be thought to be an indirect method to explain a model.
\end{itemize}
\begin{figure}[ht]
\center{\includegraphics[width=0.8\columnwidth]{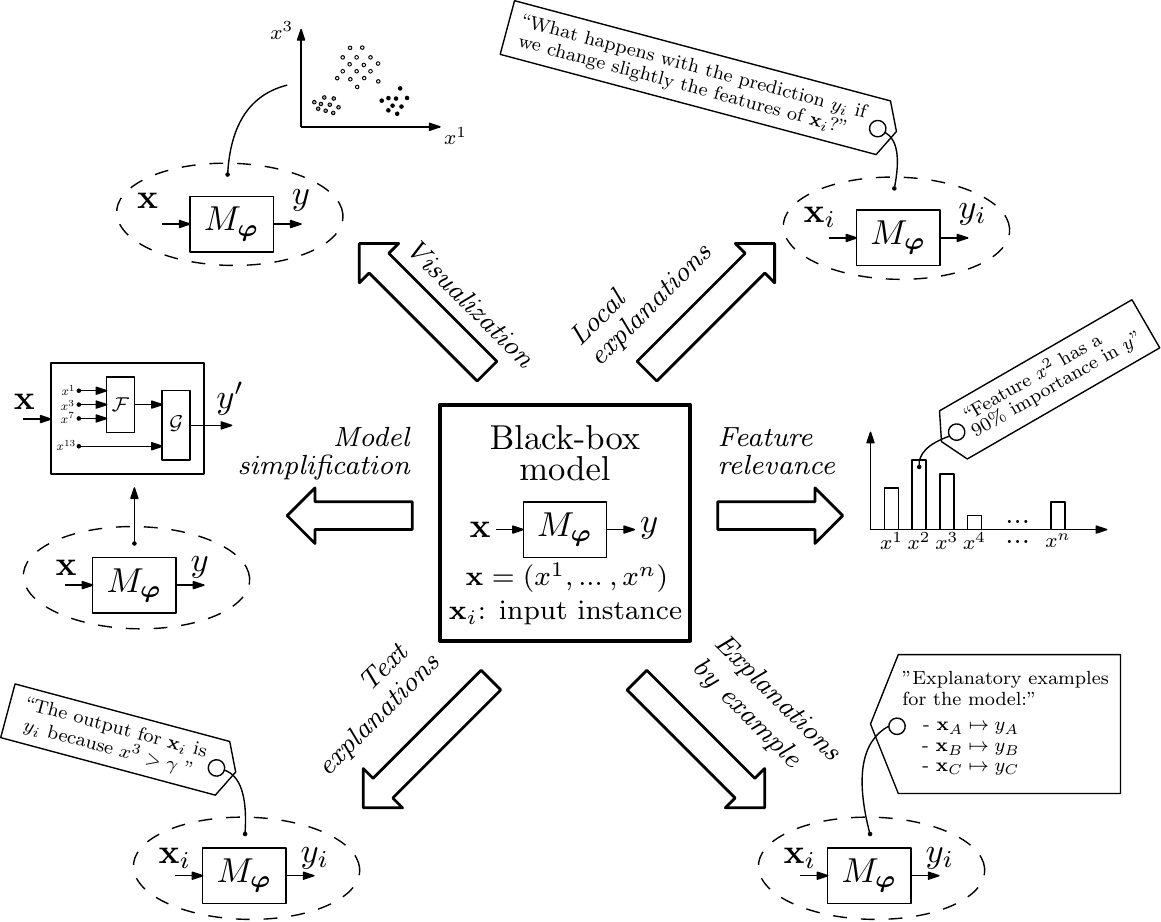}}
\caption{Conceptual diagram showing the different post-hoc explainability approaches available for a ML model $M_{\bm{\varphi}}$.}
\label{fig:post-hoc}
\end{figure}

The above classification (portrayed graphically in Figure \ref{fig:post-hoc}) will be used when reviewing specific/agnostic XAI techniques for ML models in the following sections (Table \ref{tab:ModelTaxonomy}). For each ML model, a distinction of the propositions to each of these categories is presented in order to pose an overall image of the field's trends. 
\begin{table}[h]
	\centering
	\resizebox{\textwidth}{!}{%
		\begin{tabular}{cC{6cm}C{6cm}C{6cm}C{6cm}c}
			\specialrule{.2em}{.1em}{.1em}
			\multirow{3}{*}{\textbf{Model}} & \multicolumn{3}{c}{\textbf{Transparent ML Models}} & \multirow{3}{1.5cm}{\textbf{Post-hoc analysis}} \\ \cmidrule(r){2-4}
			& \textbf{Simulatability} & \textbf{Decomposability} & \textbf{Algorithmic Transparency} & \\ \specialrule{.2em}{.1em}{.1em}
			Linear/Logistic Regression & Predictors are human readable and interactions among them are kept to a minimum & Variables are still readable, but the number of interactions and predictors involved in them have grown to force decomposition & Variables and interactions are too complex to be analyzed without mathematical tools & Not needed \\
			\hline
			Decision Trees & A human can simulate and obtain the prediction of a decision tree on his/her own, without requiring any mathematical background & The model comprises rules that do not alter data whatsoever, and preserves their readability & Human-readable rules that explain the knowledge learned from data and allows for a direct understanding of the prediction process & Not needed \\
            \hline
            K-Nearest Neighbors & The complexity of the model (number of variables, their understandability and the similarity measure under use) matches human naive capabilities for simulation & The amount of variables is too high and/or the similarity measure is too complex to be able to simulate the model completely, but the similarity measure and the set of variables can be decomposed and analyzed separately & The similarity measure cannot be decomposed and/or the number of variables is so high that the user has to rely on mathematical and statistical tools to analyze the model & Not needed\\
			\hline
			Rule Based Learners & Variables included in rules are readable, and the size of the rule set is manageable by a human user without external help & The size of the rule set becomes too large to be analyzed without decomposing it into small rule chunks & Rules have become so complicated (and the rule set size has grown so much) that mathematical tools are needed for inspecting the model behaviour & Not needed\\
			\hline
			General Additive Models & Variables and the interaction among them as per the smooth functions involved in the model must be constrained within human capabilities for understanding & Interactions become too complex to be simulated, so decomposition techniques are required for analyzing the model & Due to their complexity, variables and interactions cannot be analyzed without the application of mathematical and statistical tools & Not needed\\
			\hline
			Bayesian Models & Statistical relationships modeled among variables and the variables themselves should be directly understandable by the target audience & Statistical relationships involve so many variables that they must be decomposed in marginals so as to ease their analysis & Statistical relationships cannot be interpreted even if already decomposed, and predictors are so complex that model can be only analyzed with mathematical tools & Not needed\\
			\hline
			Tree Ensembles & \xmark & \xmark & \xmark & Needed: Usually \textit{Model simplification} or \textit{Feature relevance} techniques \\
			\hline
			Support Vector Machines & \xmark & \xmark & \xmark & Needed: Usually \textit{Model simplification} or \textit{Local explanations} techniques\\
			\hline
			Multi--layer Neural Network & \xmark & \xmark & \xmark & Needed: Usually \textit{Model simplification}, \textit{Feature relevance} or \textit{Visualization} techniques\\
			\hline
			Convolutional Neural Network & \xmark & \xmark & \xmark & Needed: Usually \textit{Feature relevance} or \textit{Visualization} techniques \\
			\hline
			Recurrent Neural Network & \xmark & \xmark & \xmark & Needed: Usually \textit{Feature relevance} techniques \\
			\specialrule{.2em}{.1em}{.1em}
		\end{tabular}%
	}
	\caption{Overall picture of the classification of ML models attending to their level of explainability.}
	\label{tab:ModelTaxonomy}
\end{table}

\section{Transparent Machine Learning Models} \label{sec:transparent}

\begin{figure}[h]
        \center{\includegraphics[width=\columnwidth]{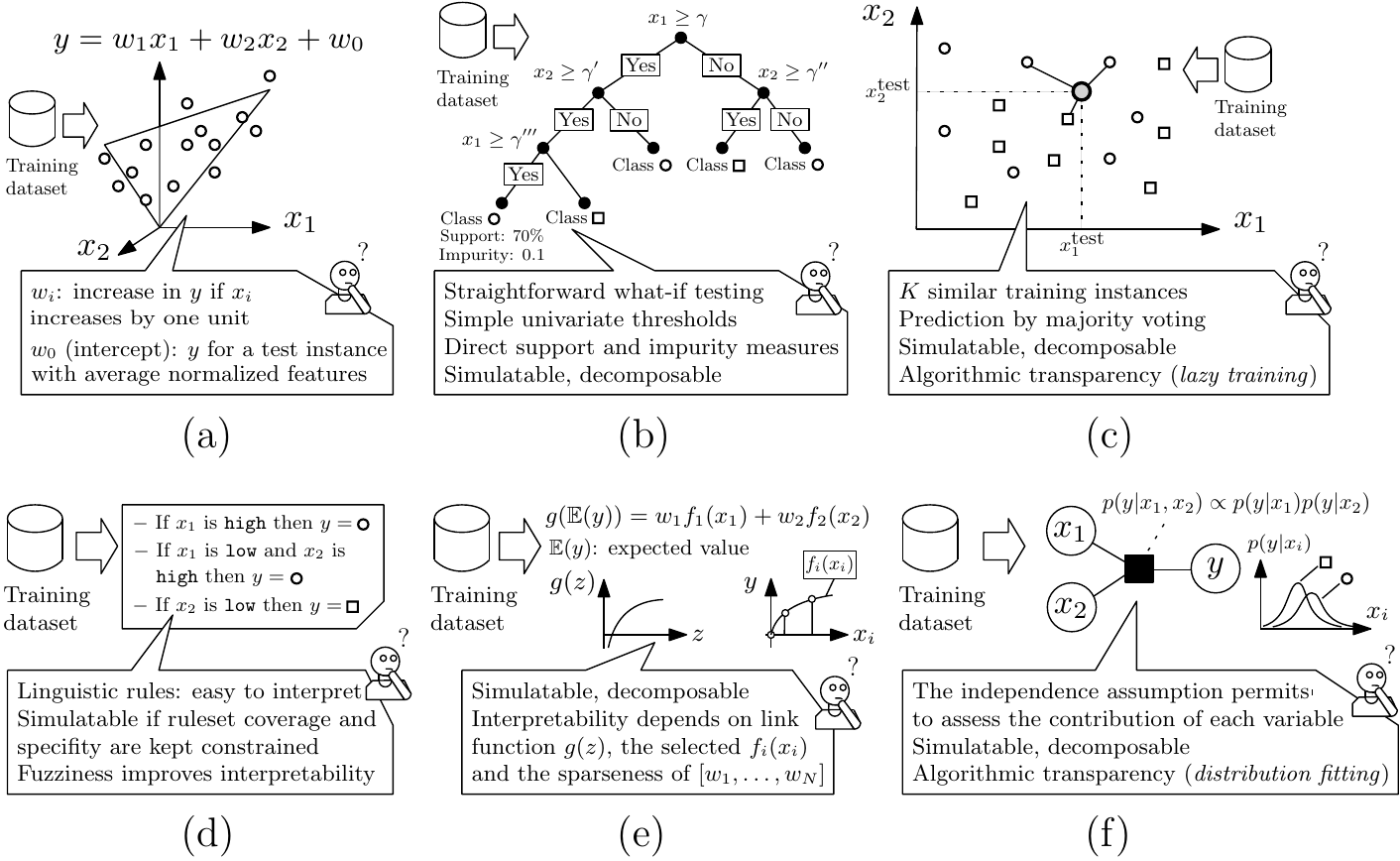}}
        \caption{Graphical illustration of the levels of transparency of different ML models considered in this overview: (a) Linear regression; (b) Decision trees; (c) K-Nearest Neighbors; (d) Rule-based Learners; (e) Generalized Additive Models; (f) Bayesian Models.}
        \label{fig:transparentShow}
      \end{figure}

The previous section introduced the concept of \textit{transparent} models. A model is considered to be transparent if by itself it is understandable. The models surveyed in this section are a suit of transparent models that can fall in one or all of the levels of model transparency described previously (namely, simulatability, decomposability and algorithmic transparency). In what follows we provide reasons for this statement, with graphical support given in Figure \ref{fig:transparentShow}.

\subsection{Linear/Logistic Regression}

Logistic Regression (LR) is a classification model to predict a dependent variable (category) that is dichotomous (binary). However, when the dependent variable is continuous, linear regression would be its homonym. This model takes the assumption of linear dependence between the predictors and the predicted variables, impeding a flexible fit to the data. This specific reason (stiffness of the model) is the one that maintains the model under the umbrella of transparent methods. However, as stated in Section 2, explainability is linked to a certain audience, which makes a model fall under both categories depending who is to interpret it. This way, logistic and linear regression, although clearly meeting the characteristics of transparent models (algorithmic transparency, decomposability and simulatability), may also demand post-hoc explainability techniques (mainly, visualization), particularly when the model is to be explained to non-expert audiences.

The usage of this model has been largely applied within Social Sciences for quite a long time, which has pushed researchers to create ways of explaining the results of the models to non-expert users. Most authors agree on the different techniques used to analyze and express the soundness of LR \cite{PurposeLR,InteractionLR,AppliedLR,IntroLR}, including the overall model evaluation, statistical tests of individual predictors, goodness-of-fit statistics and validation of the predicted probabilities. The overall model evaluation shows the improvement of the applied model over a baseline, showing if it is in fact improving the model without predictions. The statistical significance of single predictors is shown by calculating the Wald chi-square statistic. The goodness-of-fit statistics show the quality of fitness of the model to the data and how significant this is. This can be achieved by resorting to different techniques e.g. the so-called Hosmer-Lemeshow (H-L) statistic. The validation of predicted probabilities involves testing whether the output of the model corresponds to what is shown by the data. These techniques show mathematical ways of representing the fitness of the model and its behavior. 

Other techniques from other disciplines besides Statistics can be adopted for explaining these regression models. Visualization techniques are very powerful when presenting statistical conclusions to users not well-versed in statistics. For instance, the work in \cite{NaturalFrecuencies} shows that the usage of probabilities to communicate the results, implied that the users where able to estimate the outcomes correctly in 10\% of the cases, as opposed to 46\% of the cases when using natural frequencies. Although logistic regression is among the simplest classification models in supervised learning, there are concepts that must be taken care of.

In this line of reasoning, the authors of \cite{CannotDo} unveil some concerns with the interpretations derived from LR. They first mention how dangerous it might be to interpret log odds ratios and odd ratios as substantive effects, since they also represent unobserved heterogeneity. Linked to this first concern, \cite{CannotDo} also states that a comparison between these ratios across models with different variables might be problematic, since the unobserved heterogeneity is likely to vary, thereby invalidating the comparison. Finally they also mention that the comparison of these odds across different samples, groups and time is also risky, since the variation of the heterogeneity is not known across samples, groups and time points. This last paper serves the purpose of visualizing the problems a model's interpretation might entail, even when its construction is as simple as that of LR.

Also interesting is to note that, for a model such as logistic or linear regression to maintain decomposability and simulatability, its size must be limited, and the variables used must be understandable by their users. As stated in Section 2, if inputs to the model are highly engineered features that are complex or difficult to understand, the model at hand will be far from being \textit{decomposable}. Similarly, if the model is so large that a human cannot think of the model as a whole, its simulatability will be put to question. 

\subsection{Decision Trees}

Decision trees are another example of a model that can easily fulfill every constraint for transparency. Decision trees are hierarchical structures for decision making used to support regression and classification problems \cite{quinlan1987simplifying,laurent1976constructing}. In the simplest of their flavors, decision trees are \textit{simulatable} models. However, their properties can render them \textit{decomposable} or \textit{algorithmically transparent}. 

Decision trees have always lingered in between the different categories of transparent models. Their utilization has been closely linked to decision making contexts, being the reason why their complexity and understandability have always been considered a paramount matter. A proof of this relevance can be found in the upsurge of contributions to the literature dealing with decision tree simplification and generation \cite{quinlan1987simplifying,laurent1976constructing,utgoff1989incremental,quinlan1986induction}. As noted above, although being capable of fitting every category within transparent models, the individual characteristics of decision trees can push them toward the category of algorithmically transparent models. A \textit{simulatable} decision tree is one that is manageable by a human user. This means its size is somewhat small and the amount of features and their meaning are easily understandable. An increment in size transforms the model into a \textit{decomposable} one since its size impedes its full evaluation (simulation) by a human. Finally, further increasing its size and using complex feature relations will make the model \textit{algorithmically transparent} loosing the previous characteristics.

Decision trees have long been used in decision support contexts due to their off-the-shelf transparency. Many applications of these models fall out of the fields of computation and AI (even information technologies), meaning that experts from other fields usually feel comfortable interpreting the outputs of these models \cite{rokach2008data,rovnyak1994decision,nefeslioglu2010assessment}. However, their poor generalization properties in comparison with other models make this model family less interesting for their application to scenarios where a balance between predictive performance is a design driver of utmost importance. Tree ensembles aim at overcoming such a poor performance by aggregating the predictions performed by trees learned on different subsets of training data. Unfortunately, the combination of decision trees looses every transparent property, calling for the adoption of post-hoc explainability techniques as the ones reviewed later in the manuscript.

\subsection{K-Nearest Neighbors}

Another method that falls within transparent models is that of K-Nearest Neighbors (KNN), which deals with classification problems in a methodologically simple way: it predicts the class of a test sample by voting the classes of its K nearest neighbors (where the neighborhood relation is induced by a measure of distance between samples). When used in the context of regression problems, the voting is replaced by an aggregation (e.g. average) of the target values associated with the nearest neighbors. 

In terms of model explainability, it is important to observe that predictions generated by KNN models rely on the notion of distance and similarity between examples, which can be tailored depending on the specific problem being tackled. Interestingly, this prediction approach resembles that of experience-based human decision making, which decides upon the result of past similar cases. There lies the rationale of why KNN has also been adopted widely in contexts in which model interpretability is a requirement \cite{KNNimandoust2013application,KNNli2004application,KNNguo2004knn,KNNjiang2012improved}. Furthermore, aside from being simple to explain, the ability to inspect the reasons by which a new sample has been classified inside a group and to examine how these predictions evolve when the number of neighbors K is increased or decreased empowers the interaction between the users and the model.

One must keep in mind that as mentioned before, KNN's class of transparency depends on the features, the number of neighbors and the distance function used to measure the similarity between data instances. A very high K impedes a full simulation of the model performance by a human user. Similarly, the usage of complex features and/or distance functions would hinder the decomposability of the model, restricting its interpretability solely to the transparency of its algorithmic operations.

\subsection{Rule-based Learning} \label{ref:rbl}

Rule-based learning refers to every model that generates rules to characterize the data it is intended to learn from. Rules can take the form of simple conditional \textit{if-then} rules or more complex combinations of simple rules to form their knowledge. Also connected to this general family of models, fuzzy rule based systems are designed for a broader scope of action, allowing for the definition of verbally formulated rules over imprecise domains. Fuzzy systems improve two main axis relevant for this paper. First, they empower more understandable models since they operate in linguistic terms. Second, they perform better that classic rule systems in contexts with certain degrees of uncertainty. Rule based learners are clearly transparent models that have been often used to explain complex models by generating rules that explain their predictions \cite{nunez2002rule,nunez2006rule,RuleExtractionInThere,   ProductionRulesFromTrees}. 

Rule learning approaches have been extensively used for knowledge representation in expert systems \cite{RULElangley1995applications}. However, a central problem with rule generation approaches is the coverage (amount) and the specificity (length) of the rules generated. This problem relates directly to the intention for their use in the first place. When building a rule database, a typical design goal sought by the user is to be able to analyze and understand the model. The amount of rules in a model will clearly improve the performance of the model at the stake of compromising its intepretability. Similarly, the specificity of the rules plays also against interpretability, since a rule with a high number of antecedents an/or consequences might become difficult to interpret. In this same line of reasoning, these two features of a rule based learner play along with the classes of transparent models presented in Section 2. The greater the coverage or the specificity is, the closer the model will be to being just \textit{algorithmically transparent}. Sometimes, the reason to transition from classical rules to fuzzy rules is to relax the constraints of rule sizes, since a greater range can be covered with less stress on interpretability. 

Rule based learners are great models in terms of interpretability across fields. Their natural and seamless relation to human behaviour makes them very suitable to understand and explain other models. If a certain threshold of coverage is acquired, a rule wrapper can be thought to contain enough information about a model to explain its behavior to a non-expert user, without forfeiting the possibility of using the generated rules as an standalone prediction model. 

\subsection{General Additive Models}

In statistics, a Generalized Additive Model (GAM) is a linear model in which the value of the variable to be predicted is given by the aggregation of a number of unknown smooth functions defined for the predictor variables. The purpose of such model is to infer the smooth functions whose aggregate composition approximates the predicted variable. This structure is easily interpretable, since it allows the user to verify the importance of each variable, namely, how it affects (through its corresponding function) the predicted output. 

Similarly to every other transparent model, the literature is replete with case studies where GAMs are in use, specially in fields related to risk assessment. When compared to other models, these are understandable enough to make users feel confident on using them for practical applications in finance \cite{Bankruptcy,BankLoanLoss,FianceScienceTechnology}, environmental studies \cite{RelationshipsEnviromental}, geology \cite{GeositeAssesment}, healthcare \cite{caruana2015Transferability}, biology \cite{SpeciesDistribution,ButterflyTranscent} and energy \cite{ElectricityLoad}. Most of these contributions use visualization methods to further ease the interpretation of the model. GAMs might be also considered as \textit{simulatable} and \textit{decomposable} models if the properties mentioned in its definitions are fulfilled, but to an extent that depends roughly on eventual modifications to the baseline GAM model, such as the introduction of link functions to relate the aggregation with the predicted output, or the consideration of interactions between predictors.

All in all, applications of GAMs like the ones exemplified above share one common factor: understandability. The main driver for conducting these studies with GAMs is to understand the underlying relationships that build up the cases for scrutiny. In those cases the research goal is not accuracy for its own sake, but rather the need for understanding the problem behind and the relationship underneath the variables involved in data. This is why GAMs have been accepted in certain communities as their \emph{de facto} modeling choice, despite their acknowledged misperforming behavior when compared to more complex counterparts.

\subsection{Bayesian Models}

A Bayesian model usually takes the form of a probabilistic directed acyclic graphical model whose links represent the conditional dependencies between a set of variables. For example, a Bayesian network could represent the probabilistic relationships between diseases and symptoms. Given symptoms, the network can be used to compute the probabilities of the presence of various diseases. Similar to GAMs, these models also convey a clear representation of the relationships between features and the target, which in this case are  given explicitly by the connections linking variables to each other.

Once again, Bayesian models fall below the ceiling of Transparent models. Its categorization leaves it under \textit{simulatable}, \textit{decomposable} and \textit{algorithmically transparent}. However, it is worth noting that under certain circumstances (overly complex or cumbersome variables), a model may loose these first two properties. Bayesian models have been shown to lead to great insights in assorted applications such as cognitive modeling \cite{BayesianCognitive,BayesianPsychiatric}, fishery \cite{RelationshipsEnviromental,BayesianStock}, gaming \cite{BayesianRTS}, climate \cite{BayesianClimate}, econometrics \cite{BayesianEconometrics} or robotics \cite{BayesianRobot}. Furthermore, they have also been utilized to explain other models, such as averaging tree ensembles \cite{BayesianTree}.

\section{Post-hoc Explainability Techniques for Machile Learning Models: Taxonomy, Shallow Models and Deep Learning} \label{sec:posthoc}

When ML models do not meet any of the criteria imposed to declare them transparent, a separate method must be devised and applied to the model to explain its decisions. This is the purpose of post-hoc explainability techniques (also referred to as post-modeling explainability), which aim at communicating understandable information about how an already developed model produces its predictions for any given input. In this section we categorize and review different algorithmic approaches for post-hoc explainability, discriminating among 1) those that are designed for their application to ML models of any kind; and 2) those that are designed for a specific ML model and thus, can not be directly extrapolated to any other learner. We now elaborate on the trends identified around post-hoc explainability for different ML models, which are illustrated in Figure \ref{fig:treeCat} in the form of hierarchical bibliographic categories and summarized next:
\begin{itemize}[leftmargin=*]
\item Model-agnostic techniques for post-hoc explainability (Subsection \ref{sec:model-agnostic}), which can be applied seamlessly to any ML model disregarding its inner processing or internal representations.
\item Post-hoc explainability that are tailored or specifically designed to explain certain ML models. We divide our literature analysis into two main branches: contributions dealing with post-hoc explainability of \emph{shallow} ML models, which collectively refers to all ML models that do not hinge on layered structures of neural processing units (Subsection \ref{ssec:shallow-posthoc}); and techniques devised for \emph{deep} learning models, which correspondingly denote the family of neural networks and related variants, such as convolutional neural networks, recurrent neural networks (Subsection \ref{ssec:deep-posthoc}) and hybrid schemes encompassing deep neural networks and transparent models. For each model we perform a thorough review of the latest post-hoc methods proposed by the research community, along with a identification of trends followed by such contributions.
\item We end our literature analysis with Subsection \ref{ssec:second_tax}, where we present a second taxonomy that complements the more general one in Figure \ref{fig:treeCat} by classifying contributions dealing with the post-hoc explanation of Deep Learning models. To this end we focus on particular aspects related to this family of black-box ML methods, and expose how they link to the classification criteria used in the first taxonomy. 
\end{itemize}
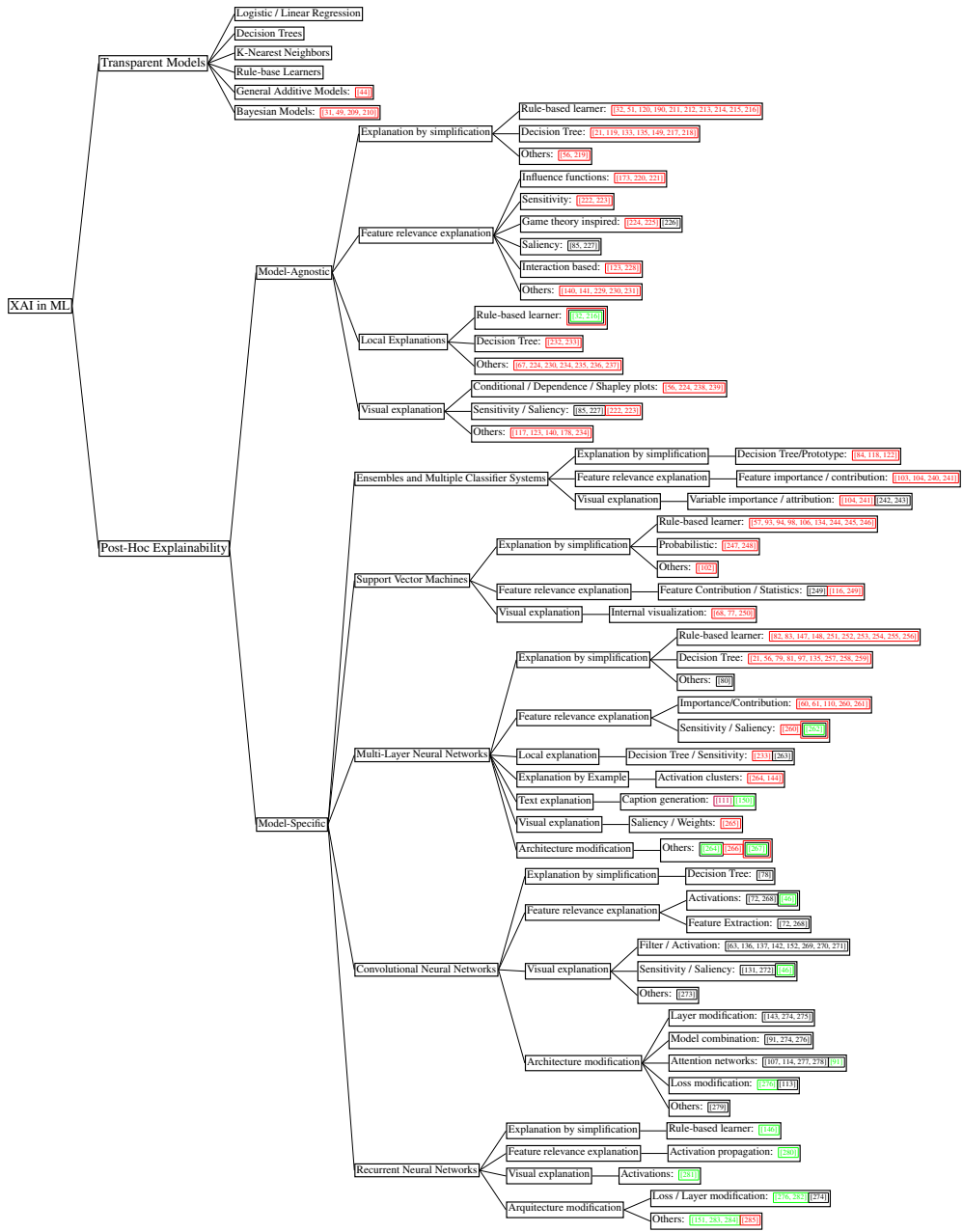
\begin{figure}[h!]
	\resizebox{0.9\columnwidth}{!}{\begin{forest}
			for tree={
				l sep=5em, s sep=1em,
				child anchor=west,
				parent anchor=east,
				grow'=0,
				line width=0.75mm,
				anchor=west,
				draw,
			}
			[{{\Huge XAI in ML}}
			[{{\Huge Transparent Models}}            
    			[{{\huge Logistic / Linear Regression}} \\                             
    			]              
    			[{{\huge Decision Trees}} \\
    			]
    			[{{\huge K-Nearest Neighbors}} \\
    			]
    			[{{\huge Rule-base Learners}} \\
    			]
    			[{{\huge General Additive Models: }} 
    			    {{\Large\color{red}\framebox{\cite{caruana2015Transferability}}}}
    			]
    			[{{\huge Bayesian Models: }}
    			    {{\Large\color{red}\framebox{\cite{kim2015Trust,letham2015interpretable,kim2014bayesian,kim2016examples}}}}
    			]
			]
			[{{\Huge Post-Hoc Explainability}}        
    			[{{\huge Model-Agnostic}} \\ 
        			[{{\huge Explanation by simplification}\vspace{1.5mm}} \\
            			[{{\huge Rule-based learner: }\vspace{1.5mm}}
                            {{\Large\color{red}\framebox{\cite{ribeiro2016trust,aung2007comparing,DistillAndCompare,RuleExtractionInThere,AccVsComp,GREX,lakkaraju2017interpretable,ModelAgnosticMusic,InterpretableTwoLevel,NothingElseMatters}}}}
            			]
            			[{{\huge Decision Tree: }\vspace{1.5mm}}
                            {{\Large\color{red}\framebox{\cite{craven1996extracting,domingos1998knowledge,SingleTreeApproximation,TreeView,johansson2009evolving,Craven96,InterpretabilityViaModelExtraction}}}}
            			]
            			[{{\huge Others: }\vspace{1.5mm}}
                            {{\Large\color{red}\framebox{\cite{InterpretableDeepICU,DiscoveringAdditive}}}}
            			]
            		]
        			[{{\huge Feature relevance explanation}\vspace{1.5mm}} \\
                        [{{\huge Influence functions: }\vspace{1.5mm}}
                            {{\Large\color{red}\framebox{\cite{AlgorithmicTransparency,adler2018auditing,ViaInfluence}}}}
            			]
            			[{{\huge Sensitivity: }\vspace{1.5mm}}
                            {{\Large\color{red}\framebox{\cite{SensitivityAnalysis,UsingSensitivityAndVisualization}}}}
            			]
            			[{{\huge Game theory inspired: }\vspace{1.5mm}}
                            {{\Large\color{red}\framebox{\cite{lundberg2017unified,EfficientExplanation}}}{\Large\color{black}\framebox{\cite{1911.11888}}}}
            			]
            			[{{\huge Saliency: }\vspace{1.5mm}}
                            {{\Large\color{black}\framebox{\cite{fong2017interpretable,RealTimeImageSaliency}}}}
            			]
            			[{{\huge Interaction based: }\vspace{1.5mm}}
                            {{\Large\color{red}\framebox{\cite{AtributeInteractions,ExploringByRandomization}}}}
            			]
            			[{{\huge Others: }\vspace{1.5mm}}
                            {{\Large\color{red}\framebox{\cite{apley2016visualizing,staniak2018explanations,moeyersoms2016explaining,IndividualClassificationDecisions,adebayo2016iterative}}}}
            			]
            		]
            		[{{\huge Local Explanations}\vspace{1.5mm}} \\
            			[{{\huge Rule-based learner: }\vspace{1.5mm}}
                            {{\Large\color{red}\framebox{\color{black}\framebox{\color{green}\framebox{\cite{ribeiro2016trust,NothingElseMatters}}}}}}
            			]
            			[{{\huge Decision Tree: }\vspace{1.5mm}}
                            {{\Large\color{red}\framebox{\cite{guidotti2018local,krishnan2017palm}}}}
            			]
            			[{{\huge Others: }\vspace{1.5mm}}
                            {{\Large\color{red}\framebox{\cite{krause2016interacting,lundberg2017unified,IndividualClassificationDecisions,ExplainingClassifications,ribeiro2018anchors,Martens:2014,chen2017enhancing}}}}
            			]
            		]
        			[{{\huge Visual explanation}\vspace{1.5mm}} \\
                        [{{\huge Conditional / Dependence / Shapley plots: }\vspace{1.5mm}}
                            {{\Large\color{red}\framebox{\cite{InterpretableDeepICU,lundberg2017unified,VisualizingStatisticalLearning,VisualizingFeatureImportance}}}}
            			]
            			[{{\huge Sensitivity / Saliency: }\vspace{1.5mm}}
                            {{\Large\color{black}\framebox{\cite{fong2017interpretable,RealTimeImageSaliency}}\Large\color{red}\framebox{\cite{SensitivityAnalysis,UsingSensitivityAndVisualization}}}}
            			]
            			[{{\huge Others: }\vspace{1.5mm}}
                            {{\Large\color{red}\framebox{\cite{xu2018interpreting,AtributeInteractions,apley2016visualizing,NaturalFrecuencies,ExplainingClassifications}}}}
            			]
        			]
    			] 
    			[{{\huge Model-Specific}} \\
        			[{{\huge Ensembles and Multiple Classifier Systems}} \\
            			[{{\huge Explanation by simplification}\vspace{1.5mm}} \\  
                            [{{\huge Decision Tree/Prototype: }\vspace{1.5mm}}
                                {{\Large\color{red}\framebox{\cite{tan2016tree,Intrees,MakingTEInterpretable}}}}
                			]
        			    ]
            			[{{\huge Feature relevance explanation}\vspace{1.5mm}} \\        
                			[{{\huge Feature importance / contribution: }\vspace{1.5mm}}
                                {{\Large\color{red}\framebox{\cite{FeatureContributionMethod,welling2016forest,FeatureTweaking,auret2012interpretation}}}}
                			]
            			]
            			[{{\huge Visual explanation}\vspace{1.5mm}} \\        
                			[{{\huge Variable importance / attribution: }\vspace{1.5mm}}
                                {{\Large\color{red}\framebox{\cite{welling2016forest,auret2012interpretation}}}{\Large\color{black}\framebox{\cite{rajani2018stacking,rajani2018ensembling}}}}
                			]    
            			]
    			    ]
    			    [{{\huge Support Vector Machines}} \\
        			    [{{\huge Explanation by simplification}\vspace{1.5mm}}\\
                            [{{\huge Rule-based learner: }\vspace{1.5mm}}
                                {{\Large\color{red}\framebox{\cite{barakat2008,barakat2007rule,Chaves2005,fu2004,zhang2005,intepretationSVM,nunez2006,chen2007,nunez2002B}}}}
                			]
                			[{{\huge Probabilistic: }\vspace{1.5mm}}
                                {{\Large\color{red}\framebox{\cite{bayesianForSVM,probabilisticSVM}}}}
                			]  
                			[{{\huge Others: }\vspace{1.5mm}}
                                {{\Large\color{red}\framebox{\cite{haasdonk2005feature}}}}
                			] 
        			    ]
        			    [{{\huge Feature relevance explanation}\vspace{1.5mm}}\\ 
                			[{{\huge Feature Contribution / Statistics: }\vspace{1.5mm}}
                                {{\Large\color{black}\framebox{\cite{landecker2013interpreting}}\Large\color{red}\framebox{\cite{interpretingNeuroSVM,landecker2013interpreting}}}}
                			] 
        			    ]
        			    [{{\huge Visual explanation}\vspace{1.5mm}}\\      
                			[{{\huge Internal visualization: }\vspace{1.5mm}}
                                {{\Large\color{red}\framebox{\cite{interpretingHeatMapSVM,ustun2007visualisation,jakulin2005nomograms}}}}
                			]
        			    ]
        		    ] 
        			[{{\huge Multi-Layer Neural Networks}} \\
            			[{{\huge Explanation by simplification}\vspace{1.5mm}}\\  
                            [{{\huge Rule-based learner: }\vspace{1.5mm}}
                                {{\Large\color{red}\framebox{\cite{augasta2012reverse,zhou2003extracting,craven1994using,arbatli1997rule,fu1994rule,Towell93,Thrun94,Setiono00,Taha99,Tsukimoto00}}}}
                			]
                			[{{\huge Decision Tree: }\vspace{1.5mm}}
                                {{\Large\color{red}\framebox{\cite{craven1996extracting,InterpretableDeepICU,wu2018beyond,frosst2017distilling,krishnan1999extracting,TreeView,zilke2016deepred,Schmitz99,CRED}}}}
                			]
                			[{{\huge Others: }\vspace{1.5mm}}
                                {{\Large\color{black}\framebox{\cite{hinton2015distilling}}}}
                			]
            			]
            			[{{\huge Feature relevance explanation}\vspace{1.5mm}}\\  
                            [{{\huge Importance/Contribution: }\vspace{1.5mm}}
                                {{\Large\color{red}\framebox{\cite{DeepTaylor,LearningHowTo,shrikumar2016not,feraud2002methodology,Shrikumar17}}}}
                			]   
                			[{{\huge Sensitivity / Saliency: }\vspace{1.5mm}}
                                {{\Large\color{red}\framebox{\cite{feraud2002methodology}}\Large\color{red}\framebox{\color{black}\framebox{\color{green}\framebox{\cite{Axiomatic}}}}}}
                			] 
            			]
                        [{{\huge Local explanation}\vspace{1.5mm}}\\  
                            [{{\huge Decision Tree / Sensitivity: }\vspace{1.5mm}}
                                {\Large\color{red}\framebox{\cite{krishnan2017palm}}{\Large\color{black}\framebox{\cite{adebayo2018local}}}}
                			]
            			]
            			[{{\huge Explanation by Example}\vspace{1.5mm}}\\  
                            [{{\huge Activation clusters: }\vspace{1.5mm}}
                                {{\Large\color{red}\framebox{\cite{papernot2018deep,kim2017interpretability}}}}
                			]
            			]
            			[{{\huge Text explanation}\vspace{1.5mm}}\\  
                            [{{\huge Caption generation: }\vspace{1.5mm}}
                                {{\Large\color{purple}\framebox{\cite{ImprovingInterpretability}}\Large\color{green}\framebox{\cite{lei2016rationalizing}}}}
                			]
            			]
            			[{{\huge Visual explanation}\vspace{1.5mm}}\\  
                            [{{\huge Saliency / Weights: }\vspace{1.5mm}}
                                {{\Large\color{red}\framebox{\cite{li2015visualizing}}}}
                			]
            			]
            			[{{\huge Architecture modification}\vspace{1.5mm}}\\ 
                			[{{\huge Others: }\vspace{1.5mm}}
                                {{\Large\color{black}\framebox{\color{green}\framebox{\cite{papernot2018deep}}}\Large\color{red}\framebox{\cite{ImprovingStimulation}}\Large\color{red}\framebox{\color{black}\framebox{\color{green}\framebox{\cite{1909.13584}}}}}}
                			]
            			]
        			] 
        			[{{\huge Convolutional Neural Networks}} \\
                        [{{\huge Explanation by simplification}\vspace{1.5mm}}\\ 
                			[{{\huge Decision Tree: }\vspace{1.5mm}}
                                {{\Large\color{black}\framebox{\cite{zhang2019interpreting}}}}
                			]
            			]
            			[{{\huge Feature relevance explanation}\vspace{1.5mm}}\\ 
                			[{{\huge Activations: }\vspace{1.5mm}}
                                {{\Large\color{black}\framebox{\cite{LayerWise,SynthesizingPreferredInputs}}\Large\color{black}\framebox{\color{green}\framebox{\cite{samek2017explainable}}}}}
                			]
                			[{{\huge Feature Extraction: }\vspace{1.5mm}}
                                {{\Large\color{black}\framebox{\cite{LayerWise,SynthesizingPreferredInputs}}}}
                			]
            			]
            			[{{\huge Visual explanation}\vspace{1.5mm}}\\  
                            [{{\huge Filter / Activation: }\vspace{1.5mm}}
                                {{\Large\color{black}\framebox{\cite{bach2016controlling,VisualizingUnderstanding,UnderstandingDeep,zeiler2010deconvolutional,selvaraju2016grad,li2016convergent,liu2016towards,goyal2016towards}}}}
                			]
            			    [{{\huge Sensitivity / Saliency: }\vspace{1.5mm}}
                                {{\Large\color{black}\framebox{\cite{zhang2015sensitivity,InsideConv}}\Large\color{black}\framebox{\color{green}\framebox{\cite{samek2017explainable}}}}}
                			]
                			[{{\huge Others: }\vspace{1.5mm}}
                                {{\Large\color{black}\framebox{\cite{nguyen2015deep}}}}
                			]
            		    ]
            			[{{\huge Architecture modification}\vspace{1.5mm}}\\ 
                            [{{\huge Layer modification: }\vspace{1.5mm}}
                                {{\Large\color{black}\framebox{\cite{springenberg2014striving,donahue2015long,NetworkInNetwork}}}}
                			]
                			[{{\huge Model combination: }\vspace{1.5mm}}
                                {{\Large\color{black}\framebox{\cite{xu2015show,donahue2015long,Hendricks16Generate}}}}
                			]
                			[{{\huge Attention networks: }\vspace{1.5mm}}
                                {{\Large\color{black}\framebox{\cite{linsley2018global,seo2017interpretable,wang2017residual,xiao2015application}}\Large\color{black}\framebox{\color{green}\cite{xu2015show}}}}
                			]
                			[{{\huge Loss modification: }\vspace{1.5mm}}
                                {{\Large\color{green}\framebox{\cite{Hendricks16Generate}}\Large\color{black}\framebox{\cite{InterpretableCNN}}}}
                			]
                			[{{\huge Others: }\vspace{1.5mm}}
                                {{\Large\color{black}\framebox{\cite{Zhang16}}}}
                			]
        		    	]
        			] 
        			[{{\huge Recurrent Neural Networks}} \\
            			[{{\huge Explanation by simplification}\vspace{1.5mm}}\\ 
                            [{{\huge Rule-based learner: }\vspace{1.5mm}}
                                {{\Large\color{green}\framebox{\cite{murdoch2017automatic}}}}
                			]
            			]
            			[{{\huge Feature relevance explanation}\vspace{1.5mm}}\\ 
                            [{{\huge Activation propagation: }\vspace{1.5mm}}
                                {{\Large\color{green}\framebox{\cite{ExplainingRNN}}}}
                			]
            			]
            			[{{\huge Visual explanation}\vspace{1.5mm}}\\ 
                            [{{\huge Activations: }\vspace{1.5mm}}
                                {{\Large\color{green}\framebox{\cite{VisualizingUnderstandingRNN}}}}
                			]
            			]
            			[{{\huge Arquitecture modification}\vspace{1.5mm}}\\ 
                            [{{\huge Loss / Layer modification: }\vspace{1.5mm}}
                                {{\Large\color{green}\framebox{\cite{Hendricks16Generate,clos2017towards}}\Large\color{black}\framebox{\cite{donahue2015long}}}}
                		    ]
                		    [{{\huge Others: }\vspace{1.5mm}}
                                {{\Large\color{green}\framebox{\cite{radford2017learning,InterpretableRNN,MarkovRNN}}\Large\color{red}\framebox{\cite{RETAIN}}}}
                		    ]
            			]
        			] 
    			] 
			] 
		] 
	\end{forest}}
	\centering
	\caption{Taxonomy of the reviewed literature and trends identified for explainability techniques related to different ML models. References boxed in {\color{black}blue}, {\color{green}green} and {\color{red}red} correspond to XAI techniques using image, text or tabular data, respectively. In order to build this taxonomy, the literature has been analyzed in depth to discriminate whether a post-hoc technique can be seamlessly applied to any ML model, even if, e.g., explicitly mentions \textit{Deep Learning} in its title and/or abstract.}
	\label{fig:treeCat}
\end{figure}

\subsection{Model-agnostic Techniques for Post-hoc Explainability}
\label{sec:model-agnostic}

Model-agnostic techniques for post-hoc explainability are designed to be plugged to any model with the intent of extracting some information from its prediction procedure. Sometimes, simplification techniques are used to generate proxies that mimic their antecedents with the purpose of having something tractable and of reduced complexity. Other times, the intent focuses on extracting knowledge directly from the models or simply visualizing them to ease the interpretation of their behavior. Following the taxonomy introduced in Section 2, model-agnostic techniques may rely on \textit{model simplification}, \textit{feature relevance} estimation and \textit{visualization} techniques:
\begin{itemize}[leftmargin=*]
	\item \textit{Explanation by simplification}. They are arguably the broadest technique under the category of model agnostic post-hoc methods.  \textit{Local explanations} are also present within this category, since sometimes, simplified models are only representative of certain sections of a model. Almost all techniques taking this path for \textit{model simplification} are based on rule extraction techniques. Among the most known contributions to this approach we encounter the technique of Local Interpretable Model-Agnostic Explanations (LIME) \cite{ribeiro2016trust} and all its variations \cite{ModelAgnosticMusic,NothingElseMatters}. LIME builds locally linear models around the predictions of an opaque model to explain it. These contributions fall under explanations by simplification as well as under \textit{local explanations}. Besides LIME and related flavors, another approach to rule extraction is G-REX \cite{GREX}. Although it was not originally intended for extracting rules from opaque models, the generic proposition of G-REX has been extended to also account for model explainability purposes \cite{ RuleExtractionInThere,AccVsComp}. In line with rule extraction methods, the work in \cite{InterpretableTwoLevel} presents a novel approach to learn rules in CNF (Conjunctive Normal Form) or DNF (Disjunctive Normal Form) to bridge from a complex model to a human-interpretable model. Another contribution that falls off the same branch is that in \cite{InterpretabilityViaModelExtraction}, where the authors formulate \textit{model simplification} as a model extraction process by approximating a transparent model to the complex one. Simplification is approached from a different perspective in \cite{DistillAndCompare}, where an approach to distill and audit black box models is presented. In it, two main ideas are exposed: a method for model distillation and comparison to audit black-box risk scoring models; and an statistical test to check if the auditing data is missing key features it was trained with. 
	The popularity of \textit{model simplification} is evident, given it temporally coincides with the most recent literature on XAI, including techniques such as LIME or G-REX. This symptomatically reveals that this post-hoc explainability approach is envisaged to continue playing a central role on XAI.
	
	\item \textit{Feature relevance explanation} techniques aim to describe the functioning of an opaque model by ranking or measuring the influence, relevance or importance each feature has in the prediction output by the model to be explained. An amalgam of propositions are found within this category, each resorting to different algorithmic approaches with the same targeted goal. One fruitful contribution to this path is that of \cite{lundberg2017unified} called SHAP (SHapley Additive exPlanations). Its authors presented a method to calculate an additive feature importance score for each particular prediction with a set of desirable properties (local accuracy, \textit{missingness} and consistency) that its antecedents lacked. Another approach to tackle the contribution of each feature to predictions has been coalitional Game Theory \cite{EfficientExplanation} and local gradients \cite{ExplainingClassifications}. Similarly, by means of local gradients \cite{IndividualClassificationDecisions} test the changes needed in each feature to produce a change in the output of the model. In \cite{ExploringByRandomization} the authors analyze the relations and dependencies found in the model by grouping features, that combined, bring insights about the data. The work in \cite{AlgorithmicTransparency} presents a broad variety of measures to tackle the quantification of the degree of influence of inputs on outputs of systems. Their QII (Quantitative Input Influence) measures account for correlated inputs while measuring influence. In contrast, in \cite{SensitivityAnalysis} the authors build upon the existing SA (Sensitivity Analysis) to construct a Global SA which extends the applicability of the existing methods. In \cite{RealTimeImageSaliency} a real-time image saliency method is proposed, which is applicable to differentiable image classifiers. The study in \cite{AtributeInteractions} presents the so-called Automatic STRucture IDentification method (ASTRID) to inspect which attributes are exploited by a classifier to generate a prediction. This method finds the largest subset of features such that the accuracy of a classifier trained with this subset of features cannot be distinguished in terms of accuracy from a classifier built on the original feature set. In \cite{ViaInfluence} the authors use influence functions to trace a model's prediction back to the training data, by only requiring an oracle version of the model with access to gradients and Hessian-vector products. Heuristics for creating counterfactual examples by modifying the input of the model have been also found to contribute to its explainability \cite{Martens:2014,chen2017enhancing}. Compared to those attempting explanations by simplification, a similar amount of publications were found tackling explainability by means of \textit{feature relevance} techniques. Many of the contributions date from 2017 and some from 2018, implying that as with \textit{model simplification} techniques, \textit{feature relevance} has also become a vibrant subject study in the current XAI landscape.
	
	\item \textit{Visual explanation} techniques are a vehicle to achieve model-agnostic explanations. Representative works in this area can be found in \cite{SensitivityAnalysis}, which present a portfolio of visualization techniques to help in the explanation of a black-box ML model built upon the set of extended techniques mentioned earlier (Global SA). Another set of visualization techniques is presented in \cite{UsingSensitivityAndVisualization}. The authors present three novel SA methods (data based SA, Monte-Carlo SA, cluster-based SA) and one novel input importance measure (Average Absolute Deviation). Finally, \cite{VisualizingStatisticalLearning} presents ICE (Individual Conditional Expectation) plots as a tool for visualizing the model estimated by any supervised learning algorithm. Visual explanations are less common in the field of model-agnostic techniques for post-hoc explainability. Since the design of these methods must ensure that they can be seamlessly applied to any ML model disregarding its inner structure, creating \textit{visualizations} from just inputs and outputs from an opaque model is a complex task. This is why almost all visualization methods falling in this category work along with \textit{feature relevance} techniques, which provide the information that is eventually displayed to the end user.
\end{itemize}

Several trends emerge from our literature analysis. To begin with, rule extraction techniques prevail in model-agnostic contributions under the umbrella of post-hoc explainability. This could have been intuitively expected if we bear in mind the wide use of rule based learning as explainability wrappers anticipated in Section \ref{ref:rbl}, and the complexity imposed by not being able to \textit{get into} the model itself. Similarly, another large group of contributions deals with \textit{feature relevance}. Lately these techniques are gathering much attention by the community when dealing with DL models, with hybrid approaches that utilize particular aspects of this class of models and therefore, compromise the independence of the \textit{feature relevance} method on the model being explained. Finally, visualization techniques propose interesting ways for visualizing the output of \textit{feature relevance} techniques to ease the task of model's interpretation. By contrast, visualization techniques for other aspects of the trained model (e.g. its structure, operations, etc) are tightly linked to the specific model to be explained.

\subsection{Post-hoc Explainability in Shallow ML Models} \label{ssec:shallow-posthoc}

Shallow ML covers a diversity of supervised learning models. Within these models, there are strictly interpretable (transparent) approaches (e.g. KNN and Decision Trees, already discussed in Section \ref{sec:transparent}). However, other shallow ML models rely on more sophisticated learning algorithms that require additional layers of explanation. Given their prominence and notable performance in predictive tasks, this section concentrates on two popular shallow ML models (tree ensembles and Support Vector Machines, SVMs) that require the adoption of post-hoc explainability techniques for explaining their decisions.

\subsubsection{Tree Ensembles, Random Forests and Multiple Classifier Systems}

Tree ensembles are arguably among the most accurate ML models in use nowadays. Their advent came as an efficient means to improve the generalization capability of single decision trees, which are usually prone to overfitting. To circumvent this issue, tree ensembles combine different trees to obtain an aggregated prediction/regression. While it results to be effective against overfitting, the combination of models makes the interpretation of the overall ensemble more complex than each of its compounding tree learners, forcing the user to draw from post-hoc explainability techniques. For tree ensembles, techniques found in the literature are explanation by simplification and \textit{feature relevance} techniques; we next examine recent advances in these techniques. 

To begin with, many contributions have been presented to simplify tree ensembles while maintaining part of the accuracy accounted for the added complexity. The author from \cite{domingos1998knowledge} poses the idea of training a single albeit less complex model from a set of random samples from the data (ideally following the real data distribution) labeled by the ensemble model. Another approach for simplification is that in \cite{Intrees}, in which authors create a Simplified Tree Ensemble Learner (STEL). Likewise, \cite{MakingTEInterpretable} presents the usage of two models (simple and complex) being the former the one in charge of interpretation and the latter of prediction by means of Expectation-Maximization and Kullback-Leibler divergence. As opposed to what was seen in model-agnostic techniques, not that many techniques to board explainability in tree ensembles by means of \textit{model simplification}. It derives from this that either the proposed techniques are good enough, or model-agnostic techniques do cover the scope of simplification already. 

Following simplification procedures, \textit{feature relevance} techniques are also used in the field of tree ensembles. Breiman \cite{CostComplexityPrunning} was the first to analyze the variable importance within Random Forests. His method is based on measuring MDA (Mean Decrease Accuracy) or MIE (Mean Increase Error) of the forest when a certain variable is randomly permuted in the out-of-bag samples. Following this contribution \cite{auret2012interpretation} shows, in an real setting, how the usage of variable importance reflects the underlying relationships of a complex system modeled by a Random Forest. Finally, a crosswise technique among post-hoc explainability, \cite{FeatureTweaking} proposes a framework that poses recommendations that, if taken, would convert an example from one class to another. This idea attempts to disentangle the variables importance in a way that is further descriptive. In the article, the authors show how these methods can be used to elevate recommendations to improve malicious online ads to make them rank higher in paying rates.

Similar to the trend shown in model-agnostic techniques, for tree ensembles again, simplification and \textit{feature relevance} techniques seem to be the most used schemes. However, contrarily to what was observed before, most papers date back from 2017 {\color{black}and place their focus mostly on bagging ensembles. When shifting the focus towards other ensemble strategies, scarce activity has been recently noted around the explainability of boosting and stacking classifiers. Among the latter, it is worth highlighting the connection between the reason why a compounding learner of the ensemble produces an specific prediction on a given data, and its contribution to the output of the ensemble. The so-called Stacking With Auxiliary Features (SWAF) approach proposed in \cite{rajani2018stacking} points in this direction by harnessing and integrating explanations in stacking ensembles to improve their generalization. This strategy allows not only relying on the output of the compounding learners, but also on the origin of that output and its consensus across the entire ensemble. Other interesting studies on the explainability of ensemble techniques include model-agnostic schemes such as DeepSHAP \cite{1911.11888}, put into practice with stacking ensembles and multiple classifier systems in addition to Deep Learning models; the combination of explanation maps of multiple classifiers to produce improved explanations of the ensemble to which they belong \cite{rajani2018ensembling}; and recent insights dealing with traditional and gradient boosting ensembles \cite{1907.02582, 1802.03888}}.

\subsubsection{Support Vector Machines}

Another shallow ML model with historical presence in the literature is the SVM. SVM models are more complex than tree ensembles, with a much opaquer structure. Many implementations of post-hoc explainability techniques have been proposed to relate what is mathematically described internally in these models, to what different authors considered explanations about the problem at hand. Technically, an SVM constructs a hyper-plane or set of hyper-planes in a high or infinite-dimensional space, which can be used for classification, regression, or other tasks such as outlier detection. Intuitively, a good separation is achieved by the hyperplane that has the largest distance (so-called functional margin) to the nearest training-data point of any class, since in general, the larger the margin, the lower the generalization error of the classifier. SVMs are among the most used ML models due to their excellent prediction and generalization capabilities. From the techniques stated in Section 2, post-hoc explainability applied to SVMs covers explanation by \textit{simplification}, \textit{local explanations}, \textit{visualizations} and \textit{explanations by example}.

Among explanation by simplification, four classes of simplifications are made. Each of them differentiates from the other by how deep they go into the algorithm inner structure. First, some authors propose techniques to build rule based models only from the support vectors of a trained model. This is the approach of \cite{barakat2007rule}, which proposes a method that extracts rules directly from the support vectors of a trained SVM using a modified sequential covering algorithm. In \cite{barakat2008} the same authors propose eclectic rule extraction, still considering only the support vectors of a trained model. The work in \cite{Chaves2005} generates fuzzy rules instead of classical propositional rules. Here, the authors argue that long antecedents reduce comprehensibility, hence, a fuzzy approach allows for a more linguistically understandable result. The second class of simplifications can be exemplified by \cite{fu2004}, which proposed the addition of the SVM's hyperplane, along with the support vectors, to the components in charge of creating the rules. His method relies on the creation of hyper-rectangles from the intersections between the support vectors and the hyper-plane. In a third approach to \textit{model simplification}, another group of authors considered adding the actual training data as a component for building the rules. In \cite{nunez2002rule,nunez2006,nunez2002B} the authors proposed a clustering method to group prototype vectors for each class. By combining them with the support vectors, it allowed defining ellipsoids and hyper-rectangles in the input space. Similarly in \cite{zhang2005}, the authors proposed the so-called Hyper-rectangle Rule Extraction, an algorithm based on SVC (Support Vector Clustering) to find prototype vectors for each class and then define small hyper-rectangles around. In \cite{fung2005}, the authors formulate the rule extraction problem as a multi-constrained optimization to create a set of non-overlapping rules. Each rule conveys a non-empty hyper-cube with a shared edge with the hyper-plane. In a similar study conducted in \cite{chen2007}, extracting rules for gene expression data, the authors presented a novel technique as a component of a multi-kernel SVM. This multi-kernel method consists of feature selection, prediction modeling and rule extraction. Finally, the study in \cite{intepretationSVM} makes use of a growing SVC to give an interpretation to SVM decisions in terms of linear rules that define the space in Voronoi sections from the extracted prototypes.

Leaving aside rule extraction, the literature has also contemplated some other techniques to contribute to the interpretation of SVMs. Three of them (visualization techniques) are clearly used toward explaining SVM models when used for concrete applications. For instance, \cite{ustun2007visualisation} presents an innovative approach to visualize trained SVM to extract the information content from the kernel matrix. They center the study on Support Vector Regression models. They show the ability of the algorithm to visualize which of the input variables are actually related with the associated output data. In \cite{interpretingHeatMapSVM} a visual way combines the output of the SVM with heatmaps to guide the modification of compounds in late stages of drug discovery. They assign colors to atoms based on the weights of a trained linear SVM that allows for a much more comprehensive way of debugging the process. In \cite{interpretingNeuroSVM} the authors argue that many of the presented studies for interpreting SVMs only account for the weight vectors, leaving the margin aside. In their study they show how this margin is important, and they create an statistic that explicitly accounts for the SVM margin. The authors show how this statistic is specific enough to explain the multivariate patterns shown in neuroimaging.

Noteworthy is also the intersection between SVMs and Bayesian systems, the latter being adopted as a post-hoc technique to explain decisions made by the SVM model. This is the case of \cite{probabilisticSVM} and \cite{bayesianForSVM}, which are studies where SVMs are interpreted as MAP (Maximum A Posteriori) solutions to inference problems with Gaussian Process priors. This framework makes tuning the hyper-parameters comprehensible and gives the capability of predicting class probabilities instead of the classical binary classification of SVMs. Interpretability of SVM models becomes even more involved when dealing with non-CPD (Conditional Positive Definite) kernels that are usually harder to interpret due to missing geometrical and theoretical understanding. The work in \cite{haasdonk2005feature} revolves around this issue with a geometrical interpretation of indefinite kernel SVMs, showing that these do not classify by hyper-plane margin optimization. Instead, they minimize the distance between convex hulls in pseudo-Euclidean spaces.

A difference might be appreciated between the post-hoc techniques applied to other models and those noted for SVMs. In previous models, \textit{model simplification} in a broad sense was the prominent method for post-hoc explainability. In SVMs, \textit{local explanations} have started to take some weight among the propositions. However, simplification based methods are, on average, much older than local explanations. 

As a final remark, none of the reviewed methods treating SVM explainability are dated beyond 2017, which might be due to the progressive proliferation of DL models in almost all disciplines. Another plausible reason is that these models are already understood, so it is hard to improve upon what has already been done.

\subsection{Explainability in Deep Learning} \label{ssec:deep-posthoc}

Post-hoc \textit{local explanations} and \textit{feature relevance} techniques are increasingly the most adopted methods for explaining DNNs. 
This section reviews explainability studies proposed for the most used DL models, namely multi-layer neural networks, Convolutional Neural Networks (CNN) and Recurrent Neural Networks (RNN).
 
\subsubsection{Multi-layer Neural Networks}

From their inception, multi-layer neural networks (also known as multi-layer perceptrons) have been warmly welcomed by the academic community due to their huge ability to infer complex relations among variables. However, as stated in the introduction, developers and engineers in charge of deploying these models in real-life production find in their questionable explainability a common reason for reluctance. That is why neural networks have been always considered as black-box models. The fact that explainability is often a must for the model to be of practical value, forced the community to generate multiple explainability techniques for multi-layer neural networks, including \textit{model simplification} approaches, \textit{feature relevance} estimators, \textit{text explanations}, \textit{local explanations} and model \textit{visualizations}. 

Several \textit{model simplification} techniques have been proposed for neural networks with one single hidden layer, however very few works have been presented for neural networks with multiple hidden layers. One of these few works is DeepRED algorithm \cite{zilke2016deepred}, which extends the decompositional approach to rule extraction (splitting at neuron level) presented in \cite{CRED} for multi-layer neural network by adding more decision trees and rules.

Some other works use \textit{model simplification} as a post-hoc explainability approach. For instance, \cite{InterpretableDeepICU} presents a simple distillation method called \emph{Interpretable Mimic Learning} to extract an interpretable model by means of gradient boosting trees. In the same direction, the authors in \cite{TreeView} propose a hierarchical partitioning of the feature space that reveals the iterative rejection of unlikely class labels, until association is predicted. In addition, several works addressed the distillation of knowledge from  an ensemble of models into a single model \cite{hinton2015distilling,bucilua2006model,Traore19} .

Given the fact that the simplification of  multi-layer neural networks is more complex as the number of layers increases, explaining these models by \textit{feature relevance} methods has become progressively more popular. One of the representative works in this area is \cite{DeepTaylor}, which presents a method to decompose the network classification decision into contributions of its input elements. They consider each neuron as an object that can be decomposed and expanded then aggregate and back-propagate these decompositions through the network, resulting in a \emph{deep} Taylor decomposition. In the same direction, the authors in \cite{shrikumar2016not} proposed DeepLIFT, an approach for computing importance scores in a multi-layer neural network. Their method compares the activation of a neuron to the reference activation and assigns the score according to the difference. 

On the other hand, some works try to verify the theoretical soundness of current explainability methods. For example, the authors in \cite{Axiomatic},  bring up a fundamental problem of most \textit{feature relevance} techniques, designed for multi-layer networks. They showed that two axioms that such techniques ought to fulfill namely, \emph{sensitivity} and \emph{implementation invariance}, are violated in practice by most approaches. Following these axioms, the authors of \cite{Axiomatic} created \emph{integrated gradients}, a new \textit{feature relevance} method proven to meet the aforementioned axioms. Similarly, the authors in \cite{LearningHowTo} analyzed the correctness of  current  \textit{feature relevance} explanation approaches designed for Deep Neural Networks, e,g., DeConvNet, Guided BackProp and LRP, on simple linear neural networks. Their analysis showed that these methods do not  produce the theoretically correct explanation and presented two new explanation methods \textit{PatternNet} and \textit{PatternAttribution} that are more theoretically sound for both, simple and deep neural networks.

\subsubsection{Convolutional Neural Networks}

Currently, CNNs constitute the state-of-art models in all fundamental computer vision tasks, from image classification and object detection to instance segmentation. Typically, these models are built as a sequence of convolutional layers and pooling layers to automatically learn increasingly higher level features. At the end of the sequence, one or multiple fully connected layers are used to map the output features map into scores. This structure entails extremely complex internal relations that are very difficult to explain. Fortunately, the road to explainability for CNNs is easier than for other types of models, as the human cognitive skills favors the understanding of visual data. 
 
Existing works that aim at understanding what CNNs learn can be divided into two broad categories: 1) those that try to understand the decision process by mapping back the output in the input space to see which parts of the input were discriminative for the output; and 2) those that try to delve inside the network and interpret how the intermediate layers see the external world, not necessarily related to any specific input, but in general. 
  
One of the seminal works in the first category was \cite{AdaptiveDeconv}. When an input image runs feed-forward through a CNN, each layer outputs a number of feature maps with strong and soft activations. The authors in \cite{AdaptiveDeconv} used Deconvnet, a network designed previously by the same authors \cite{zeiler2010deconvolutional} that, when fed with a feature map from a selected layer, reconstructs the maximum activations. These reconstructions can give an idea about the parts of the image that produced that effect. To visualize these strongest activations in the input image, the same authors used the occlusion sensitivity method to generate a saliency map \cite{VisualizingUnderstanding}, which consists of iteratively forwarding the same image through the network occluding a different region at a time. 

To improve the quality of the mapping on the input space, several subsequent papers proposed simplifying both the CNN architecture and the visualization method. In particular, \cite{LearningDeepFeatures} included a global average pooling layer between the last convolutional layer of the CNN and the fully-connected layer that predicts the object class. With this simple architectural modification of the CNN, the authors built a class activation map that helps identify the image regions that were particularly important for a specific object class by projecting back the weights of the output layer on the convolutional feature maps. Later, in \cite{springenberg2014striving}, the authors showed that max-pooling layers can be used to replace convolutional layers with a large stride without loss in accuracy on several image recognition benchmarks. They obtained a cleaner visualization than Deconvnet by using a guided backpropagation method. 

\begin{figure}[htbp!]
\centering
\resizebox{\columnwidth}{!}{
\begin{tabular}{c@{\qquad}c@{\qquad}c}
\includegraphics[height=1.35in]{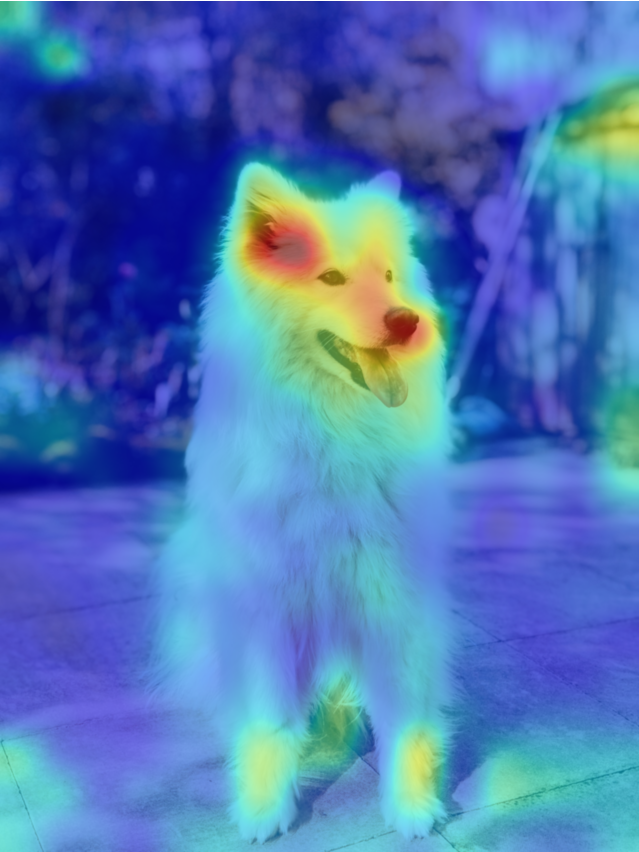}  & 
\includegraphics[height=1.35in]{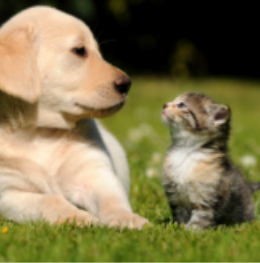} \includegraphics[height=1.35in]{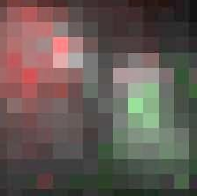}& 
\includegraphics[height=1.35in]{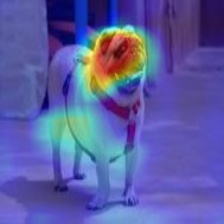} \\ 
      (a) Heatmap \cite{Burns18} & 
      (b) Attribution \cite{Olah17} & 
      (c) Grad-CAM \cite{selvaraju2017grad}
  \end{tabular} 
}
  \caption{Examples of rendering for different XAI visualization techniques on images.\label{fig:visualisation}}
\end{figure}

To increase the interpretability of classical CNNs, the authors in \cite{InterpretableCNN} used a loss for each filter in high level convolutional layers to force each filter to learn very specific object components. The obtained activation patterns are much more interpretable for their exclusiveness with respect to the different labels to be predicted. The authors in \cite{LayerWise} proposed visualizing the contribution to the prediction of each single pixel of the input image in the form of a heatmap. They used a Layer-wise Relevance Propagation (LRP) technique, which relies on a Taylor series close to the prediction point rather than partial derivatives at the prediction point itself. To further improve the quality of the visualization, attribution methods such as heatmaps, saliency maps or class activation methods (\textit{GradCAM} \cite{selvaraju2017grad}) are used (see Figure \ref{fig:visualisation}). In particular, the authors in \cite{selvaraju2017grad} proposed a Gradient-weighted Class Activation Mapping (Grad-CAM), which uses the gradients of any target concept, flowing into the final convolutional layer to produce a coarse localization map, highlighting the important regions in the image for predicting the concept.

In addition to the aforementioned {\it feature relevance} and {\it visual} explanation methods, some works proposed generating  \textit{text explanations} of the visual content of the image. For example, the authors in \cite{xu2015show} combined a CNN feature extractor with an RNN attention model to automatically learn to describe the content of images. In the same line, \cite{xiao2015application} presented a three-level attention model to perform a fine-grained classification task. The general model is a pipeline that integrates three types of attention: the object level attention model proposes candidate image regions or patches from the input image, the part-level attention model filters out non-relevant patches to a certain object, and the last attention model localizes discriminative patches. In the task of video captioning, the authors in \cite{ImprovingInterpretability} use a CNN model combined with a bi-directional LSTM model as encoder to extract video features and then feed these features to an LSTM decoder to generate textual descriptions. 

One of the seminal works in  the second category is \cite{UnderstandingDeep}. In order to analyse the visual information contained inside the CNN, the authors proposed a general framework that reconstruct an image from the CNN internal representations and showed that several layers retain photographically accurate information about the image, with different degrees of geometric and photometric invariance. To visualize the notion of a class captured by a CNN, the same authors created an image that maximizes the class score based on computing the gradient of the class score with respect to the input image \cite{InsideConv}. In the same direction, the authors in \cite{SynthesizingPreferredInputs} introduced a Deep Generator Network (DGN) that generates the most representative image for a given output neuron in a CNN. 

For quantifying the interpretability of the latent representations of CNNs, the authors in \cite{QuantifyingInterpretability} used a different approach called network dissection. They run a large number of images through a CNN and then analyze the top activated images by considering each unit as a concept detector to further evaluate each unit for semantic segmentation. This paper also examines the effects of classical training techniques on the interpretability of the learned model. 

Although many of the techniques examined above utilize \textit{local explanations} to achieve an overall explanation of a CNN model, others explicitly focus on building global explanations based on locally found prototypes. In \cite{adebayo2018local,adebayo2018sanity}, the authors empirically showed how \textit{local explanations} in deep networks are strongly dominated by their lower level features. They demonstrated that deep architectures provide strong priors that prevent the altering of how these low-level representations are captured. All in all, \textit{visualization} mixed with \textit{feature relevance} methods are arguably the most adopted approach to explainability in CNNs. 

\begin{figure}[htbp!]
\centering
\resizebox{0.8\columnwidth}{!}{
\begin{tabular}{c@{\qquad}c@{\qquad}c}
\includegraphics[height=1.35in]{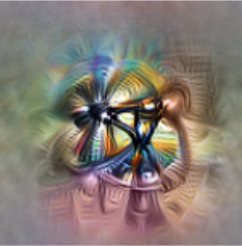}  & 
\includegraphics[height=1.35in]{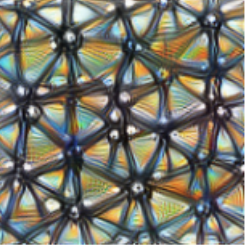} & 
\includegraphics[height=1.35in]{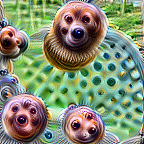} \\ 
(a) Neuron & (b) Channel & (c) Layer 
  \end{tabular}}
  \caption{Feature visualization at different levels of a certain network \cite{Olah17}.\label{fig:feature}}
\end{figure}

Instead of using one single interpretability technique, the framework proposed in \cite{Olah18} combines several methods to provide much more information  about the network. For example, combining feature visualization (\textit{what is a neuron looking for?}) with attribution (\textit{how does it affect the output?}) allows exploring how the network decides between labels. This visual interpretability interface displays different blocks such as feature visualization and attribution depending on the visualization goal. This interface can be thought of as a union of individual elements that belong to layers (input, hidden, output), atoms (a neuron, channel, spatial or neuron group), content (activations -- the amount a neuron fires, attribution -- which classes a spatial position most contributes to, which tends to be more meaningful in later layers), and presentation (information visualization, feature visualization). Figure \ref{fig:feature} shows some examples. Attribution methods normally rely on pixel association, displaying what part of an input example is responsible for the network activating in a particular way \cite{Olah17}. 

\begin{figure}[htbp!]
\centering
  \begin{tabular}{c@{\qquad}c@{\qquad}c}
\includegraphics[height=1.35in]{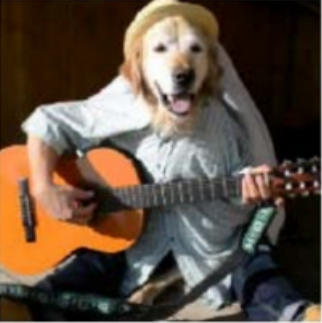}  & 
\includegraphics[height=1.35in]{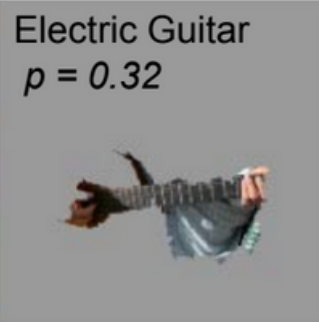} & 
\includegraphics[height=1.35in]{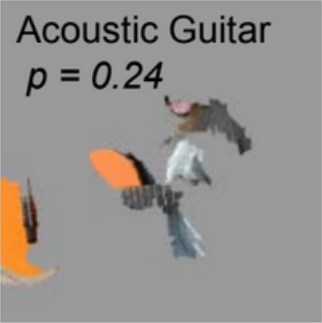} \\ 
(a) Original image & (b) Explaining \textit{electric guitar} & (c) Explaining \textit{acoustic guitar} 
  \end{tabular} 
  \caption{Examples of explanation when using LIME on images \cite{LIME}.\label{fig:lime}}
\end{figure}

A much simpler approach to all the previously cited methods was proposed in LIME framework \cite{LIME}, as was described in Subsection \ref{sec:model-agnostic} LIME perturbs the input and sees how the predictions change. In image classification, LIME creates a set of perturbed instances by dividing the input image into interpretable components (contiguous \textit{superpixels}), and runs each perturbed instance through the model to get a probability. A simple linear model learns on this data set, which is locally weighted. At the end of the process, LIME presents the superpixels with highest positive weights as an explanation (see Figure \ref{fig:lime}).

A completely different explainability approach is proposed in adversarial detection. To understand model failures in detecting adversarial examples, the authors in \cite{papernot2018deep} apply the k-nearest neighbors algorithm on the representations of the data learned by each layer of the CNN. A test input image  is considered as adversarial if its representations are far from the representations of the training images.

\subsubsection{Recurrent Neural Networks}

As occurs with CNNs in the visual domain, RNNs have lately been used extensively for predictive problems defined over inherently sequential data, with a notable presence in natural language processing and time series analysis. These types of data exhibit long-term dependencies that are complex to be captured by a ML model. RNNs are able to retrieve such time-dependent relationships by formulating the retention of knowledge in the neuron as another parametric characteristic that can be learned from data.  

Few contributions have been made for explaining RNN models. These studies can be divided into two groups: 1) explainability by understanding what a RNN model has learned (mainly via \textit{feature relevance} methods); and 2) explainability by modifying RNN architectures to provide insights about the decisions they make (\textit{local explanations}). 

In the first group, the authors in  \cite{ExplainingRNN} extend the usage of LRP to RNNs. They  propose a specific propagation rule that works with multiplicative connections as those in LSTMs (Long Short Term Memory) units and GRUs (Gated Recurrent Units). The authors in \cite{VisualizingUnderstandingRNN} propose a visualization technique based on finite horizon n-grams that discriminates interpretable cells within LSTM and GRU networks. Following the premise of not altering the architecture, \cite{DistillingRNN} extends the interpretable mimic learning distillation method used for CNN models to LSTM networks, so that interpretable features are learned by fitting Gradient Boosting Trees to the trained LSTM network under focus. 

Aside from the approaches that do not change the inner workings of the RNNs, \cite{RETAIN} presents RETAIN (REverse Time AttentIoN) model, which detects influential past patterns by means of a two-level neural attention model. To create an interpretable RNN, the authors in \cite{InterpretableRNN} propose an RNN based on SISTA (Sequential Iterative Soft-Thresholding Algorithm) that models a sequence of correlated observations with a sequence of sparse latent vectors, making its weights interpretable as the parameters of a principled statistical model. Finally, \cite{MarkovRNN} constructs a combination of an HMM (Hidden Markov Model) and an RNN, so that the overall model approach harnesses the interpretability of the HMM and the accuracy of the RNN model.

\subsubsection{Hybrid Transparent and Black-box Methods} 

The use of background knowledge in the form of logical statements or constraints in Knowledge Bases (KBs) has shown to not only improve explainability but also performance with respect to purely data-driven approaches \cite{Donadello17,donadello2018semantic,dAvilaGarcez19NeSy}. A positive side effect shown is that this hybrid approach provides robustness to the learning system when errors are present in the training data labels. Other approaches have shown to be able to jointly learn and reason with both symbolic and sub-symbolic representations and inference. The interesting aspect is that this blend allows for expressive probabilistic-logical reasoning in an end-to-end fashion \cite{manhaeve2018deepproblog}. A successful use case is on dietary recommendations, where explanations are extracted from the reasoning behind (non-deep but KB-based) models \cite{Donadello19}.

Future data fusion approaches may thus consider endowing DL models with explainability by externalizing other domain information sources. Deep formulation of classical ML models has been done, e.g. in Deep Kalman filters (DKFs) \cite{Krishnan15}, Deep Variational Bayes Filters (DVBFs) \cite{Karl16}, Structural Variational Autoencoders (SVAE) \cite{Johnson16}, or conditional random fields as RNNs \cite{Zheng15}. These approaches provide deep models with the interpretability inherent to probabilistic graphical models. For instance, SVAE combines probabilistic graphical models in the embedding space with neural networks to enhance the interpretability of DKFs. A particular example of classical ML model enhanced with its DL counterpart is Deep Nearest Neighbors DkNN \cite{papernot2018deep}, where the neighbors constitute human-interpretable explanations of predictions. The intuition is based on the rationalization of a DNN prediction based on evidence. This evidence consists of a characterization of confidence termed \textit{credibility} that spans the hierarchy of representations within a DNN, that must be supported by the training data \cite{papernot2018deep}. 
\begin{figure}[ht]
        \center{\includegraphics[width=0.8\columnwidth]{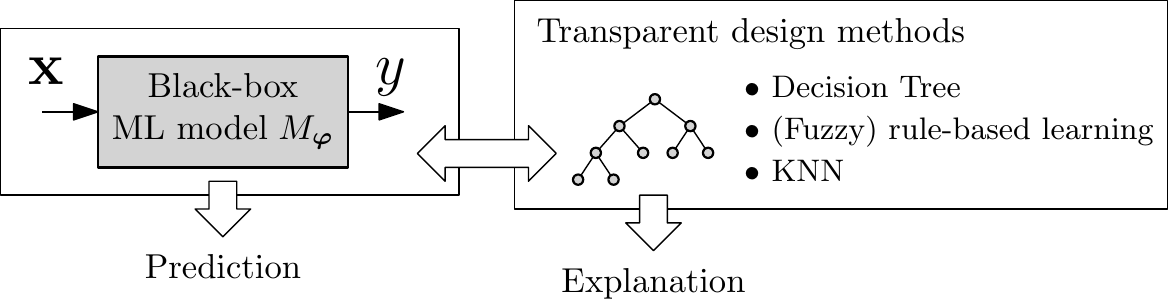}}
        \caption{\label{fig:hybrid} Pictorial representation of a hybrid model. A neural network considered as a black-box can be explained by associating it to a more interpretable model such as a Decision Tree \cite{Narodytska18}, a (fuzzy) rule-based system \cite{Fernandez19} or KNN \cite{papernot2018deep}. 
        }
\end{figure}

A different perspective on hybrid XAI models consists of enriching black-box models knowledge with that one of transparent ones, as proposed in \cite{WhatDoesExplainableAImean} and further refined in \cite{Bennetot19} {\color{black}and \cite{loyola2019black}. In particular, this can be done by constraining the neural network thanks to a semantic KB and bias-prone concepts \cite{Bennetot19}, or by stacking ensembles jointly encompassing white- and black-box models \cite{loyola2019black}.} 

Other examples of hybrid symbolic and sub-symbolic methods where a knowledge-base tool or graph-perspective enhances the neural (e.g., language \cite{petroni2019language}) model are in \cite{Bollacker19,Shang19}. In reinforcement learning, very few examples of symbolic (graphical \cite{Zolotas19} or relational  \cite{santoro2017simple,garnelo2016towards}) hybrid models exist, while in recommendation systems, for instance, explainable autoencoders are proposed \cite{Bellini18}. A specific transformer architecture symbolic visualization method (applied to music) pictorially shows how soft-max attention works \cite{huang2018music}. By visualizing self-reference, i.e., the last layer of attention weights, arcs show which notes in the past are informing the future and how attention is skip over less relevant sections. Transformers can also help explain image captions visually \cite{cornia2019smart}. 

Another hybrid approach consists of mapping an uninterpretable black-box system to a white-box \textit{twin} that is more interpretable. For example, an opaque neural network can be combined with a transparent Case Based Reasoning (CBR) system \cite{Aamodt94, Caruana99}. In \cite{Keane19}, the DNN and the CBR (in this case a kNN) are paired in order to improve interpretability while keeping the same accuracy. The \textit{explanation by example} consists of analyzing the feature weights of the DNN which are then used in the CBR, in order to retrieve nearest-neighbor cases to explain the DNN’s prediction. 

\subsection{Alternative Taxonomy of Post-hoc Explainability Techniques for Deep Learning} \label{ssec:second_tax}

DL is the model family where most research has been concentrated in recent times and they have become central for most of the recent literature on XAI. While the division between model-agnostic and model-specific is the most common distinction made, the community has not only relied on this criteria to classify XAI methods. For instance, some model-agnostic methods such as \textit{SHAP} \cite{lundberg2017unified} are widely used to explain DL models. That is why several XAI methods can be easily categorized in different taxonomy branches depending on the angle the method is looked at. An example is LIME which can also be used over CNNs, despite not being exclusive to deal with images. Searching within the alternative DL taxonomy shows us that LIME can explicitly be used for \textit{Explaining a Deep Network Processing}, as a kind of \textit{Linear Proxy Model}. Another type of classification is indeed proposed in \cite{Gilpin18} with a segmentation based on 3 categories. The first category groups methods explaining the processing of data by the network, thus answering to the question \emph{``why does this particular input leads to this particular output?''}. The second one concerns methods explaining the representation of data inside the network, i.e., answering to the question \emph{``what information does the network contain?''}. The third approach concerns models specifically designed to simplify the interpretation of their own behavior. Such a multiplicity of classification possibilities leads to different ways of constructing XAI taxonomies.
\begin{figure}[h!]
\begin{tabular}{c@{\qquad}c}
	\resizebox{0.55\columnwidth}{!}{\begin{forest} 
			for tree={
				l sep=5em, s sep=1em,
				child anchor=west,
				parent anchor=east,
				grow'=0,
				line width=0.75mm,
				anchor=west,
				draw,
			}
			[{{\Huge XAI in DL}}
			[{{\Huge Explanation of Deep} \\ { \Huge Network Processing }}            
			[{{\huge Linear Proxy Models  }} \\                             
			{{\Large \cite{ribeiro2016trust}}}\\
			]
			[{{\huge Decision Trees  }} \\                             
			{{\Large \cite{augasta2012reverse, zilke2016deepred, Schmitz99, CRED}}}\\
			]
			[{{\huge Automatic-Rule Extraction  }} \\                             
			{{\Large \cite{Craven96, fu1994rule, Towell93, Thrun94, Setiono00, Taha99, Tsukimoto00, hailesilassie2016rule, Benitez97, Johansson05}}}\\
			]
			[{{\huge Salience Mapping  }} \\                             
			{{\Large \cite{LearningDeepFeatures, VisualizingUnderstanding, Shrikumar17, Axiomatic, InsideConv, ExplainingRNN, Smilkov17, Ancona17}}}\\
			]
			]             
			[{{\Huge Explanation of Deep} \\ {\Huge Network Representation}}        
			[{{\huge  Role of Layers}} \\ 
			{{
			\Large \cite{Yosinski_Bengio14, Razavian14}
			}}\\
			]
			[{{\huge Role of Individual} {\huge Units}} \\
			{{\Large \cite{QuantifyingInterpretability, CarSteer, Zhou14, Zhang18, Franckle18}}}\\
			]
			[{{\huge Role of Representation} {\huge Vectors}} \\
			{{\Large \cite{kim2017interpretability}}}\\
			]
			]
			[{{\Huge Explanation Producing} \\ {\Huge Systems}}        
			[{{\huge Attention Networks}} \\ 
			{{\Large \cite{1909.13584, xiao2015application, Vaswani17, Lu16, Das16, HukPark18, Ross17}}}\\
			]
			[{{\huge Representation Disentanglement}} \\
			{{\Large \cite{InterpretableCNN, Zhang16, Jolliffe86, Hyvarinen00, Berry07, Kingma13, Higgins17, Chen16, Zhang18Disentangle, Sabour17}}}\\
			]
			[{{\huge Explanation Generation}} \\
			{{
			\Large \cite{Hendricks16Generate, Agrawal15, Fukui16, Bouchacourt19}
			}}\\
			]
			]
			[{{\Huge Hybrid Transparent} \\ {\Huge and Black-box Methods}}        
			[{{\huge Neural-symbolic Systems}} \\
			{{\Large \cite{Donadello17,donadello2018semantic,dAvilaGarcez19NeSy, manhaeve2018deepproblog}}}\\
			]
			[{{\huge KB-enhanced Systems}} \\
			{{\Large \cite{WhatDoesExplainableAImean, Bennetot19, Donadello19, petroni2019language, Bollacker19,Shang19}}}\\
			]
			[{{\huge Deep Formulation}} \\
			{{\Large \cite{papernot2018deep, Krishnan15, Karl16, Johnson16, Zheng15}}}\\
			]
			[{{\huge Relational Reasoning}} \\
			{{\Large \cite{santoro2017simple,garnelo2016towards, Bellini18, huang2018music}}}\\
			]
			[{{\huge Case-base Reasoning}} \\
			{{\Large \cite{Aamodt94, Caruana99, Keane19}}}\\
			]
			]
			]
	\end{forest}} & \begin{tabular}{c}\resizebox{0.4\columnwidth}{!}{\includegraphics{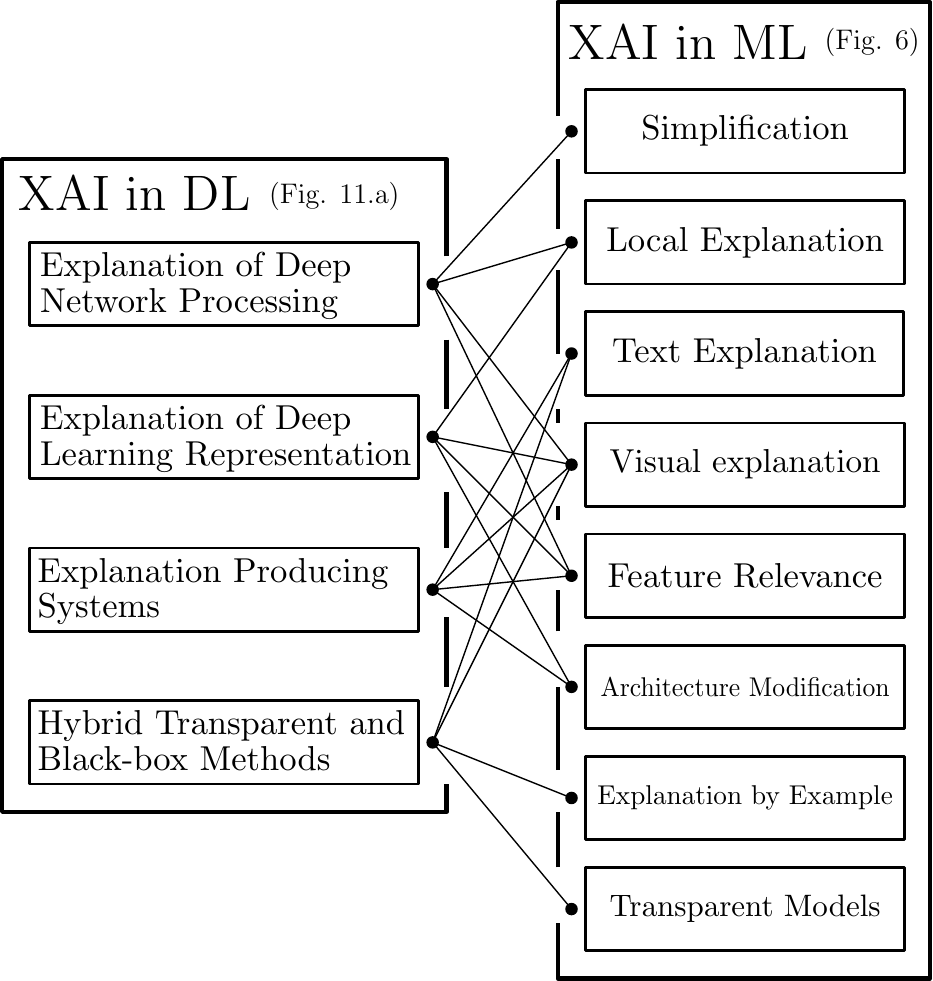}}\end{tabular}\\
	(a) & (b)
\end{tabular}
	\centering
	\caption{(a) Alternative Deep Learning specific taxonomy extended from the categorization from \cite{Gilpin18}; and (b) its connection to the taxonomy in Figure \ref{fig:treeCat}.}
	\label{fig:treeGilpin}
\end{figure}

Figure \ref{fig:treeGilpin} shows the alternative Deep Learning taxonomy inferred from \cite{Gilpin18}. From the latter, it can be deduced the complementarity and overlapping of this taxonomy to Figure \ref{fig:treeCat} as:
\begin{itemize}[leftmargin=*]
    \item Some methods \cite{InsideConv, ExplainingRNN} classified in distinct categories (namely \textit{feature relevance for CNN} and \textit{feature relevance for RNN}) in Figure \ref{fig:treeCat} are included in a single category (\textit{Explanation of Deep Network Processing with Salience Mapping}) when considering the classification from \cite{Gilpin18}. 
    \item Some methods \cite{augasta2012reverse, kim2017interpretability} are classified on a single category (\textit{Explanation by simplification for Multi-Layer Neural Network}) in Figure \ref{fig:treeCat} while being in 2 different categories (namely, \textit{Explanation of Deep Network Processing with Decision Trees} and \textit{Explanation of Deep Network Representation with the Role of Representation Vectors}) in \cite{Gilpin18}, as shown in Figure \ref{fig:treeGilpin}.
\end{itemize}

A classification based on explanations of model processing and explanations of model representation is relevant, as it leads to a differentiation between the execution trace of the model and its internal data structure. This means that depending of the failure reasons of a complex model, it would be possible to pick-up the right XAI method according to the information needed: the execution trace or the data structure. This idea is analogous to testing and debugging methods used in regular programming paradigms \cite{Hofer06}.

\section{XAI: Opportunities, Challenges and {\color{black}Future} Research Needs} \label{sec:challenges}

We now capitalize on the performed literature review to put forward a critique of the achievements, trends and challenges that are still to be addressed in the field of explainability of ML and data fusion models. Actually our discussion on the advances taken so far in this field has already anticipated some of these challenges. In this section we revisit them and explore new research opportunities for XAI, identifying possible research paths that can be followed to address them effectively in years to come:
\begin{itemize}[leftmargin=*]
    \item When introducing the overview in Section \ref{sec:intro} we already mentioned the existence of a tradeoff between model interpretability and performance, in the sense that making a ML model more understandable could eventually degrade the quality of its produced decisions. In Subsection \ref{ssec:tradeoff} we will stress on the potential of XAI developments to effectively achieve an optimal balance between the interpretability and performance of ML models.
    \item In Subsection \ref{sec:what} we stressed on the imperative need for reaching a consensus on \emph{what} explainability entails within the AI realm. Reasons for pursuing explainability are also assorted and, under our own assessment of the literature so far, not unambiguously mentioned throughout related works. In Subsection \ref{ssec:concepts_and_metrics} we will further delve into this important issue.
    \item Given its notable prevalence in the XAI literature, Subsections \ref{ssec:deep-posthoc} and \ref{ssec:second_tax} revolved on the explainability of Deep Learning models, examining advances reported so far around a specific bibliographic taxonomy. We go in this same direction with Subsection \ref{ssec:deep_learning_challenges}, which exposes several challenges that hold in regards to the explainability of this family of models.
    \item Finally, we close up this prospective discussion with Subsections \ref{ssec:robust_adv} to \ref{ssec:guidelines}, which place on the table several research niches that despite its connection to model explainability, remain insufficiently studied by the community. 
\end{itemize}

{\color{black}Before delving into these identified challenges, it is important to bear in mind that this prospective section is complemented by Section \ref{sec:responsibleAI}, which enumerates research needs and open questions related to XAI within a broader context: the need for responsible AI.}

\subsection{On the Tradeoff between Interpretability and Performance} \label{ssec:tradeoff}

The matter of interpretability versus performance is one that repeats itself through time, but as any other big statement, has its surroundings filled with myths and misconceptions.

As perfectly stated in \cite{rudin2018please}, it is not necessarily true that models that are more complex are inherently more accurate. This statement is false in cases in which the data is well structured and features at our disposal are of great quality and value. This case is somewhat common in some industry environments, since features being analyzed are constrained within very controlled physical problems, in which all of the features are highly correlated, and not much of the possible landscape of values can be explored in the data \cite{diez2019data}. What can be hold as true, is that more complex models enjoy much more flexibility than their simpler counterparts, allowing for more complex functions to be approximated. Now, returning to the statement \emph{``models that are more complex are more accurate''}, given the premise that the function to be approximated entails certain complexity, that the data available for study is greatly widespread among the world of suitable values for each variable and that there is enough data to harness a complex model, the statement presents itself as a true statement. It is in this situation that the trade-off between performance and interpretability can be observed. It should be noted that the attempt at solving problems that do not respect the aforementioned premises will fall on the trap of attempting to solve a problem that does not provide enough data diversity (variance). Hence, the added complexity of the model will only fight against the task of accurately solving the problem. 
\begin{figure}[h!]
    	\center{\includegraphics[width=0.75\columnwidth]{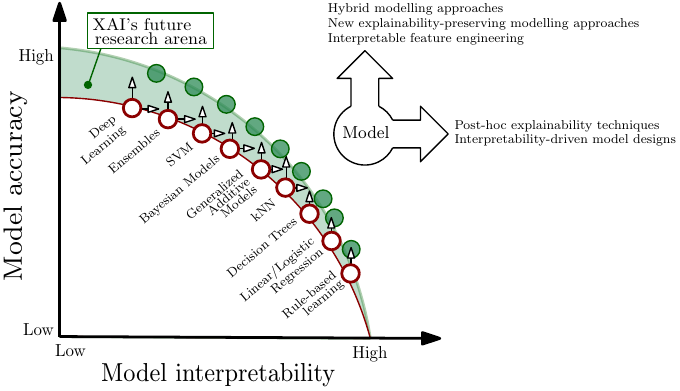}}
	\caption{Trade-off between model interpretability and performance, and a representation of the area of improvement where the potential of XAI techniques and tools resides.}
	\label{fig:tradeoff}
\end{figure}

In this path toward performance, when the performance comes hand in hand with complexity, interpretability encounters itself on a downwards slope that until now appeared unavoidable. However, the apparition of more sophisticated methods for explainability could invert or at least cancel that slope. Figure \ref{fig:tradeoff} shows a tentative representation inspired by previous works \cite{gunning2017explainable}, in which XAI shows its power to improve the common trade-off between model interpretability and performance. {\color{black}Another aspect worth mentioning at this point due to its close link to model interpretability and performance is the \emph{approximation dilemma}: explanations made for a ML model must be made drastic and approximate enough to match the requirements of the audience for which they are sought, ensuring that explanations are representative of the studied model and do not oversimplify its essential features.}

\subsection{On the Concept and Metrics} \label{ssec:concepts_and_metrics}

The literature clearly asks for an unified concept of explainability. In order for the field to thrive, it is imperative to place a common ground upon which the community is enabled to contribute new techniques and methods. A common concept must convey the needs expressed in the field. It should propose a common structure for every XAI system. This paper attempted a new proposition of a concept of explainability that is built upon that from Gunning \cite{gunning2017explainable}. In that proposition and the following strokes to complete it (Subsection \ref{sec:what}), explainability is defined as the ability a model has to make its functioning clearer to an audience. To address it, post-hoc type methods exist. The concept portrayed in this survey might not be complete but as it stands, allows for a first common ground and reference point to sustain a profitable discussion in this matter. It is paramount that the field of XAI reaches an agreement in this respect combining the shattered efforts of a widespread field behind the same banner.

Another key feature needed to relate a certain model to this concrete concept is the existence of a metric. A metric, or group of them should allow for a meaningful comparison of how well a model fits the definition of explainable. Without such tool, any claim in this respect dilutes among the literature, not providing a solid ground on which to stand. These metrics, as the classic ones (accuracy, F1, sensitivity...), should express how well the model performs in a certain aspect of explainability. {\color{black}Some attempts have been done recently around the measurement of XAI, as reviewed thoroughly in \cite{1812.04608,1811.11839}. In general, XAI measurements should evaluate the goodness, usefulness and satisfaction of explanations, the improvement of the mental model of the audience induced by model explanations, and the impact of explanations on the performance of the model and on the trust and reliance of the audience. Measurement techniques surveyed in \cite{1812.04608} and \cite{1811.11839} (e.g., goodness checklist, explanation satisfaction scale, elicitation methods for mental models, computational measures for explainer fidelity, explanation trustworthiness and model reliability) seem to be a good push in the direction of evaluating XAI techniques. Unfortunately, conclusions drawn from these overviews are aligned with our prospects on the field: more quantifiable, general XAI metrics are really needed to support the existing measurement procedures and tools proposed by the community.}

This survey does not tackle the problem of designing such a suite of metrics, since such a task should be approached by the community as a whole prior acceptance of the broader concept of explainability, which on the other hand, is one of the aims of the current work. {\color{black}Nevertheless, we advocate for further efforts towards new proposals to evaluate the performance of XAI techniques, as well as comparison methodologies among XAI approaches that allow contrasting them quantitatively under different application context, models and purposes.}

\subsection{Challenges to achieve Explainable Deep Learning} \label{ssec:deep_learning_challenges}

While many efforts are currently being made in the area of XAI, there are still many challenges to be faced before being able to obtain explainability in DL models. First, as explained in Subsection \ref{sec:what}, there is a lack of agreement on the vocabulary and the different definitions surrounding XAI. As an example, we often see the terms \textit{feature importance} and \textit{feature relevance} referring to the same concept. This is even more obvious for visualization methods, where there is absolutely no consistency behind what is known as saliency maps, salient masks, heatmaps, neuron activations, attribution, and other approaches alike. As XAI is a relatively young field, the community does not have a standardized terminology yet. 

As it has been commented in Subsection \ref{ssec:tradeoff}, there is a trade-off between interpretability and accuracy \cite{Gilpin18}, i.e., between the simplicity of the information given by the system on its internal functioning, and the exhaustiveness of this description. Whether the observer is an expert in the field, a policy-maker or a user without machine learning knowledge, intelligibility does not have to be at the same level in order to provide the \textit{audience} an understanding \cite{Preece18Stakeholders}. This is one of the reasons why, as mentioned above, a challenge in XAI is establishing objective metrics on what constitutes a good explanation. A possibility to reduce this subjectivity is taking inspiration from experiments on human psychology, sociology or cognitive sciences to create objectively convincing explanations. Relevant findings to be considered when creating an explainable AI model are highlighted in \cite{Miller19}: First, explanations are better when \textit{constrictive}, meaning that a prerequisite for a good explanation is that it does not only indicate why the model made a decision X, but also why it made decision X rather than decision Y. It is also explained that probabilities are not as important as causal links in order to provide a satisfying explanation. Considering that black box models tend to process data in a quantitative manner, it would be necessary to translate the probabilistic results into qualitative notions containing causal links. In addition, they state that explanations are \textit{selective}, meaning that focusing solely on the main causes of a decision-making process is sufficient. It was also shown that the use of counterfactual explanations can help the user to understand the decision of a model \cite{goudet2018learning, Lopez-Paz17, Byrne19}.

Combining connectionist and symbolic paradigms seems a favourable way to address this challenge \cite{Bennetot19,dAvilaGarcez19NeSy,garnelo2016towards,garnelo2019reconciling,marra2019integrating}. On one hand, connectionist methods are more precise but opaque. On the other hand, symbolic methods are popularly considered less efficient, while they offer a greater explainability thus respecting the conditions mentioned above:
\begin{itemize}[leftmargin=*]
    \item The ability to refer to established reasoning rules allows symbolic methods to be constrictive.
    \item The use of a KB formalized e.g. by an ontology can allow data to be processed directly in a qualitative way.
    \item Being selective is less straightforward for connectionist models than for symbolic ones. 
\end{itemize}

Recalling that a good explanation needs to influence the mental model of the user, i.e. the representation of the external reality using, among other things, symbols, it seems obvious that the use of the symbolic learning paradigm is appropriate to produce an explanation. Therefore, neural-symbolic interpretability could provide convincing explanations while keeping or improving generic performance \cite{Donadello17}. 

As stated in \cite{WhatDoesExplainableAImean}, a truly explainable model should not leave explanation generation to the users as different explanations may be deduced depending on their background knowledge. Having a semantic representation of the knowledge can help a model to have the ability to produce explanations (e.g., in natural language \cite{Bennetot19}) combining common sense reasoning and human-understandable features. 

Furthermore, until an objective metric has been adopted, it appears necessary to make an effort to rigorously formalize evaluation methods. One way may be drawing inspiration from the social sciences, e.g., by being consistent when choosing the evaluation questions and the population sample used \cite{Kelley03}.

A final challenge XAI methods for DL need to address is providing explanations that are accessible for society, policy makers and the law as a whole. In particular, conveying explanations that require non-technical expertise will be paramount to both handle ambiguities, and to develop the social right to the (not-yet available) right for explanation in the EU General Data Protection Regulation (GDPR) \cite{Wachter17}.

\subsection{{\color{black}Explanations for AI Security: XAI and Adversarial Machine Learning}} \label{ssec:robust_adv}

Nothing has been said about confidentiality concerns linked to XAI. One of the last surveys very briefly introduced the idea of algorithm property and trade secrets \cite{adadi2018peeking}. However, not much attention has been payed to these concepts. If \emph{confidential} is the property that makes something \textit{secret}, in the AI context many aspects involved in a model may hold this property. For example, imagine a  model that some company has developed through many years of research in a specific field. The knowledge synthesized in the model built might be considered to be confidential, and it may be compromised even by providing only input and output access \cite{Orekondy18}. The latter shows that, under minimal assumptions, \textit{data model functionality stealing} is possible. An approach that has served to make DL models more robust against intellectual property exposure based on a sequence of non accessible queries is in \cite{Oh19}. This recent work exposes the need for further research toward the development of XAI tools capable of explaining ML models while keeping the model's confidentiality in mind. 

Ideally, XAI should be able to explain the knowledge within an AI model and it should be able to reason about what the model acts upon. However, the information revealed by XAI techniques can be used both to generate more effective attacks in adversarial contexts aimed at confusing the model, at the same time as to develop techniques to better protect against private content exposure by using such information. Adversarial attacks \cite{goodfellow2014explaining} try to manipulate a ML algorithm after learning what is the specific information that should be fed to the system so as to lead it to a specific output. For instance, regarding a supervised ML classification model, adversarial attacks try to discover the minimum changes that should be applied to the input data in order to cause a different classification. This has happened regarding computer vision systems of autonomous vehicles; a minimal change in a stop signal, imperceptible to the human eye, led vehicles to detect it as a 45 mph signal \cite{eykholt2017robust}. For the particular case of DL models, available solutions such as Cleverhans \cite{DBLP:journals/corr/GoodfellowPM16} seek to detect adversarial vulnerabilities, and provide different approaches to harden the model against them. Other examples include AlfaSVMLib \cite{Xiao:2015:SVM:2779626.2779777} for SVM models, and AdversarialLib \cite{Biggio:2013:EAA:3120190.3120221} for evasion attacks. There are even available solutions for unsupervised ML, like clustering algorithms \cite{DBLP:journals/corr/abs-1811-09982}.

{\color{black}While XAI techniques can be used to furnish more effective adversarial attacks or to reveal confidential aspects of the model itself, some recent contributions have capitalized on the possibilities of Generative Adversarial Networks (GANs \cite{pan2019recent}), Variational Autoencoders \cite{charte2018practical} and other generative models towards explaining data-based decisions. Once trained, generative models can generate instances of what they have learned based on a noise input vector that can be interpreted as a latent representation of the data at hand. By manipulating this latent representation and examining its impact on the output of the generative model, it is possible to draw insights and discover specific patterns related to the class to be predicted. This generative framework has been adopted by several recent studies \cite{baumgartner2018visual,biffi2018learning} mainly as an attribution method to relate a particular output of a Deep Learning model to their input variables. Another interesting research direction is the use of generative models for the creation of counterfactuals, i.e., modifications to the input data that could eventually alter the original prediction of the model \cite{1907.03077}. Counterfactual prototypes help the user understand the performance boundaries of the model under consideration for his/her improved trust and informed criticism. In light of this recent trend, we definitely believe that there is road ahead for generative ML models to take their part in scenarios demanding understandable machine decisions.}

\subsection{XAI and Output Confidence} \label{ssec:output_confidence}

Safety issues have also been studied in regards to processes that depend on the output of AI models, such as vehicular perception and self-driving in autonomous vehicles, automated surgery, data-based support for medical diagnosis, insurance risk assessment and cyber-physical systems in manufacturing, among others \cite{varshney2017safety}. In all these scenarios erroneous model outputs can lead to harmful consequences, which has yielded comprehensive regulatory efforts aimed at ensuring that no decision is made solely on the basis of data processing \cite{goodman2017Fair}. 

In parallel, research has been conducted towards minimizing both risk and uncertainty of harms derived from decisions made on the output of a ML model. As a result, many techniques have been reported to reduce such a risk, among which we pause at the evaluation of the model's output confidence to decide upon. In this case, the inspection of the share of epistemic uncertainty (namely, the uncertainty due to lack of knowledge) of the input data and its correspondence with the model's output confidence can inform the user and eventually trigger his/her rejection of the model's output \cite{weiss2004mining,attenberg2015beat}. To this end, explaining via XAI techniques which region of the input data the model is focused on when producing a given output can discriminate possible sources of epistemic uncertainty within the input domain.

\subsection{{\color{black}XAI, Rationale Explanation, and Critical Data Studies}} \label{ssec:critical_data}

When shifting the focus to the research practices seen in Data Science, it has been noted that reproducibility is stringently subject not only to the mere sharing of data, models and results to the community, but also to the availability of information about the full discourse around data collection, understanding, assumptions held and insights drawn from model construction and results' analyses \cite{neff2017critique}. In other words, in order to transform data into a valuable actionable asset, individuals must engage in collaborative sense-making by sharing the context producing their findings, wherein context refers to sets of narrative stories around how data were processed, cleaned, modeled and analyzed. In this discourse we find also an interesting space for the adoption of XAI techniques due to their powerful ability to describe black-box models in an understandable, hence conveyable fashion towards colleagues from Social Science, Politics, Humanities and Legal fields. 

{\color{black}XAI can effectively ease the process of explaining the reasons why a model reached a decision in an accessible way to non-expert users, i.e. the \emph{rationale explanation}.} This confluence of multi-disciplinary teams in projects related to Data Science and the search for methodologies to make them appraise the ethical implications of their data-based choices has been lately coined as Critical Data studies \cite{iliadis2016critical}. It is in this field where XAI can significantly boost the exchange of information among heterogeneous audiences about the knowledge learned by models.

\subsection{XAI and Theory-guided Data Science} \label{ssec:tgds}

We envision an exciting synergy between the XAI realm and \emph{Theory-guided Data Science}, a paradigm exposed in \cite{TGDSkarpatne2017theory} that merges both Data Science and the classic theoretical principles underlying the application/context where data are produced. The rationale behind this rising paradigm is the need for data-based models to generate knowledge that is the prior knowledge brought by the field in which it operates. This means that the model type should be chosen according to the type of relations we intend to encounter. The structure should also follow what is previously known. Similarly, the training approach should not allow for the optimization process to enter regions that are not plausible. Accordingly, regularization terms should stand the prior premises of the field, avoiding the elimination of badly represented true relations for spurious and deceptive false relations. Finally, the output of the model should inform about everything the model has come to learn, allowing to reason and merge the new knowledge with what was already known in the field.

Many examples of the implementation of this approach are currently available with promising results. The studies in \cite{TGDShautier2010finding}-\nocite{TGDSfischer2006predicting,TGDScurtarolo2013high,TGDSwong2009active,TGDSxu2015robust,Lesort:17,TGDSleibo2017view}\cite{TGDSschrodt2015bhpmf} were carried out in diverse fields, showcasing the potential of this new paradigm for data science. Above all, it is relevant to notice the resemblance that all concepts and requirements of Theory-guided Data Science share with XAI. All the additions presented in \cite{TGDSkarpatne2017theory} push toward techniques that would eventually render a model explainable, and furthermore, knowledge consistent. The concept of \emph{knowledge from the beginning}, central to Theory-guided Data Science, must also consider how the knowledge captured by a model should be explained for assessing its compliance with theoretical principles known beforehand. This, again, opens a magnificent window of opportunity for XAI.

{\color{black}\subsection{Guidelines for ensuring Interpretable AI Models} \label{ssec:guidelines}

Recent surveys have emphasized on the multidisciplinary, inclusive nature of the process of making an AI-based model interpretable. Along this process, it is of utmost importance to scrutinize and take into proper account the interests, demands and requirements of all stakeholders interacting with the system to be explained, from the designers of the system to the decision makers consuming its produced outputs and users undergoing the consequences of decisions made therefrom.

Given the confluence of multiple criteria and the need for having the human in the loop, some attempts at establishing the procedural guidelines to implement and explain AI systems have been recently contributed. Among them, we pause at the thorough study in \cite{1906.05684}, which suggests that the incorporation and consideration of explainability in practical AI design and deployment workflows should comprise four major methodological steps:
\begin{enumerate}[leftmargin=*]
\item Contextual factors, potential impacts and domain-specific needs must be taken into account when devising an approach to interpretability: These include a thorough understanding of the purpose for which the AI model is built, the complexity of explanations that are required by the audience, and the performance and interpretability levels of existing technology, models and methods. The latter pose a reference point for the AI system to be deployed in lieu thereof.

\item Interpretable techniques should be preferred when possible: when considering explainability in the development of an AI system, the decision of which XAI approach should be chosen should gauge domain-specific risks and needs, the available data resources and existing domain knowledge, and the suitability of the ML model to meet the requirements of the computational task to be addressed. It is in the confluence of these three design drivers where the guidelines postulated in \cite{1906.05684} (and other studies in this same line of thinking \cite{1811.10154}) recommend first the consideration of standard interpretable models rather than sophisticated yet opaque modeling methods. In practice, the aforementioned aspects (contextual factors, impacts and domain-specific needs) can make transparent models preferable over complex modeling alternatives whose interpretability require the application of post-hoc XAI techniques. By contrast, black-box models such as those reviewed in this work (namely, support vector machines, ensemble methods and neural networks) should be selected only when their superior modeling capabilities fit best the characteristics of the problem at hand. 

\item If a black-box model has been chosen, the third guideline establishes that ethics-, fairness- and safety-related impacts should be weighed. Specifically, responsibility in the design and implementation of the AI system should be ensured by checking whether such identified impacts can be mitigated and counteracted by supplementing the system with XAI tools that provide the level of explainability required by the domain in which it is deployed. To this end, the third guideline suggests 1) a detailed articulation, examination and evaluation of the applicable explanatory strategies, 2) the analysis of whether the coverage and scope of the available explanatory approaches match the requirements of the domain and application context where the model is to be deployed; and 3) the formulation of an interpretability action plan that sets forth the explanation delivery strategy, including a detailed time frame for the execution of the plan, and a clearance of the roles and responsibilities of the team involved in the workflow. 

\item Finally, the fourth guideline encourages to rethink interpretability in terms of the cognitive skills, capacities and limitations of the individual human. This is an important question on which studies on measures of explainability are intensively revolving by considering human mental models, the accessibility of the audience to vocabularies of explanatory outcomes, and other means to involve the expertise of the audience into the decision of what explanations should provide.
\end{enumerate}

We foresee that the set of guidelines proposed in \cite{1906.05684} and summarized above will be complemented and enriched further by future methodological studies, ultimately heading to a more \emph{responsible} use of AI. Methodological principles ensure that the purpose for which explainability is pursued is met by bringing the manifold of requirements of all participants into the process, along with other universal aspects of equal relevance such as no discrimination, sustainability, privacy or accountability. A challenge remains in harnessing the potential of XAI to realize a \emph{Responsible AI}, as we discuss in the next section.}

\section{Toward Responsible AI: Principles of Artificial Intelligence, Fairness, Privacy and Data Fusion} \label{sec:responsibleAI}

Over the years many organizations, both private and public, have published guidelines to indicate how AI should be developed and used. These guidelines are commonly referred to as AI \emph{principles}, and they tackle issues related to potential AI threats to both individuals and to the society as a whole. This section presents some of the most important and widely recognized principles in order to link XAI -- which normally appears inside its own principle -- to all of them. Should a responsible implementation and use of AI models be sought in practice, it is our firm claim that XAI does not suffice on its own. Other important principles of Artificial Intelligence such as privacy and fairness must be carefully addressed in practice. In the following sections we elaborate on the concept of Responsible AI, along with the implications of XAI and data fusion in the fulfillment of its postulated principles.

\subsection{Principles of Artificial Intelligence} \label{ssec:principlesAI}

A recent review of some of the main AI principles published since 2016 appears in \cite{fjeld2019principled}. In this work, the authors show a visual framework where different organizations are classified according to the following parameters:
\begin{itemize}[leftmargin=*]
\item Nature, which could be private sector, government, inter-governmental organization, civil society or multistakeholder.

\item Content of the principles: eight possible principles such as privacy, explainability, or fairness. They also consider the coverage that the document grants for each of the considered principles.

\item Target audience: to whom the principles are aimed. They are normally for the organization that developed them, but they could also be destined for another audience (see Figure \ref{fig:audiences}).

\item Whether or not they are rooted on the International Human Rights, as well as whether they explicitly talk about them.
\end{itemize}

For instance, \cite{benjamins2019responsible} is an illustrative example of a document of AI principles for the purpose of this overview, since it accounts for some of the most common principles, and deals explicitly with explainability. Here, the authors propose five principles mainly to guide the development of AI within their company, while also indicating that they could also be used within other organizations and businesses. 

The authors of those principles aim to develop AI in a way that it directly reinforces inclusion, gives equal opportunities for everyone, and contributes to the common good. To this end, the following aspects should be considered:
\begin{itemize}[leftmargin=*]
\item The outputs after using AI systems should not lead to any kind of discrimination against individuals or collectives in relation to race, religion, gender, sexual orientation, disability, ethnic, origin or any other personal condition. Thus, a fundamental criteria to consider while optimizing the results of an AI system is not only their outputs in terms of error optimization, but also how the system deals with those groups. This defines the principle of \textit{Fair AI}.  

\item People should always know when they are communicating with a person, and when they are communicating with an AI system. People should also be aware if their personal information is being used by the AI system and for what purpose. It is crucial to ensure a certain level of understanding about the decisions taken by an AI system. This can be achieved through the usage of XAI techniques. It is important that the generated explanations consider the profile of the user that will receive those explanations (the so-called \emph{audience} as per the definition given in Subsection \ref{sec:what}) in order to adjust the transparency level, as indicated in \cite{theodorou2017designing}. This defines the principle of \textit{Transparent and Explainable AI}.

\item AI products and services should always be aligned with the United Nation's Sustainable Development Goals \cite{RePEc:ess:wpaper:id:7559} and contribute to them in a positive and tangible way. Thus, AI should always generate a benefit for humanity and the common good. This defines the principle of \textit{Human-centric AI} (also referred to as \textit{AI for Social Good} \cite{Hager:19}).

\item AI systems, specially when they are fed by data, should always consider privacy and security standards during all of its life cycle. This principle is not exclusive of AI systems since it is shared with many other software products. Thus, it can be inherited from processes that already exist within a company. This defines the principle of \textit{Privacy and Security by Design}, which was also identified as one of the core ethical and societal challenges faced by Smart Information Systems under the Responsible Research and Innovation paradigm (RRI, \cite{stahl2018ethics}). RRI refers to a package of methodological guidelines and recommendations aimed at considering a wider context for scientific research, from the perspective of the lab to global societal challenges such as sustainability, public engagement, ethics, science education, gender equality, open access, and governance. Interestingly, RRI also requires openness and transparency to be ensured in projects embracing its principles, which links directly to the principle of Transparent and Explainable AI mentioned previously.

\item The authors emphasize that all these principles should always be extended to any third-party (providers, consultants, partners...).
\end{itemize}

Going beyond the scope of these five AI principles, the European Commission (EC) has recently published ethical guidelines for Trustworthy AI \cite{hleg2019high} through an assessment checklist that can be completed by different profiles related to AI systems (namely, product managers, developers and other roles). The assessment is based in a series of principles: 1) human agency and oversight; 2) technical robustness and safety; 3) privacy and data governance; 4) transparency, diversity, non-discrimination and fairness; 5) societal and environmental well-being; 6) accountability. These principles are aligned with the ones detailed in this section, though the scope for the EC principles is more general, including any type of organization involved in the development of AI. 

It is worth mentioning that most of these AI principles guides directly approach XAI as a key aspect to consider and include in AI systems. In fact, the overview for these principles introduced before \cite{fjeld2019principled}, indicates that 28 out of the 32 AI principles guides covered in the analysis, explicitly include XAI as a crucial component. Thus, the work and scope of this article deals directly with one of the most important aspects regarding AI at a worldwide level.

\subsection{Fairness and Accountability} \label{ssec:fairnessaccount}

As mentioned in the previous section, there are many critical aspects, beyond XAI, included within the different AI principles guidelines published during the last decade. However, those aspects are not completely detached from XAI; in fact, they are intertwined. This section presents two key components with a huge relevance within the AI principles guides, Fairness and Accountability. It also highlights how they are connected to XAI.

\subsubsection{Fairness {\color{black}and Discrimination}} 

It is in the identification of implicit correlations between protected and unprotected features where XAI techniques find their place within discrimination-aware data mining methods. By analyzing how the output of the model behaves with respect to the input feature, the model designer may unveil hidden correlations between the input variables amenable to cause discrimination. XAI techniques such as SHAP \cite{lundberg2017unified} could be used to generate counterfactual outcomes explaining the decisions of a ML model when fed with protected and unprotected variables.

Recalling the Fair AI principle introduced in the previous section, \cite{benjamins2019responsible} reminds that fairness is a discipline that generally includes proposals for bias detection within datasets regarding sensitive data that affect protected groups (through variables like gender, race...). {\color{black}Indeed, ethical concerns with black-box models arise from their tendency to unintentionally create unfair decisions by considering sensitive factors such as the individual's race, age or gender \cite{d2017conscientious}. Unfortunately, such unfair decisions can give rise to discriminatory issues, either by explicitly considering sensitive attributes or implicitly by using factors that correlate with sensitive data. In fact, an attribute may implicitly encode a protected factor, as occurs with postal code in credit rating \cite{barocas2016big}. The aforementioned proposals centered on fairness aspects permit to discover correlations between non-sensitive variables and sensitive ones, detect imbalanced outcomes from the algorithms that penalize a specific subgroup of people {\color{black}(\emph{discrimination})}, and mitigate the effect of bias on the model's decisions. These approaches can deal with:}
\begin{itemize}[leftmargin=*]
\item Individual fairness: here, fairness is analyzed by modeling the differences between each subject and the rest of the population.

\item Group fairness: it deals with fairness from the perspective of all individuals.

\item Counterfactual fairness: it tries to interpret the causes of bias using, for example, causal graphs.
\end{itemize}

The sources for bias, as indicated in \cite{barocas2016big}, can be traced to:
\begin{itemize}[leftmargin=*]
\item Skewed data: bias within the data acquisition process.

\item Tainted data: errors in the data modelling definition, wrong feature labelling, and other possible causes.

\item Limited features: using too few features could lead to an inference of false feature relationships that can lead to bias.

\item Sample size disparities: when using sensitive features, disparities between different subgroups can induce bias.

\item Proxy features: there may be correlated features with sensitive ones that can induce bias even when the sensitive features are not present in the dataset.
\end{itemize}

The next question that can be asked is what criteria could be used to define when AI is not biased. For supervised ML, \cite{hardt2016equality} presents a framework that uses three criteria to evaluate group fairness when there is a sensitive feature present within the dataset:
\begin{itemize}[leftmargin=*]
\item Independence: this criterion is fulfilled when the model predictions are independent of the sensitive feature. Thus, the proportion of positive samples (namely, those ones belonging to the class of interest) given by the model is the same for all the subgroups within the sensitive feature.

\item Separation: it is met when the model predictions are independent of the sensitive feature given the target variable. For instance, in classification models, the True Positive (TP) rate and the False Positive (FP) rate are the same in all the subgroups within the sensitive feature. This criteria is also known as \emph{Equalized Odds}.

\item Sufficiency: it is accomplished when the target variable is independent of the sensitive feature given the model output. Thus, the Positive Predictive Value is the same for all subgroups within the sensitive feature. This criteria is also known as Predictive Rate Parity.
\end{itemize}

Although not all of the criteria can be fulfilled at the same time, they can be optimized together in order to minimize the bias within the ML model.

There are two possible actions that could be used in order to achieve those criteria. On one hand, evaluation includes measuring the amount of bias present within the model (regarding one of the criteria aforementioned). There are many different metrics that can be used, depending on the criteria considered. Regarding independence criterion, possible metrics are \textit{statistical parity difference} or \textit{disparate impact}. In case of the separation criterion, possible metrics are \textit{equal opportunity difference} and \textit{average odds difference} \cite{hardt2016equality}. Another possible metric is the \textit{Theil index} \cite{speicher2018unified}, which measures inequality both in terms of individual and group fairness. 

On the other hand, mitigation refers to the process of fixing some aspects in the model in order to remove the effect of the bias in terms of one or several sensitive features. Several techniques exist within the literature, classified in the following categories:  
    \begin{itemize}[leftmargin=*]
    \item Pre-processing: these groups of techniques are applied before the ML model is trained, looking to remove the bias at the first step of the learning process. An example is Reweighing \cite{kamiran2012data}, which modifies the weights of the features in order to remove discrimination in sensitive attributes. Another example is \cite{zemel2013learning}, which hinges on transforming the input data in order to find a good representation that obfuscates information about membership in sensitive features.
    
    \item In-processing: these techniques are applied during the training process of the ML model. Normally, they include Fairness optimization constraints along with cost functions of the ML model. An example is Adversarial Debiasing, \cite{zhang2018mitigating}. This technique optimizes jointly the ability of predicting the target variable while minimizing the ability of predicting sensitive features using a GAN.
    
    \item Post-processing: these techniques are applied after the ML model is trained. They are less intrusive because they do not modify the input data or the ML model. An example is Equalized Odds \cite{hardt2016equality}. This techniques allows to adjust the thresholds in the classification model in order to reduce the differences between the TP rate and the FP rate for each sensitive subgroup.
    \end{itemize}

Even though these references apparently address an AI principle that appears to be independent of XAI, the literature shows that they are intertwined. For instance, the survey in \cite{fjeld2019principled} evinces that 26 out of the 28 AI principles that deal with XAI, also talk about fairness explicitly. This fact elucidates that organizations usually consider both aspects together when implementing Responsible AI. 

The literature also exploses that XAI proposals can be used for bias detection. For example, \cite{ahn2019fairsight} proposes a framework to visually analyze the bias present in a model (both for individual and group fairness). Thus, the fairness report is shown just like the visual summaries used within XAI. This explainability approach eases the understanding and measurement of bias. The system must report that there is bias, justify it quantitatively, indicate the degree of fairness, and explain why a user or group would be treated unfairly with the available data. {\color{black}Similarly, XAI techniques such as SHAP \cite{lundberg2017unified} could be used to generate counterfactual outcomes explaining the decisions of a ML model when fed with protected and unprotected variables. By identifying implicit correlations between protected and unprotected features through XAI techniques, the model designer may unveil hidden correlations between the input variables amenable to cause discrimination.} 

Another example is \cite{soares2019fair}, where the authors propose a fair-by-design approach in order to develop ML models that jointly have less bias and include as explanations human comprehensible rules. The proposal is based in self-learning locally generative models that use only a small part of the whole dataset available (weak supervision). It first finds recursively relevant prototypes within the dataset, and extracts the empirical distribution and density of the points around them. Then it generates rules in an IF/THEN format that explain that a data point is classified within a specific category because it is \textit{similar} to some prototypes. The proposal then includes an algorithm that both generates explanations and reduces bias, as it is demonstrated for the use case of recidivism using the Correctional Offender Management Profiling for Alternative Sanctions (COMPAS) dataset \cite{dressel2018accuracy}. The same goal has been recently pursued in \cite{aivodji2019fairwashing}, showing that post-hoc XAI techniques 
can forge fairer explanations from truly unfair black-box models. Finally, CERTIFAI (Counterfactual Explanations for Robustness, Transparency, Interpretability, and Fairness of Artificial Intelligence models) \cite{sharma2019certifai} uses a customized genetic algorithm to generate counterfactuals that can help to see the robustness of a ML model, generate explanations, and examine fairness (both at the individual level and at the group level) at the same time.

{\color{black}Strongly linked to the concept of fairness, much attention has been lately devoted to the concept of \emph{data diversity}, which essentially refers to the capability of an algorithmic model to ensure that all different types of objects are represented in its output \cite{drosou2017diversity}. Therefore, diversity can be thought to be an indicator of the quality of a collection of items that, when taking the form of a model's output, can quantify the proneness of the model to produce diverse results rather than highly accurate predictions. Diversity comes into play in human-centered applications with ethical restrictions that permeate to the AI modeling phase \cite{lerman2013big}. Likewise, certain AI problems (such as content recommendation or information retrieval) also aim at producing diverse recommendations rather than highly-scoring yet similar results \cite{agrawal2009diversifying,smyth2001similarity}. In these scenarios, dissecting the internals of a black-box model via XAI techniques can help identifying the capability of the model to maintain the input data diversity at its output. Learning strategies to endow a model with diversity keeping capabilities could be complemented with XAI techniques in order to shed transparency over the model internals, and assess the effectiveness of such strategies with respect to the diversity of the data from which the model was trained. Conversely, XAI could help to discriminate which parts of the model are compromising its overall ability to preserve diversity.}

\subsubsection{Accountability}

Regarding accountability, the EC \cite{hleg2019high} defines the following aspects to consider:
    \begin{itemize}[leftmargin=*]
    \item Auditability: it includes the assessment of algorithms, data and design processes, but preserving the intellectual property related to the AI systems. Performing the assessment by both internal and external auditors, and making the reports available, could contribute to the trustworthiness of the technology. When the AI system affects fundamental rights, including safety-critical applications, it should always be audited by an external third party.
    
    \item Minimization and reporting of negative impacts: it consists of reporting actions or decisions that yield a certain outcome by the system. It also comprises the assessment of those outcomes and how to respond to them. To address that, the development of AI systems should also consider the identification, assessment, documentation and minimization of their potential negative impacts. In order to minimize the potential negative impact, impact assessments should be carried out both prior to and during the development, deployment and use of AI systems. It is also important to guarantee protection for anyone who raises concerns about an AI system (e.g., \emph{whistle-blowers}). All assessments must be proportionate to the risk that the AI systems pose.
    
    \item Trade-offs: in case any tension arises due to the implementation of the above requirements, trade-offs could be considered but only if they are ethically acceptable. Such trade-offs should be reasoned, explicitly acknowledged and documented, and they must be evaluated in terms of their risk to ethical principles. The decision maker must be accountable for the manner in which the appropriate trade-off is being made, and the trade-off decided should be continually reviewed to ensure the appropriateness of the decision. If there is no ethically acceptable trade-off, the development, deployment and use of the AI system should not proceed in that form.
    
    \item Redress: it includes mechanisms that ensure an adequate redress for situations when unforeseen unjust adverse impacts take place. Guaranteeing a redress for those non-predicted scenarios is a key to ensure trust. Special attention should be paid to vulnerable persons or groups. 
    \end{itemize}

These aspects addressed by the EC highlight different connections of XAI with accountability. First, XAI contributes to auditability as it can help explaining AI systems for different profiles, including regulatory ones. Also, since there is a connection between fairness and XAI as stated before, XAI can also contribute to the minimization and report of negative impacts.

\subsection{Privacy and Data Fusion} \label{ssec:privacydatafusion}

The ever-growing number of information sources that nowadays coexist in almost all domains of activity calls for data fusion approaches aimed at exploiting them simultaneously toward solving a learning task. By merging heterogeneous information, data fusion has been proven to improve the performance of ML models in many applications{\color{black}, such as industrial prognosis \cite{diez2019data}, cyber-physical social systems \cite{wang2019data} or the Internet of Things \cite{ding2019survey}, among others}. This section speculates with the potential of data fusion techniques to enrich the explainability of ML models, and to compromise the privacy of the data from which ML models are learned. To this end, we briefly overview different data fusion paradigms, and later analyze them from the perspective of data privacy. As we will later, despite its relevance in the context of Responsible AI, the confluence between XAI and data fusion is an uncharted research area in the current research mainstream. 

\subsubsection{Basic Levels of Data Fusion} \label{sssec:levels_data_fusion}

We depart from the different levels of data fusion that have been identified in comprehensive surveys on the matter \cite{SMIRNOV201931,DING2019129,WANG201942,LAU2019357}. In the context of this subsection, we will distinguish among fusion at data level, fusion at model level and fusion at knowledge level. Furthermore, a parallel categorization can be established depending on where such data is processed and fused, yielding centralized and distributed methods for data fusion. In a centralized approach, nodes deliver their locally captured data to a centralized processing system to merge them together. In contrast, in a distributed approach, each of the nodes merges its locally captured information, eventually sharing the result of the local fusion with its counterparts.

\begin{figure}[ht]
	\center{\includegraphics[width=\columnwidth]{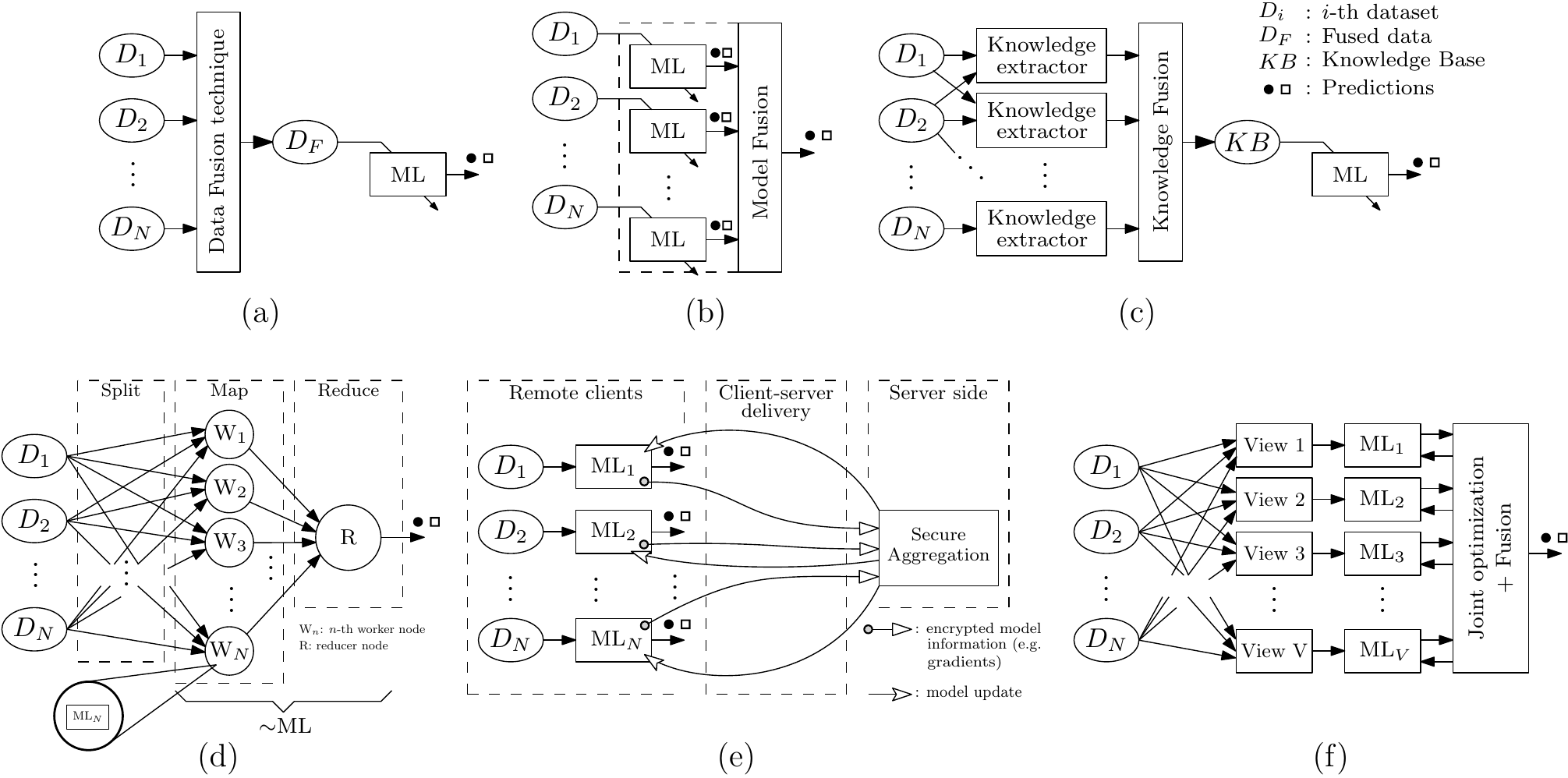}}
	\caption{Diagrams showing different levels at which data fusion can be performed: (a) data level; (b) model level; (c) knowledge level; (d) Big Data fusion; (e) Federated Learning and (f) Multiview Learning.}
	\label{fig:fusion_all}
\end{figure}

Fusion through the information generation process has properties and peculiarities depending on the level at which the fusion is performed. At the so-called \emph{data level}, fusion deals with raw data. As schematically shown in Figure \ref{fig:fusion_all}, a fusion model at this stage receives raw data from different information sources, and combines them to create a more coherent, compliant, robust or simply representative data flow. On the other hand, fusion at the \emph{model level} aggregates models, each learned from a subset of the data sets that were to be fused. Finally, at the \emph{knowledge level} the fusion approach deals with knowledge in the form of rules, ontologies or other knowledge representation techniques with the intention of merging them to create new, better or more complete knowledge from what was originally provided. Structured knowledge information is extracted from each data source and for every item in the data set using multiple \emph{knowledge extractors} (e.g. a reasoning engine operating on an open semantic database). All produced information is then fused to further ensure the quality, correctness and manageability of the produced knowledge about the items in the data set.

Other data fusion approaches exist beyonds the ones represented in Figure \ref{fig:fusion_all}. As such, data-level fusion can be performed either by a technique specifically devoted to this end (as depicted in Figure \ref{fig:fusion_all}.b) or, instead, performed along the learning process of the ML model (as done in e.g. DL models). Similarly, model-level data fusion can be made by combining the decisions of different models (as done in tree ensembles).

\subsubsection{Emerging Data Fusion Approaches} \label{sssec:advanced_levels_data_fusion}

In the next subsection we examine other data fusion approaches that have recently come into scene due to their implications in terms of data privacy:
\begin{itemize}[leftmargin=*]
\item In Big Data fusion (Figure \ref{fig:fusion_all}.d), local models are learned on a split of the original data sources, each submitted to a Worker node in charge of performing this learning process (\emph{Map} task). Then, a \emph{Reduce} node (or several \emph{Reduce} nodes, depending on the application) combines the outputs produced by each \emph{Map} task. Therefore, Big Data fusion can be conceived as a means to distribute the complexity of learning a ML model over a pool of Worker nodes, wherein the strategy to design how information/models are fused together between the \emph{Map} and the \emph{Reduce} tasks is what defines the quality of the finally generated outcome \cite{ramirez2018big}. 
\item By contrast, in Federated Learning \cite{1610.02527,mcmahan2017communication,konevcny2016federated}, the computation of ML models is made on data captured locally by remote client devices (Figure \ref{fig:fusion_all}.e). Upon local model training, clients transmit encrypted information about their learned knowledge to a central server, which can take the form of layer-wise gradients (in the case of neural ML models) or any other model-dependent content alike. The central server aggregates (fuses) the knowledge contributions received from all clients to yield a shared model harnessing the collected information from the pool of clients. It is important to observe that no client data is delivered to the central server, which elicits the privacy-preserving nature of Federated Learning. Furthermore, computation is set closer to the collected data, which reduces the processing latency and alleviates the computational burden of the central server.

\item Finally, Multiview Learning \cite{sun2013survey} constructs different \emph{views} of the object as per the information contained in the different data sources (Figure \ref{fig:fusion_all}.f). These views can be produced from multiple sources of information and/or different feature subsets \cite{zhang2019feature}. Multiview Learning devises strategies to jointly optimize ML models learned from the aforementioned views to enhance the generalization performance, specially in those applications with weak data supervision and hence, prone to model overfitting. This joint optimization resorts to different algorithmic means, from co-training to co-regularization \cite{zhao2017multi}. 
\end{itemize}

\subsubsection{Opportunities and Challenges in Privacy and Data Fusion under the Responsible AI Paradigm} \label{sssec:opportunities_data_fusion}

AI systems, specially when dealing with multiple data sources, need to explicitly include privacy considerations during the system's life cycle. This is specially critical when working with personal data, because respecting people's right to privacy should always be addressed. The EC highlights that privacy should also address data governance, covering the quality and integrity of the used data \cite{hleg2019high}. It should also include the definition of access protocols and the capability to process data in a way that ensures privacy. The EC guide breaks down the privacy principle into three aspects:
\begin{itemize}[leftmargin=*]
\item Privacy and data protection: they should be guaranteed in AI systems throughout its entire lifecycle. It includes both information provided by users and information generated about those users derived from their interactions with the system. Since digital information about a user could be used in a negative way against them (discrimination due to sensitive features, unfair treatment...), it is crucial to ensure proper usage of all the data collected.

\item Quality and integrity of data: quality of data sets is fundamental to reach good performance with AI systems that are fueled with data, like ML. However, sometimes the data collected contains socially constructed biases, inaccuracies, errors and mistakes. This should be tackled before training any model with the data collected. Additionally, the integrity of the data sets should be ensured. 

\item Access to data: if there is individual personal data, there should always be data protocols for data governance. These protocols should indicate who may access data and under which circumstances. 
\end{itemize}

The aforementioned examples from the EC shows how data fusion is directly intertwined with privacy and with fairness, regardless of the technique employed for it. 

Notwithstanding this explicit concern from regulatory bodies, loss of privacy has been compromised by DL methods in scenarios where no data fusion is performed. For instance, a few images are enough to threaten users' privacy even in the presence of image obfuscation \cite{Oh16}, and the model parameters of a DNN can be exposed by simply performing input queries on the model \cite{Orekondy18,Oh19}. An approach to explain loss of privacy is by using \textit{privacy loss} and \textit{intent loss} subjective scores. The former provides a subjective measure of the severity of the privacy violation depending on the role of a face in the image, while the latter captures the intent of the bystanders to appear in the picture. These kind of explanations have motivated, for instance, secure matching cryptographic protocols for photographer and bystanders to preserve privacy \cite{Orekondy18, Aditya16, sun2018hybrid}. We definite advocate for more efforts invested in this direction, namely, in ensuring that XAI methods do not pose a threat in regards to the privacy of the data used for training the ML model under target.

When data fusion enters the picture, different implications arise with the context of explainability covered in this survey. To begin with, classical techniques for fusion at the data level only deal with data and have no connection to the ML model, so they have little to do with explainability. However, the advent of DL models has blurred the distinction between information fusion and predictive modeling. The first layers of DL architectures are in charge of learning high-level features from raw data that possess relevance for the task at hand. This learning process can be thought to aim at solving a data level fusion problem, yet in a directed learning fashion that makes the fusion process tightly coupled to the task to be solved. 

In this context, many techniques in the field of XAI have been proposed to deal with the analysis of correlation between features. This paves the way to explaining how data sources are actually fused through the DL model, which can yield interesting insights on how the predictive task at hand induces correlations among the data sources over the spatial and/or time domain. Ultimately, this gained information on the fusion could not only improve the usability of the model as a result of its enhanced understanding by the user, but could also help identifying other data sources of potential interest that could be incorporated to the model, or even contribute to a more efficient data fusion in other contexts. 

Unfortunately, this previously mentioned concept of fusion at data level contemplates data under certain constraints of known form and source origin. As presented in \cite{dong2013big}, the Big Data era presents an environment in which these premises cannot be taken for granted, and methods to board Big Data fusion (as that illustrated in Figure \ref{fig:fusion_all}.d) have to be thought. Conversely, a concern with model fusion context emerges in the possibility that XAI techniques could be explanatory enough to compromise the confidentiality of private data. This could eventually occur if sensitive information (e.g. ownership) could be inferred from the explained fusion among protected and unprotected features.

When turning our prospects to data fusion at model level, we have already argued that the fusion of the outputs of several transparent models (as in tree ensembles) could make the overall model opaque, thereby making it necessary to resort to post-hoc explainability solutions. However, model fusion may entail other drawbacks when endowed with powerful post-hoc XAI techniques. Let us imagine that relationships of a model's input features have been discovered by means of a post-hoc technique) and that one of those features is hidden or unknown. Will it be possible to infer another model's features if that previous feature was known to be used in that model? Would this possibility uncover a problem as privacy breaches in cases in which related protected input variables are not even shared in the first place?

To get the example clearer, in \cite{zhang2015comobile} a multiview perspective is utilized in which different single views (representing the sources they attend to) models are fused. These models contain among others, cell-phone data, transportation data, etc. which might introduce the problem that information that is not even shared can be discovered through other sources that are actually shared. In the example above, what if instead of features, a model shares with another a layer or part of its architecture as in Federated Learning? Would this sharing make possible to infer information from that exchanged part of its model, to the extent of allowing for the design of adversarial attacks with better success rate upon the antecedent model? 

If focused at knowledge level fusion, a similar reasoning holds: XAI comprises techniques that extract knowledge from ML model(s). This ability to explain models could have an impact on the necessity of discovering new knowledge through the complex interactions formed within ML models. If so, XAI might enrich knowledge fusion paradigms, bringing the possibility of discovering new knowledge extractors of relevance for the task at hand. For this purpose, it is of paramount importance that the knowledge extracted from a model by means of XAI techniques can be understood and extrapolated to the domain in which knowledge extractors operate. The concept matches with ease with that of transfer learning portrayed in \cite{pan2009survey}. Although XAI is not contemplated in the surveyed processes of extracting knowledge from models trained in certain feature spaces and distributions, to then be utilized in environments where previous conditions do not hold, when deployed, XAI can pose a threat if the explanations given about the model can be reversely engineered through the knowledge fusion paradigm to eventually compromise, for instance, the differential privacy of the overall model.

The distinction between centralized and distributed data fusion also spurs further challenges in regards to privacy and explainability. The centralized approach does not bring any further concerns that those presented above. However, distributed fusion does arise new problems. Distributed fusion might be applied for different reasons, mainly due to environmental constraints or due to security or privacy issues. The latter context may indulge some dangers. Among other goals (e.g. computational efficiency), model-level data fusion is performed in a distributed fashion to ensure that no actual data is actually shared, but rather parts of an ML model trained on local data. This rationale lies at the heart of Federated Learning, where models exchange locally learned information among nodes. Since data do not leave the local device, only the transmission of model updates is required across distributed devices. This lightens the training process for network-compromised settings and guarantees data privacy \cite{konevcny2016federated}. Upon the use of post-hoc explainability techniques, a node could disguise sensitive information about the local context in which the received ML model part was trained. In fact, it was shown that a black-box model based on a DNN from which an input/output query interface is given can be used to accurately predict every single hyperparameter value used for training, allowing for potential privacy-related consequences \cite{Oh19, Oh16, Aditya16}. This relates to studies showing that blurring images does not guarantee privacy preservation.

Data fusion, privacy and model explainability are concepts that have not been analysed together so far. From the above discussion it is clear that there are unsolved concerns and caveats that demand further study by the community in forthcoming times. 

\subsection{Implementing Responsible AI Principles in an Organization}

While increasingly more organizations are publishing AI principles to declare that they care about avoiding unintended negative consequences, there is much less experience on how to actually implement the principles into an organization. Looking at several examples of principles declared by different organizations \cite{fjeld2019principled}, we can divide them into two groups:
\begin{itemize}[leftmargin=*]
\item AI-specific principles that focus on aspects that are specific to AI, such as explainability, fairness and human agency.
\item End-to-end principles that cover all aspects involved in AI, including also privacy, security and safety. 
\end{itemize}

The EC Guidelines for Trustworthy AI are an example of end-to-end principles \cite{hleg2019high}, while those of Telefonica (a large Spanish ICT company operating worldwide) are more AI-specific \cite{benjamins2019responsible}. For example, safety and security are relevant for any connected IT system, and therefore also for AI systems. The same holds for privacy, but it is probably true that privacy in the context of AI systems is even more important than for general IT systems, due to the fact that ML models need huge amounts of data and most importantly, because XAI tools and data fusion techniques pose new challenges to preserve the privacy of protected records. 

When it comes to implement the AI Principles into an organization, it is important to operationalize the AI-specific parts and, at the same time, leverage the processes already existing for the more generic principles. Indeed, in many organizations there already exist norms and procedures for privacy, security and safety. Implementing AI principles requires a methodology such as that presented in \cite{benjamins2019responsible} that breaks down the process into different parts. The ingredients of such a methodology should include, at least:
\begin{itemize}[leftmargin=*]
\item AI principles (already discussed earlier), which set the values and boundaries.
\item Awareness and training about the potential issues, both technical and non-technical.
\item A questionnaire that forces people to think about certain impacts of the AI system {\color{black}(\emph{impact explanation})}. This questionnaire should give concrete guidance on what to do if certain undesired impacts are detected.
\item Tools that help answering some of the questions, and help mitigating any problems identified. XAI tools and fairness tools fall in this category{\color{black}, as well as other recent proposals such as \emph{model cards} \cite{mitchell2019model}.}
\item A governance model assigning responsibilities and accountabilities {\color{black}(\emph{responsibility explanation})}. There are two philosophies for governance: 1) based on committees that review and approve AI developments, and 2) based on the self-responsibility of the employees. While both are possible, given the fact that agility is key for being successful in the digital world, it seems wiser to focus on awareness and employee responsibility, and only use committees when there are specific, but important issues.
\end{itemize}

From the above elaborations, it is clear that the implementation of Responsible AI principles in companies should balance between two requirements: 1) major cultural and organizational changes needed to enforce such principles over processes endowed with AI functionalities; and 2) the feasibility and compliance of the implementation of such principles with the IT assets, policies and resources already available at the company. It is in the gradual process of rising corporate awareness around the principles and values of Responsible AI where we envision that XAI will make its place and create huge impact.

\section{Conclusions and Outlook} \label{sec:conc}

This overview has revolved around eXplainable Artificial Intelligence (XAI), which has been identified in recent times as an utmost need for the adoption of ML methods in real-life applications. Our study has elaborated on this topic by first clarifying different concepts underlying model explainability, as well as by showing the diverse purposes that motivate the search for more interpretable ML methods. These conceptual remarks have served as a solid baseline for a systematic review of recent literature dealing with explainability, which has been approached from two different perspectives: 1) ML models that feature some degree of transparency, thereby interpretable to an extent by themselves; and 2) post-hoc XAI techniques devised to make ML models more interpretable. This literature analysis has yielded a global taxonomy of different proposals reported by the community, classifying them under uniform criteria. Given the prevalence of contributions dealing with the explainability of Deep Learning models, we have inspected in depth the literature dealing with this family of models, giving rise to an alternative taxonomy that connects more closely with the specific domains in which explainability can be realized for Deep Learning models. 

\begin{figure}[ht]
	\center{\includegraphics[width=\columnwidth]{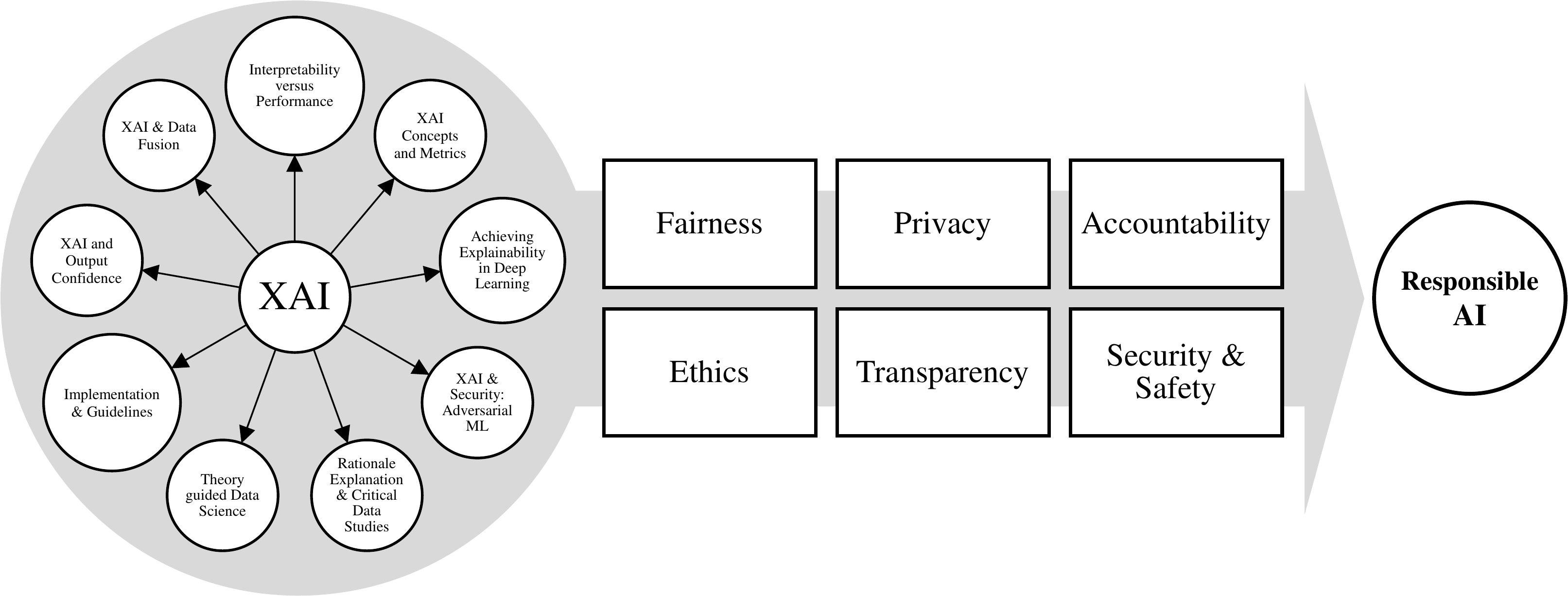}}
	\caption{{\color{black}Summary of XAI challenges discussed in this overview and its impact on the principles for Responsible AI.}}
	\label{fig:challenges_XAI_RAI}
\end{figure}

We have moved our discussions beyond what has been made so far in the XAI realm toward the concept of Responsible AI, a paradigm that imposes a series of AI principles to be met when implementing AI models in practice, including fairness, transparency, and privacy. We have also discussed the implications of adopting XAI techniques in the context of data fusion, unveiling the potential of XAI to compromise the privacy of protected data involved in the fusion process. Implications of XAI in fairness have also been discussed in detail. {\color{black}This vision of XAI as a core concept to ensure the aforementioned principles for Responsible AI is summarized graphically in Figure \ref{fig:challenges_XAI_RAI}.}

Our reflections about the future of XAI, conveyed in the discussions held throughout this work, agree on the compelling need for a proper understanding of the potentiality and caveats opened up by XAI techniques. It is our vision that model interpretability must be addressed jointly with requirements and constraints related to data privacy, model confidentiality, fairness and accountability. A responsible implementation and use of AI methods in organizations and institutions worldwide will be only guaranteed if all these AI principles are studied jointly.

\section*{Acknowledgments}

Alejandro Barredo-Arrieta, Javier Del Ser and Sergio Gil-Lopez would like to thank the Basque Government for the funding support received through the EMAITEK and ELKARTEK programs. {\color{black}Javier Del Ser also acknowledges funding support from the Consolidated Research Group MATHMODE (IT1294-19) granted by the Department of Education of the Basque Government}. Siham Tabik, Salvador Garcia, Daniel Molina and Francisco Herrera would like to thank the Spanish Government for its funding support (SMART-DaSCI project, TIN2017-89517-P), as well as the BBVA Foundation through its \emph{Ayudas Fundaci\'on BBVA a Equipos de Investigaci\'on Cient\'ifica} 2018 call (DeepSCOP project). {\color{black}This work was also funded in part by the European Union's Horizon 2020 research and innovation programme AI4EU under grant agreement 825619}. We also thank Chris Olah, Alexander Mordvintsev and Ludwig Schubert for borrowing images for illustration purposes. Part of this overview is inspired by a preliminary work of the concept of Responsible AI: R. Benjamins, A. Barbado, D. Sierra, \emph{``Responsible AI by Design''}, to appear in the Proceedings of the Human-Centered AI: Trustworthiness of AI Models \& Data (HAI) track at AAAI Fall Symposium, DC, November 7-9, 2019 \cite{benjamins2019responsible}. 

\bibliography{mybibfile}
\end{document}